\documentstyle[12pt,chicagob]{article}

\title{Updating Probabilities}
\author{Peter D. Gr{\"u}nwald\\
CWI, P.O. Box 94079 \\ 1090 GB Amsterdam\\
pdg@cwi.nl\\
www.cwi.nl/\~\/pdg
\and
Joseph Y.\ Halpern\\
Cornell University\\
Ithaca, NY 14853\\
halpern@cs.cornell.edu\\
www.cs.cornell.edu/home/halpern
}
\date{}

\begin{document}
\renewcommand{\mid}{\, | \,}

\renewcommand{\sc}{\bf}

%
%

%

\newtheorem{THEOREM}{Theorem}[section]
\newenvironment{theorem}{\begin{THEOREM} \hspace{-.85em} {\bf :} }%
                        {\end{THEOREM}}
\newtheorem{LEMMA}[THEOREM]{Lemma}
\newenvironment{lemma}{\begin{LEMMA} \hspace{-.85em} {\bf :} }%
                      {\end{LEMMA}}
\newtheorem{COROLLARY}[THEOREM]{Corollary}
\newenvironment{corollary}{\begin{COROLLARY} \hspace{-.85em} {\bf :} }%
                          {\end{COROLLARY}}
\newtheorem{PROPOSITION}[THEOREM]{Proposition}
\newenvironment{proposition}{\begin{PROPOSITION} \hspace{-.85em} {\bf :} }%
                            {\end{PROPOSITION}}
\newtheorem{DEFINITION}[THEOREM]{Definition}
\newenvironment{definition}{\begin{DEFINITION} \hspace{-.85em} {\bf :} \rm}%
                            {\end{DEFINITION}}
\newtheorem{CLAIM}[THEOREM]{Claim}
\newenvironment{claim}{\begin{CLAIM} \hspace{-.85em} {\bf :} \rm}%
                            {\end{CLAIM}}
\newtheorem{EXAMPLE}[THEOREM]{Example}
\newenvironment{example}{\begin{EXAMPLE} \hspace{-.85em} {\bf :} \rm}%
                            {\end{EXAMPLE}}
\newtheorem{REMARK}[THEOREM]{Remark}
\newenvironment{remark}{\begin{REMARK} \hspace{-.85em} {\bf :} \rm}%
                            {\end{REMARK}}

\newcommand{\thm}{\begin{theorem}}
\newcommand{\lem}{\begin{lemma}}
\newcommand{\pro}{\begin{proposition}}
\newcommand{\dfn}{\begin{definition}}
\newcommand{\rem}{\begin{remark}}
\newcommand{\xam}{\begin{example}}
\newcommand{\cor}{\begin{corollary}}
\newcommand{\prf}{\noindent{\bf Proof:} }
\newcommand{\ethm}{\end{theorem}}
\newcommand{\elem}{\end{lemma}}
\newcommand{\epro}{\end{proposition}}
\newcommand{\edfn}{\bbox\end{definition}}
\newcommand{\erem}{\bbox\end{remark}}
\newcommand{\exam}{\bbox\end{example}}
\newcommand{\ecor}{\end{corollary}}
\newcommand{\eprf}{\bbox\vspace{0.1in}}
\newcommand{\beqn}{\begin{equation}}
\newcommand{\eeqn}{\end{equation}}
\newcommand{\wbox}{\mbox{$\sqcap$\llap{$\sqcup$}}}
\newcommand{\bbox}{\vrule height7pt width4pt depth1pt}
\newcommand{\qed}{\eprf}
\newcommand{\clm}{\begin{claim}}
\newcommand{\eclm}{\end{claim}}
\let\member=\in
\let\notmember=\notin
\newcommand{\sub}{_}
\def\su{^}
\newcommand{\rarrow}{\rightarrow}
\newcommand{\larrow}{\leftarrow}
\newcommand{\boldsymbol}[1]{\mbox{\boldmath $\bf #1$}}
\newcommand{\bolda}{{\bf a}}
\newcommand{\boldb}{{\bf b}}
\newcommand{\boldc}{{\bf c}}
\newcommand{\boldd}{{\bf d}}
\newcommand{\bolde}{{\bf e}}
\newcommand{\boldf}{{\bf f}}
\newcommand{\boldg}{{\bf g}}
\newcommand{\boldh}{{\bf h}}
\newcommand{\boldi}{{\bf i}}
\newcommand{\boldj}{{\bf j}}
\newcommand{\boldk}{{\bf k}}
\newcommand{\boldl}{{\bf l}}
\newcommand{\boldm}{{\bf m}}
\newcommand{\boldn}{{\bf n}}
\newcommand{\boldo}{{\bf o}}
\newcommand{\boldp}{{\bf p}}
\newcommand{\boldq}{{\bf q}}
\newcommand{\boldr}{{\bf r}}
\newcommand{\bolds}{{\bf s}}
\newcommand{\boldt}{{\bf t}}
\newcommand{\boldu}{{\bf u}}
\newcommand{\boldv}{{\bf v}}
\newcommand{\boldw}{{\bf w}}
\newcommand{\boldx}{{\bf x}}
\newcommand{\boldy}{{\bf y}}
\newcommand{\boldz}{{\bf z}}
\newcommand{\boldA}{{\bf A}}
\newcommand{\boldB}{{\bf B}}
\newcommand{\boldC}{{\bf C}}
\newcommand{\boldD}{{\bf D}}
\newcommand{\boldE}{{\bf E}}
\newcommand{\boldF}{{\bf F}}
\newcommand{\boldG}{{\bf G}}
\newcommand{\boldH}{{\bf H}}
\newcommand{\boldI}{{\bf I}}
\newcommand{\boldJ}{{\bf J}}
\newcommand{\boldK}{{\bf K}}
\newcommand{\boldL}{{\bf L}}
\newcommand{\boldM}{{\bf M}}
\newcommand{\boldN}{{\bf N}}
\newcommand{\boldO}{{\bf O}}
\newcommand{\boldP}{{\bf P}}
\newcommand{\boldQ}{{\bf Q}}
\newcommand{\boldR}{{\bf R}}
\newcommand{\boldS}{{\bf S}}
\newcommand{\boldT}{{\bf T}}
\newcommand{\boldU}{{\bf U}}
\newcommand{\boldV}{{\bf V}}
\newcommand{\boldW}{{\bf W}}
\newcommand{\boldX}{{\bf X}}
\newcommand{\boldY}{{\bf Y}}
\newcommand{\boldZ}{{\bf Z}}
\newcommand{\sat}{\models}
\newcommand{\dtur}{\models}
\newcommand{\infers}{\vdash}
\newcommand{\stur}{\vdash}
\newcommand{\rimp}{\Rightarrow}
\newcommand{\limp}{\Leftarrow}
\newcommand{\dimp}{\Leftrightarrow}
\newcommand{\bor}{\bigvee}
\newcommand{\band}{\bigwedge}
\newcommand{\union}{\cup}
\newcommand{\inter}{\cap}
\newcommand{\xx}{{\bf x}}
\newcommand{\yy}{{\bf y}}
\newcommand{\uu}{{\bf u}}
\newcommand{\vv}{{\bf v}}
\newcommand{\FF}{{\bf F}}
\newcommand{\natnum}{{\sl N}}
\newcommand{\IR}{\mbox{$I\!\!R$}}
\newcommand{\IP}{\mbox{$I\!\!P$}}
\newcommand{\IN}{\mbox{$I\!\!N$}}
\newcommand{\IC}{\mbox{$C\!\!\!\!\raisebox{.75pt}{\mbox{\sqi I}}$}}
\newcommand{\marrow}{\hbox{$\rightarrow$ \hskip -10pt
                      $\rightarrow$ \hskip 3pt}}
\renewcommand{\phi}{\varphi}
\newcommand{\Circ}{\mbox{{\small $\bigcirc$}}}
\newcommand{\lt}{<}
\newcommand{\gt}{>}
\newcommand{\all}{\forall}
\newcommand{\infinity}{\infty}
\newcommand{\bc}[2]{\left( \begin{array}{c} #1 \\ #2  \end{array} \right)}
\newcommand{\cross}{\times}
\newcommand{\bigfootnote}[1]{{\footnote{\normalsize #1}}}
\newcommand{\medfootnote}[1]{{\footnote{\small #1}}}
\newcommand{\bd}{\bf}


\newcommand{\imp}{\Rightarrow}

\newcommand{\A}{{\cal A}}
\newcommand{\B}{{\cal B}}
\newcommand{\C}{{\cal C}}
\newcommand{\D}{{\cal D}}
\newcommand{\E}{{\cal E}}
\newcommand{\F}{{\cal F}}
\newcommand{\G}{{\cal G}}
\newcommand{\I}{{\cal I}}
\newcommand{\J}{{\cal J}}
\newcommand{\K}{{\cal K}}
\newcommand{\M}{{\cal M}}
\newcommand{\N}{{\cal N}}
\newcommand{\Ocal}{{\cal O}}
\newcommand{\Hcal}{{\cal H}}
\renewcommand{\P}{{\cal P}}
\newcommand{\Q}{{\cal Q}}
\newcommand{\R}{{\cal R}}
\newcommand{\T}{{\cal T}}
\newcommand{\U}{{\cal U}}
\newcommand{\V}{{\cal V}}
\newcommand{\W}{{\cal W}}
\newcommand{\X}{{\cal X}}
\newcommand{\Y}{{\cal Y}}
\newcommand{\Z}{{\cal Z}}

\newcommand{\Kone}{{\cal K}_1}
\newcommand{\abs}[1]{\left| #1\right|}
\newcommand{\set}[1]{\left\{ #1 \right\}}
\newcommand{\Ki}{{\cal K}_i}
\newcommand{\Kn}{{\cal K}_n}
\newcommand{\st}{\, \vert \,} 
\newcommand{\stc}{\, : \,} 
\newcommand{\la}{\langle}
\newcommand{\ra}{\rangle}
\newcommand{\<}{\langle}
\renewcommand{\>}{\rangle}
\newcommand{\lang}{\mbox{${\cal L}_n$}}
\newcommand{\langd}{\mbox{${\cal L}_n^D$}}

\newtheorem{nlem}{Lemma}
\newtheorem{Ob}{Observation}
\newtheorem{pps}{Proposition}
\newtheorem{defn}{Definition}
\newtheorem{crl}{Corollary}
\newtheorem{cl}{Claim}
\newcommand{\pf}{\par\noindent{\bf Proof}~~}
\newcommand{\eg}{e.g.,~}
\newcommand{\ie}{i.e.,~}
\newcommand{\vs}{vs.~}
\newcommand{\cf}{cf.~}
\newcommand{\etal}{et al.\ }
\newcommand{\resp}{resp.\ }
\newcommand{\respc}{resp.,\ }
\newcommand{\comment}[1]{\marginpar{\scriptsize\raggedright #1}}
\newcommand{\wrt}{with respect to~}
\newcommand{\re}{r.e.}
\newcommand{\nind}{\noindent}
\newcommand{\distributed}{distributed\ }
\newcommand{\bn}{\bigskip\markright{NOTES}
\section*{Notes}}
\newcommand{\Exer}{
\bigskip\markright{EXERCISES}
\section*{Exercises}}
\newcommand{\DG}{D_G}
\newcommand{\Sm}{{\rm S5}_m}
\newcommand{\Smc}{{\rm S5C}_m}
\newcommand{\Smi}{{\rm S5I}_m}
\newcommand{\Smic}{{\rm S5CI}_m}
\newcommand{\Martin}{Mart\'\i n\ }
\newcommand{\ol}{\setlength{\itemsep}{0pt}\begin{enumerate}}
\newcommand{\eol}{\end{enumerate}\setlength{\itemsep}{-\parsep}}
\newcommand{\ul}{\setlength{\itemsep}{0pt}\begin{itemize}}
\newcommand{\dl}{\setlength{\itemsep}{0pt}\begin{description}}
\newcommand{\edl}{\end{description}\setlength{\itemsep}{-\parsep}}
\newcommand{\eul}{\end{itemize}\setlength{\itemsep}{-\parsep}}
\newtheorem{fthm}{Theorem}
\newtheorem{flem}[fthm]{Lemma}
\newtheorem{fcor}[fthm]{Corollary}
\newcommand{\slidehead}[1]{
\eject
\Huge
\begin{center}
{\bf #1 }
\end{center}
\vspace{.5in}
\LARGE}

\newcommand{\subG}{_G}
\newcommand{\If}{{\bf if}}

\newcommand{\attime}{{\tt \ at\_time\ }}
\newcommand{\hatell}{\skew6\hat\ell\,}
\newcommand{\Then}{{\bf then}}
\newcommand{\Until}{{\bf until}}
\newcommand{\Else}{{\bf else}}
\newcommand{\Repeat}{{\bf repeat}}
\newcommand{\cA}{{\cal A}}
\newcommand{\cE}{{\cal E}}
\newcommand{\cF}{{\cal F}}
\newcommand{\cI}{{\cal I}}
\newcommand{\cN}{{\cal N}}
\newcommand{\cR}{{\cal R}}
\newcommand{\cS}{{\cal S}}
\newcommand{\BN}{B^{\scriptscriptstyle \cN}}
\newcommand{\BS}{B^{\scriptscriptstyle \cS}}
\newcommand{\cW}{{\cal W}}
\newcommand{\EG}{E_G}
\newcommand{\CG}{C_G}
\newcommand{\CN}{C_\cN}
\newcommand{\ES}{E_\cS}
\newcommand{\EN}{E_\cN}
\newcommand{\CS}{C_\cS}

\newcommand{\attack}{\mbox{{\it attack}}}
\newcommand{\attacking}{\mbox{{\it attacking}}}
\newcommand{\delivered}{\mbox{{\it delivered}}}
\newcommand{\exist}{\mbox{{\it exist}}}
\newcommand{\decide}{\mbox{{\it decide}}}
\newcommand{\clean}{{\it clean}}
\newcommand{\diff}{{\it diff}}
\newcommand{\Failed}{{\it failed}}
\newcommand\eqdef{=_{\rm def}}
\newcommand{\true}{\mbox{{\it true}}}
\newcommand{\false}{\mbox{{\it false}}}

\newcommand{\DN}{D_{\cN}}
\newcommand{\DS}{D_{\cS}}
\newcommand{\tyme}{{\it time}}
\newcommand{\fp}{f}

\newcommand{\Kax}{{\rm K}_n}
\newcommand{\Kaxc}{{\rm K}_n^C}
\newcommand{\Kaxd}{{\rm K}_n^D}
\newcommand{\Tax}{{\rm T}_n}
\newcommand{\Taxc}{{\rm T}_n^C}
\newcommand{\Taxd}{{\rm T}_n^D}
\newcommand{\fourax}{{\rm S4}_n}
\newcommand{\fouraxc}{{\rm S4}_n^C}
\newcommand{\fouraxd}{{\rm S4}_n^D}
\newcommand{\fiveax}{{\rm S5}_n}
\newcommand{\fiveaxc}{{\rm S5}_n^C}
\newcommand{\fiveaxd}{{\rm S5}_n^D}
\newcommand{\Dax}{{\rm KD45}_n}
\newcommand{\Daxc}{{\rm KD45}_n^C}
\newcommand{\Daxd}{{\rm KD45}_n^D}
\newcommand{\LP}{{\cal L}_n}
\newcommand{\LCP}{{\cal L}_n^C}
\newcommand{\LDP}{{\cal L}_n^D}
\newcommand{\LCDP}{{\cal L}_n^{CD}}
\newcommand{\MP}{{\cal M}_n}
\newcommand{\MPr}{{\cal M}_n^r}
\newcommand{\MPrt}{\M_n^{\mbox{\scriptsize{{\it rt}}}}}
\newcommand{\MPrst}{\M_n^{\mbox{\scriptsize{{\it rst}}}}}
\newcommand{\MPelt}{\M_n^{\mbox{\scriptsize{{\it elt}}}}}
\renewcommand{\lang}{\mbox{${\cal L}_{n} (\Phi)$}}
\renewcommand{\langd}{\mbox{${\cal L}_{n}^D (\Phi)$}}
\newcommand{\fiveaxdu}{{\rm S5}_n^{DU}}
\newcommand{\LPD}{{\cal L}_n^D}
\newcommand{\fiveaxu}{{\rm S5}_n^U}
\newcommand{\fiveaxcu}{{\rm S5}_n^{CU}}
\newcommand{\LPU}{{\cal L}^{U}_n}
\newcommand{\LPCU}{{\cal L}_n^{CU}}
\newcommand{\LDPU}{{\cal L}_n^{DU}}
\newcommand{\LCPU}{{\cal L}_n^{CU}}
\newcommand{\LPDU}{{\cal L}_n^{DU}}
\newcommand{\LPCDU}{{\cal L}_n^{\it CDU}}
\newcommand{\Cn}{\C_n}
\newcommand{\CSnp}{\I_n^{oa}(\Phi')}
\newcommand{\CSc}{\C_n^{oa}(\Phi)}
\newcommand{\Ccs}{\C_n^{oa}}
\newcommand{\CSAX}{OA$_{n,\Phi}$}
\newcommand{\CSAXN}{OA$_{n,{\Phi}}'$}
\newcommand{\untill}{U}
\newcommand{\until}{\, U \,}
\newcommand{\amp}{{\rm a.m.p.}}
\newcommand{\commentout}[1]{}
\newcommand{\msgc}[1]{ @ #1 }
\newcommand{\Camp}{{\C_n^{\it amp}}}
\newcommand{\bi}{\begin{itemize}}
\newcommand{\ei}{\end{itemize}}
\newcommand{\be}{\begin{enumerate}}
\newcommand{\ee}{\end{enumerate}}
\newcommand{\rarrowr}{\stackrel{r}{\rightarrow}}
\newcommand{\ack}{\mbox{\it ack}}
\newcommand{\Gz}{\G_0}
\newcommand{\denselist}{\itemsep 0pt\partopsep 0pt}
\def\seealso#1#2{({\em see also\/} #1), #2}
\newcommand{\cents}{\hbox{\rm \rlap{/}c}}

\newenvironment{oldthm}[1]{\par\noindent{\bf Theorem #1:} \em \noindent}{\par}
\newenvironment{oldlem}[1]{\par\noindent{\bf Lemma #1:} \em \noindent}{\par}
\newenvironment{oldcor}[1]{\par\noindent{\bf Corollary #1:} \em \noindent}{\par}
\newenvironment{oldpro}[1]{\par\noindent{\bf Proposition #1:} \em \noindent}{\par}
\newcommand{\othm}[1]{\begin{oldthm}{\ref{#1}}}
\newcommand{\eothm}{\end{oldthm} \medskip}
\newcommand{\olem}[1]{\begin{oldlem}{\ref{#1}}}
\newcommand{\eolem}{\end{oldlem} \medskip}
\newcommand{\ocor}[1]{\begin{oldcor}{\ref{#1}}}
\newcommand{\eocor}{\end{oldcor} \medskip}
\newcommand{\opro}[1]{\begin{oldpro}{\ref{#1}}}
\newcommand{\eopro}{\end{oldpro} \medskip}

\newcommand{\world}{W}
\newcommand{\WN}{\W_N}
\newcommand{\Winf}{\W^*}
\newcommand{\tends}{\rightarrow}
\newcommand{\tendsto}{\tends}
\newcommand{\ninfty}{{N \rightarrow \infty}}
\newcommand{\nworldsv}[1]{{\it \#worlds}^{#1}}
\newcommand{\nworlds}{{{\it \#worlds}}_{N}^{\epsvec}}
\newcommand{\nwrldPnt}[1]{\nworlds[#1]}
\newcommand{\nworldsarg}[1]{\nworlds[#1]}
\newcommand{\binco}[2]{{{#1}\choose{#2}}}
\newcommand{\closure}[1]{{\overline{#1}}}
\newcommand{\balpha}{\bar{\alpha}}
\newcommand{\bbeta}{\bar{\beta}}
\newcommand{\bgamma}{\bar{\gamma}}
\newcommand{\half}{\frac{1}{2}}
\newcommand{\bQ}{\overline{Q}}
\newcommand{\Vector}[1]{{\langle #1 \rangle}}
\newcommand{\Algzeroone}{\mbox{\em Compute01}}
\newcommand{\Algcompute}{\mbox{{\em Compute-Pr}$_\infty$}}

\newcommand{\Prinfv}[1]{{\Pr}^{#1}_\infty}
\newcommand{\PrNv}[1]{{\Pr}^{#1}_N}
\newcommand{\Prinf}{{\Pr}_\infty}
\newcommand{\PrN}{{\Pr}_N}
\newcommand{\pN}{\PrN (\phi | \KB)}
\newcommand{\IPrinf}{{\Box\Diamond\Prinf}}
\newcommand{\PPrinf}{{\Diamond\Box\Prinf}}
\newcommand{\prNw}{{\Pr}_{N}^{\epsvec}}
\newcommand{\prNwi}{{\Pr}_{N^i}^{\epsvec^i}}
\newcommand{\priw}{{\Pr}_{\infty}}
\newcommand{\beliefprob}[1]{\Pr(#1)} 
\newcommand{\Prinfeps}{{{\Pr}_\infty^{\epsvec}}}
\newcommand{\PrNeps}{{{\Pr}_N^{\epsvec}}}
\newcommand{\Pinf}[2]{{\Pi_\infty^{#1}[#2]}}

\newcommand{\infvocab}{\Omega}
\newcommand{\infunvocab}{\Upsilon}
\newcommand{\vocab}{\Phi}
\newcommand{\nonunaryvocab}{\vocab}
\newcommand{\unvocab}{\Psi}

\newcommand{\cL}{{\cal L}}
\newcommand{\cLne}{{\cal L}^-}
\newcommand{\finitelang}{\cL_i^d(\Phi)}
\newcommand{\fLd}{\cL_d^d(\Phi)}
\newcommand{\cLd}{\cL_{\mbox{\em \scriptsize$d$}}(\Phi)}
\newcommand{\Laeq}{{\cal L}^{\aeq}}
\newcommand{\Leq}{{\cal L}^{=}}
\newcommand{\Lunaeq}{\Laeq_1}
\newcommand{\Luneq}{\Leq_1}

\newcommand{\KB}{{\it KB}}
\newcommand{\phieven}{\phi_{\mbox{\footnotesize\it even}}}
\newcommand{\phiodd}{\phi_{\mbox{\footnotesize\it odd}}}
\newcommand{\KBM}{\KB_{\boldM}}
\newcommand{\Pstar}{P^*}
\newcommand{\barP}{\neg P}
\newcommand{\barQ}{\neg Q}
\newcommand{\maxarity}{\rho}
\newcommand{\Init}{\mbox{\it Init\/}}
\newcommand{\Rep}{\mbox{\it Rep\/}}
\newcommand{\Acc}{\mbox{\it Acc\/}}
\newcommand{\Comp}{\mbox{\it Comp\/}}
\newcommand{\Step}{\mbox{\it Step\/}}
\newcommand{\Univ}{\mbox{\it Univ\/}}
\newcommand{\Exis}{\mbox{\it Exis\/}}
\newcommand{\Between}{\mbox{\it Between\/}}
\newcommand{\Count}{\mbox{\it count\/}}
\newcommand{\phiq}{\phi_Q}
\newcommand{\propform}{\beta}
\newcommand{\allphi}{\ast}
\newcommand{\ID}{\mbox{\it ID}}
\newcommand{\atomdesc}{{\psi_*}}
\newcommand{\rigid}{\mbox{\it rigid}}
\newcommand{\abdesc}{\widehat{D}}
\newcommand{\KBtwo}{\theta}
\newcommand{\psits}{\psi[\KBtwo,\abdesc]}
\newcommand{\psitsnohat}{\psi[\KBtwo,D]}
\newcommand{\KBfly}{\KB_{\mbox{\scriptsize \it fly}}}
\newcommand{\KBchirps}{\KB_{\mbox{\scriptsize \it chirps}}}
\newcommand{\KBmagpie}{\KB_{\mbox{\scriptsize \it magpie}}}
\newcommand{\KBhep}{\KB_{\mbox{\scriptsize \it hep}}}
\newcommand{\KBnixon}{\KB_{\mbox{\scriptsize \it Nixon}}}
\newcommand{\KBel}{\KB_{\mbox{\scriptsize \it likes}}}
\newcommand{\KBtax}{\KB_{\mbox{\scriptsize \it taxonomy}}}
\newcommand{\KBarm}{\KB_{\mbox{\scriptsize \it arm}}}
\newcommand{\KBlate}{\KB_{\mbox{\scriptsize \it late}}}
\newcommand{\KBp}{{\KB'}}
\newcommand{\KBflyp}{\KB_{\mbox{\scriptsize \it fly}}'}
\newcommand{\KBpp}{{\KB''}}
\newcommand{\KBdef}{\KB_{\mbox{\scriptsize \it def}}}
\newcommand{\KBdishep}{\KB_{\mbox{\scriptsize \it $\lor$hep}}}

\newcommand{\canKB}{\widehat{\KB}}
\newcommand{\canxi}{\widehat{\xi}}
\newcommand{\KBfo}{\KB_{\it fo}}
\newcommand{\KBconst}{\psi}
\newcommand{\KBprop}{\KBp}

\newcommand{\quak}{\mbox{\it Quaker\/}}
\newcommand{\repub}{\mbox{\it Republican\/}}
\newcommand{\pac}{\mbox{\it Pacifist\/}}
\newcommand{\Nixon}{\mbox{\it Nixon\/}}
\newcommand{\Winged}{\mbox{\it Winged\/}}
\newcommand{\Winner}{\mbox{\it Winner\/}}
\newcommand{\Child}{\mbox{\it Child\/}}
\newcommand{\Boy}{\mbox{\it Boy\/}}
\newcommand{\Tall}{\mbox{\it Tall\/}}
\newcommand{\Elephant}{\mbox{\it Elephant\/}}
\newcommand{\Gray}{\mbox{\it Gray\/}}
\newcommand{\Yellow}{\mbox{\it Yellow\/}}
\newcommand{\Clyde}{\mbox{\it Clyde\/}}
\newcommand{\Tweety}{\mbox{\it Tweety\/}}
\newcommand{\Opus}{\mbox{\it Opus\/}}
\newcommand{\Bird}{\mbox{\it Bird\/}}
\newcommand{\Penguin}{{\it Penguin\/}}
\newcommand{\Fish}{\mbox{\it Fish\/}}
\newcommand{\Fly}{\mbox{\it Fly\/}}
\newcommand{\Warmblooded}{\mbox{\it Warm-blooded\/}}
\newcommand{\White}{\mbox{\it White\/}}
\newcommand{\Red}{\mbox{\it Red\/}}
\newcommand{\Giraffe}{\mbox{\it Giraffe\/}}
\newcommand{\Visible}{\mbox{\it Easy-to-see\/}}
\newcommand{\Bat}{\mbox{\it Bat\/}}
\newcommand{\Blue}{\mbox{\it Blue\/}}
\newcommand{\Fever}{\mbox{\it Fever\/}}
\newcommand{\Jaun}{\mbox{\it Jaun\/}}
\newcommand{\Hep}{\mbox{\it Hep\/}}
\newcommand{\Eric}{\mbox{\it Eric\/}}
\newcommand{\Alice}{\mbox{\it Alice\/}}
\newcommand{\Tom}{\mbox{\it Tom\/}}
\newcommand{\Lottery}{\mbox{\it Lottery\/}}
\newcommand{\Zookeeper}{\mbox{\it Zookeeper\/}}
\newcommand{\Fred}{\mbox{\it Fred\/}}
\newcommand{\Likes}{\mbox{\it Likes\/}}
\newcommand{\Day}{\mbox{\it Day\/}}
\newcommand{\Nextday}{\mbox{\it Next-day\/}}
\newcommand{\Sleepslate}{\mbox{\it To-bed-late\/}}
\newcommand{\Riseslate}{\mbox{\it Rises-late\/}}
\newcommand{\TS}{\mbox{\it TS\/}}
\newcommand{\EEJ}{\mbox{\it EEJ\/}}
\newcommand{\FC}{\mbox{\it FC\/}}
\newcommand{\Dodo}{\mbox{\it Dodo\/}}
\newcommand{\Ab}{\mbox{\it Ab\/}}
\newcommand{\Chirps}{\mbox{\it Chirps\/}}
\newcommand{\Swims}{\mbox{\it Swims\/}}
\newcommand{\Magpie}{\mbox{\it Magpie\/}}
\newcommand{\Moody}{\mbox{\it Moody\/}}
\newcommand{\SomeMorning}{\mbox{\it Tomorrow\/}}
\newcommand{\Animal}{\mbox{\it Animal\/}}
\newcommand{\Sparrow}{\mbox{\it Sparrow\/}}
\newcommand{\Turtle}{\mbox{\it Turtle\/}}
\newcommand{\Older}{\mbox{\it Over60\/}}
\newcommand{\Patient}{\mbox{\it Patient\/}}
\newcommand{\Black}{\mbox{\it Black\/}}
\newcommand{\Ray}{\mbox{\it Ray\/}}
\newcommand{\Reiter}{\mbox{\it Reiter\/}}
\newcommand{\Drew}{\mbox{\it Drew\/}}
\newcommand{\McDermott}{\mbox{\it McDermott\/}}
\newcommand{\Emu}{\mbox{\it Emu\/}}
\newcommand{\Canary}{\mbox{\it Canary\/}}
\newcommand{\BlueCanary}{\mbox{\it BlueCanary\/}}
\newcommand{\FlyingBird}{\mbox{\it FlyingBird\/}}
\newcommand{\UL}{\mbox{\it LeftUsable\/}}
\newcommand{\UR}{\mbox{\it RightUsable\/}}
\newcommand{\BL}{\mbox{\it LeftBroken\/}}
\newcommand{\BR}{\mbox{\it RightBroken\/}}
\newcommand{\Ticket}{\mbox{\it Ticket\/}}
\newcommand{\BlueEyed}{{\mbox{\it BlueEyed\/}}}
\newcommand{\Jaundice}{{\it Jaundice\/}}
\newcommand{\Hepatitis}{{\it Hepatitis\/}}
\newcommand{\HeartDisease}{{\mbox{\it Heart-disease\/}}}
\newcommand{\bJ}{{\overline{J}\,}}
\newcommand{\bH}{{\overline{H}\,}}
\newcommand{\bB}{{\overline{B}\,}}
\newcommand{\Prem}{{\mbox{\it Child\/}}}
\newcommand{\David}{{\mbox{\it David\/}}}
\newcommand{\Son}{{\mbox{\it Son\/}}}	

\newcommand{\ceslim}{Ces\`{a}ro limit}
\newcommand{\Sigmad}{\Sigma^d_i}
\newcommand{\Liogonkii}{Liogon'ki\u\i}
\newcommand{\Vstar}{{\modfrag_*}}
\newcommand{\moddesc}{\psi \land \modfrag}
\newcommand{\sumact}{\Degr_2}
\newcommand{\degree}{\Degr_1}
\newcommand{\Active}{\alpha}
\newcommand{\ActiveAtoms}{\boldA}
\newcommand{\named}{n}
\newcommand{\aactive}{a}
\newcommand{\degr}{\delta}
\newcommand{\chsize}{f}
\newcommand{\chconst}{g}
\newcommand{\Chconst}{G}
\newcommand{\Degr}{\Delta}
\newcommand{\frags}{\M}
\newcommand{\const}{H}
\newcommand{\modfrag}{\V}
\newcommand{\AD}{\A}
\newcommand{\Named}{\nu}
\newcommand{\bit}{b}
\newcommand{\guess}{\gamma}
\newcommand{\bxor}[1]{\dot{\bor}}
\newcommand{\weight}{\omega}
\newcommand{\arity}{{\it arity}}
\newcommand{\assigned}{\leftarrow}
\newcommand{\snum}[2]{{{#1} \brace {#2}}}

\newcommand{\eps}{\tau}
\newcommand{\vareps}{\varepsilon}
\newcommand{\varepsvec}{{\vec{\vareps}}}
\newcommand{\xtuple}{\vec{x}}
\newcommand{\ctuple}{{\vec{c}\,}}
\newcommand{\uvec}{{\!{\vec{\,u}}}}
\newcommand{\ucom}{u}
\newcommand{\pvec}{{\!{\vec{\,p}}}}
\newcommand{\pcom}{p}
\newcommand{\vvec}{\vec{v}}
\newcommand{\wvec}{\vec{w}}
\newcommand{\xvec}{\vec{x}}
\newcommand{\yvec}{\vec{y}}
\newcommand{\zvec}{\vec{z}}
\newcommand{\zerovec}{\vec{0}}

\newcommand{\epscom}{\eps}
\newcommand{\epsvec}{{\vec{\eps}\/}}
\newcommand{\prop}[2]{{||{#1}||_{{#2}}}}
\newcommand{\aeq}{\approx} 
\newcommand{\app}{\approx}
\newcommand{\alt}{\prec}
\newcommand{\aleq}{\preceq}
\newcommand{\agt}{\succ}
\newcommand{\ageq}{\succeq}
\newcommand{\altne}{\prec}
\newcommand{\naeq}{\not\approx}
\newcommand{\cprop}[3]{{\|{#1}|{#2}\|_{{#3}}}}
\newcommand{\Bigcprop}[3]{{\Bigl\|{#1}\Bigm|{#2}\Bigr\|_{{#3}}}}

\newcommand{\reals}{\IR}
\newcommand{\qsep}{\,}
\newcommand{\perm}{\pi}
\newcommand{\val}{V}
\newcommand{\rwent}{\mbox{$\;|\!\!\!\sim$}_{\mbox{\scriptsize \it rw}}\;}
\newcommand{\notrwent}{\mbox{$\;|\!\!\!\not\sim$}_{\mbox{\scriptsize\it rw}}\;}
\newcommand{\dempster}{\delta}
\newcommand{\dentails}{{\;|\!\!\!\sim\;}}
\newcommand{\notdentails}{{\;|\!\!\!\not\sim\;}}
\newcommand{\dentailssub}[1]{\dentails\hspace{-0.4em}_{
          \mbox{\scriptsize{\it #1}}}\;}
\newcommand{\notdentailssub}[1]{\notdentails\hspace{-0.4em}_{
          \mbox{\scriptsize{\it #1}}}\;}
\newcommand{\default}{\rightarrow}

\newcommand{\vecof}[1]{{\pi({#1})}}
\newcommand{\PIN}[2]{{\Pi_N^{#1}[#2]}}
\newcommand{\SS}[2]{{S^{#1}[#2]}}
\newcommand{\SSc}[2]{{S^{#1}[#2]}}
\newcommand{\SSzero}[1]{\SS{\zerovec}{#1}}
\newcommand{\SSczero}[1]{\SSc{\zerovec}{#1}}
\newcommand{\SSpos}[1]{\SS{\leq \zerovec}{#1}}
\newcommand{\Sol}{{\it Sol}}
\newcommand{\poscon}{\gamma}
\newcommand{\constraints}{\Gamma}
\newcommand{\constpos}{\constraints^{\leq}}
\newcommand{\propspace}{\Delta^K}
\newcommand{\mept}{\vvec}
\newcommand{\mecoord}{v}
\newcommand{\mepts}{{\Q}}
\newcommand{\OS}{{\cal O}}
\renewcommand{\S}{{\cal S}}
\newcommand{\meval}{{\rho}}
\newcommand{\meptmin}{{\uvec^{\ast}_{\mbox{\scriptsize min}}}}
\newcommand{\sizeof}[1]{{\sigma(#1)}}
\newcommand{\mesize}{{\sigma^{\ast}}}
\renewcommand{\pf}{\alpha}
\newcommand{\ps}{\beta}
\newcommand{\Atoms}{\AD}
\newcommand{\limNstar}{{\lim_{N \rightarrow \infty}}^{\!\!\!*}\:}
\newcommand{\probf}[2]{F_{[#1|#2]}}
\newcommand{\probfun}[1]{F_{[#1]}}
\newcommand{\Prmu}[1]{{\mu}_{#1}}
\newcommand{\foversion}[1]{\xi_{#1}}
\newcommand{\fly}{\mbox{\it fly\/}}
\newcommand{\bird}{\mbox{\it bird\/}}
\newcommand{\yellow}{\mbox{\it yellow\/}}
\newcommand{\propconsts}{\Lambda}
\newcommand{\alldiff}{\chi^{\neq}}
\newcommand{\unaryD}{D^1}
\newcommand{\nonunD}{D^{> 1}}
\newcommand{\eqD}{D^{=}}

\maketitle  
\begin{abstract}
As examples such as the Monty Hall puzzle show,
applying conditioning to update a probability distribution 
on
a ``naive space'', which does not take into account the protocol used,
can often lead to counterintuitive results.  Here we
examine why.  
A criterion known as CAR
(``coarsening at random'') in the statistical literature 
characterizes when ``naive'' conditioning in a naive space works.  
We show that the CAR condition holds
rather infrequently, and we provide 
a procedural characterization of it, by giving a
randomized algorithm 
that generates all and
only 
distributions for which CAR holds. This substantially extends previous
characterizations of CAR.
We also
consider more generalized
notions of update such as Jeffrey conditioning and minimizing relative
entropy (MRE).  We give a generalization of the CAR condition that
characterizes when Jeffrey conditioning leads to appropriate answers,
and show that there exist some very simple settings in which MRE
essentially never gives the right results.
This generalizes and
interconnects previous results obtained in the literature on CAR and
MRE.
\end{abstract}
\section{Introduction}\label{sec:intro}
Suppose an agent represents her uncertainty about a domain 
using a
probability distribution. At some point, she receives some new
information about the domain. How should she update her distribution in
the light of this information? {\em Conditioning\/} is by far the most
common
method in case the information comes in the form of an
event.  However, there are numerous well-known examples showing
that naive conditioning can lead to problems.  We give just two of them
here.

\xam\label{xam:Monty} The {\em Monty Hall puzzle}
\cite{Mosteller,vScomb}:  Suppose 
that
you're on a game
show and given
a choice of three doors.  Behind one is a car; behind the others are
goats.  You pick door  1.  Before opening door 1, Monty
Hall, the host (who
knows what is behind each door) opens door  3, which has a goat.
He then asks you if you still want to take what's behind door  1,
or to take what's behind door 2 instead.  Should you switch?
Assuming that, initially, the car was equally likely to be behind each
of the doors, naive conditioning suggests that, given that it is not
behind door 3, it is equally likely to be behind door 1 and door 2.  Thus,
there is no reason to switch.  However, another argument suggests you
should switch: if a goat is behind door 1 (which happens with
probability 2/3), switching helps; if a car is behind door 1 (which
happens with probability 1/3), switching hurts.  
Which argument is right?
\exam

\xam\label{xam:3pris} The {\em three-prisoners puzzle}
\cite{BF82,Gardner61,Mosteller}:
Of three prisoners $a$, $b$, and $c$, two are to be executed,
but $a$ does not know which. 
Thus,
$a$ thinks that the probability that $i$
will be executed is $2/3$ for $i \in \{a,b,c\}$.
He says to the jailer,
``Since either $b$ or $c$ is certainly going to be executed,
you will give me no information about my own chances if you give me the
name of one man, either $b$ or $c$, who is going to be executed.''
But then, 
no matter what the jailer says, naive
conditioning leads $a$ to believe that his chance of execution went
down from $2/3$ to $1/2$.
\exam

There are numerous other well-known examples where naive conditioning gives 
what seems to be an inappropriate answer, including the {\em
two-children puzzle}
\cite{Gardner82,vScombb,vScombc}
and the {\em second-ace puzzle\/} \cite{Freund65,Shafer85,HT}.%
\footnote{Both the Monty Hall  
puzzle
and the two-children
puzzle were discussed in {\em Ask Marilyn}, 
Marilyn vos Savant's weekly column in 
``Parade   Magazine''. 
Of all {\em Ask Marilyn\/} columns ever published, they
  reportedly \cite{vScombc} generated respectively the
 most and the second-most response.} 

Why does naive conditioning give the wrong answer in such examples?
As argued in \cite{HT,Shafer85}, the real problem is that we are not
conditioning in the right space.  If we work in a larger
``sophisticated'' space, where we
take the protocol used by Monty (in Example~\ref{xam:Monty}) and the
jailer (in Example~\ref{xam:3pris}) into account, conditioning does
deliver the right answer.  Roughly speaking, the sophisticated space
consists
of all the possible sequences of events that could happen (for example,
what Monty would say in each circumstance, or what the jailer would say
in each circumstance), with their probability.%
\footnote{The notions of ``naive space'' and ``sophisticated space'' will be
formalized in Section~\ref{sec:naive}.  This introduction is meant only to
give
an intuitive feel for the issues.}
However, working in the sophisticated space has
problems too.
For one thing, it is not always clear what the relevant probabilities in
the sophisticated space are.  For example, what is the probability that the
jailer says $b$ if 
$b$ and $c$ are
to be executed?  Indeed, in some cases, it is
not even clear what the elements of the larger space are.
Moreover, even when the elements and the
relevant probabilities are known, the size of the sophisticated space may
become an issue, as the following example shows.

\xam\label{pch4.xam1} Suppose that a world describes which of 100 people
have a certain disease.  A world can be characterized by a tuple
of 100 0s and 1s, where the $i$th component is 1 iff individual $i$
has the disease.   There are $2^{100}$ possible worlds.
Further suppose that the
``agent'' in question is a computer system.
Initially, the agent has no information, and considers all $2^{100}$
worlds equally likely.  The agent then
receives information that is assumed to be true about which world is the
actual world.  This information comes in the form of statements like
``individual $i$ is sick or individual $j$ is healthy''
or ``at least 7 people have the disease''.  Each such statement can be
identified with a set of possible worlds.  For example, the statement
``at least 7 people have the disease'' can be identified with the set of
tuples with at least 7 1s.  For simplicity, assume that the
agent is given information saying
``the actual world is in set $U$'', for various sets $U$.
Suppose at some point the agent has been told that the actual world is in
$U_1, \ldots, U_n$.  Then, after doing conditioning, the agent
has a uniform probability on $U_1 \inter \ldots \inter U_n$.

But how does the agent keep track of the
worlds it considers possible? It certainly will not explicitly list
them; there are simply too many.  One possibility is that it keeps track of
what it has been told; the possible worlds are then
the ones consistent with what it has been told.  But this leads to two
obvious problems: checking for consistency with what it has been told
may be hard, and if it has been told $n$ things for large $n$,
remembering them all may be infeasible.  In situations where these two
problems arise, an agent may not be able to condition appropriately.
\exam
Example~\ref{pch4.xam1} provides some motivation for working in the
smaller, more naive space.
Examples~\ref{xam:Monty} and~\ref{xam:3pris}
show that this is not always appropriate.  Thus, an obvious question is
when it is appropriate. 
It turns out that this question is highly relevant in the statistical
areas of {\em selectively reported data\/} and {\em missing data}.  Originally
studied within these contexts \cite{Rubin76,DawidD77}, it was later
found that it also plays a fundamental role in the statistical work on
{\em survival analysis\/} \cite{Kleinbaum99}. 
Building on previous approaches, 
Heitjan and Rubin \citeyear{HeitjanR91}
presented a necessary and sufficient condition for when
conditioning in the ``naive space'' is appropriate. Nowadays this so-called {\em
CAR (Coarsening at Random)\/} condition is an established tool in
survival analysis. 
(See \cite{GillLR97,Nielsen98} for overviews.)
We examine this criterion
in our own, rather different context, and show that it applies
rather rarely.
Specifically, we show that there are realistic settings where the
sample space is structured in such a way that it  is
impossible to satisfy CAR, and we provide a criterion to help
determine whether or not this is the case. We also give a {\em
  procedural\/} characterization of the CAR condition,
by giving a randomized algorithm that generates all and
only distributions for which CAR holds,
thereby solving an open 
problem posed in \cite{GillLR97}.
We then show that the
situation is 
worse if the information does not come in the form
of an event.  For that case, several generalizations of conditioning
have been proposed.  Perhaps the best known are {\em  Jeffrey
conditioning\/}
\cite{Jeffrey68} (also known as {\em Jeffrey's rule\/})
and {\em Minimum Relative Entropy (MRE) Updating\/}
\cite{Kullback59,Csiszar75,ShoreJohnson} (also known as  {\em
  cross-entropy}).
Jeffrey conditioning is a generalization of ordinary conditioning; MRE
updating is a generalization of Jeffrey conditioning.

We show that
Jeffrey conditioning, when applicable, can be justified under
an appropriate generalization of the CAR condition.   
Although it has been argued,
using mostly axiomatic characterizations, that MRE updating (and hence also
Jeffrey conditioning) is, when applicable, the {\em only\/}
reasonable way to update probability (see, e.g.,
\cite{Csiszar91,ShoreJohnson}),
it is well known that there are
situations where applying MRE leads to paradoxical, highly counterintuitive
results
\cite{Hunter89,Seidenfeld86,vF4}.
\xam Consider the {\em Judy Benjamin\/} problem \cite{vF4}:
Judy is lost in a region that is divided into
two halves, Blue and Red territory, each of which is further
divided into Headquarters Company area and Second Company
area.  A priori, Judy considers it equally likely that she is in any of
these four
quadrants.   She contacts her own headquarters by radio, and is told
``I can't be sure where you are.  If you are in
Red territory, the odds are 3:1 that you are in HQ Company area ...''
At this point the radio gives out.
MRE updating on this information leads to a distribution where the
posterior probability of being in Blue territory is greater than $1/2$.
Indeed, if HQ had said ``If you are in Red
territory, the odds are $\alpha:1$ that you are in HQ company area
\ldots'', then 
for all $\alpha \ne 1$,
according to MRE updating,
the posterior probability of being in Blue territory
is always greater than 
$1/2$.
\exam

In \cite{GroveHalpern97}, a ``sophisticated 
space'' is
provided where conditioning gives what is arguably the more intuitive
answer in the Judy Benjamin problem, namely that if HQ sends 
a
message of
the form  
``if you are in Red territory, then
the odds are $\alpha:1$ that you are in HQ company area''
then Judy's
posterior probability of being in each of the two quadrants in Blue
remains at $1/4$.  Seidenfeld
\citeyear{Seidenfeld86}, strengthening results of Friedman and Shimony
\citeyear{FriedmanS71}, showed that there is {\em no\/} sophisticated
space in which
conditioning will give the same answer as MRE in this case.  (See also
\cite{Dawid01} for similar results along these lines.)  We strengthen
these results by showing that, 
even in a class of much simpler situations 
(where Jeffrey conditioning cannot be applied), 
using MRE in the naive space corresponds
to conditioning in the sophisticated space in essentially only trivial
cases. These results taken together show that {\em generally speaking, working with the naive
space, while an attractive approach, is likely to give highly
misleading answers}. That is the main message of this paper.
We remark that, although there are certain similarities, our results are
quite different in spirit from the well-known results of Diaconis and
Zabell \citeyear{DZ}.  They considered when a posterior probability
could be viewed as the result of conditioning a prior
probability on some larger space.  
By way of contrast,
we have a fixed larger space in mind
(the ``sophisticated space''),
and are interested in when conditioning in the naive space and the
sophisticated space agree.

It is also worth stressing that the distinction between the naive and the
sophisticated space is entirely unrelated to the philosophical view that
one has of probability and how one should do probabilistic inference.
For example, the probabilities in the Monty Hall puzzle can be
viewed as the participant's subjective probabilities about the location
of the car and about what Monty will say under what circumstances;
alternatively, they can be viewed as ``frequentist'' probabilities,
inferred from watching the Monty Hall show  on television for many weeks
and then setting the probabilities equal to observed frequencies. The
problem we address occurs both from a frequentist and from a subjective
stance.

The rest of this paper is organized as follows.  In
Section~\ref{sec:naive} we formalize the notion of naive and
sophisticated spaces.  In Section~\ref{sec:CAR}, we consider the case
where the information comes in the form of an event.  We describe the
CAR condition and show 
that it is violated in a general setting of which 
the Monty Hall and three-prisoners puzzle are
special cases. In Section~\ref{sec:char} we give several
characterizations of CAR. We supply conditions under which it is
guaranteed to hold and guaranteed not to hold, and we give a
randomized algorithm that generates all and only distributions for
which CAR holds.
In
Section~\ref{sec:MRE} we consider the case where the information is not
in the form of an event.  %
We first consider situations where Jeffrey conditioning can be
applied. We
show that Jeffrey conditioning in the naive space gives the appropriate
answer iff a generalized CAR condition holds.  We then show that,
typically,
applying MRE in the naive space does 
not
give the
appropriate answer.  
We conclude with some discussion of the implication
of these results in Section~\ref{sec:discussion}.
\section{Naive vs.~Sophisticated Spaces}\label{sec:naive}

Our formal model is a special case of the multi-agent systems framework
\cite{HFfull}, which is essentially the same as that used in \cite{FrH1full}
to model belief revision.  We assume that there is some external world
in a set $W$, and an agent who makes observations or gets information
about that world.  We can describe the situation by a pair $(w,l)$,
where $w \in W$ is the actual world, and $l$ is the agent's {\em local
state\/}, which essentially characterizes her information.  $W$ is
what we called the ``naive space'' in the introduction.
For the
purposes of this paper, we assume that $l$ has the form $\<o_1, \ldots,
o_n\>$, where $o_j$ is the observation that the agent makes at time $j$,
$j = 1, \ldots, n$.  
This representation implicitly assumes that the
agent remembers
everything she has observed (since her local state encodes all the
previous observations).  Thus, we ignore memory issues here.  We 
also ignore computational issues, just so as to be able to focus on when
conditioning is appropriate.

A pair $(w,\<o_1, \ldots, o_n\>)$ is
called a 
run. A run may be viewed as a 
complete description of what happens over time in
one possible execution of the system.
For simplicity, in this paper, we assume that the state of the world
does not change over time.
\commentout{
We restrict 
to runs $r$ that satisfy the following constraints:
\begin{itemize}
\item $r_W(m) = r_W(0)$ for all $m$ (so the external world does not
change over time);
\item $r_O(0) = \<\,\>$;
\item $r_O(m+1)$ has the form $r_O(m) \cdot o$ for some $o \in O$ (so
that the agent's local state at time $m+1$ is the result of extending
her observation at time $m$ by one observation).
\end{itemize}
}
The ``sophisticated space'' is the set of all possible runs.

In the Monty Hall puzzle, the
naive space has three worlds, representing the three possible locations
of the car.  The sophisticated space describes what Monty would have said in
all circumstances (i.e., Monty's {\em protocol\/}) as well as where
the car is.  The three-prisoners puzzle is treated in 
detail in Example~\ref{xam:3prisb} below.
While in these cases the sophisticated
space is
still relatively simple, this is no longer the case for the Judy
Benjamin puzzle.  Although the naive space has only four elements,
constructing the sophisticated space involves considering all the things
that HQ could have said, which is far from clear, and the conditions
under which HQ says any particular thing.
Grove and Halpern \citeyear{GroveHalpern97} discuss the 
difficulties in constructing such a sophisticated space.
In general, not only is it not clear what the sophisticated space is,
but the need for a sophisticated space and the form it must take
may become clear only after the fact.  For example, in the Judy Benjamin
problem, before contacting headquarters, Judy would almost certainly not
have had a sophisticated space in mind (even assuming she was an expert
in probability), and could not have known the form it would 
have to take until after hearing headquarter's response.

In any case, if the agent has a prior probability on the set $\R$ of 
possible
runs in
the sophisticated space, after hearing or observing $\<o_1, \ldots,
o_k\>$, she can condition, to get a posterior on $\R$.  Formally,
the agent is conditioning her prior on the set 
of runs where her local state at time $k$ is $\<o_1, \ldots,
o_k\>$.

Clearly the agent's probability $\Pr$ on $\R$ induces a probability
$\Pr_W$ 
on $W$
by marginalization.  We are interested in whether the agent can
compute her posterior on $W$ after observing $\<o_1, \ldots, o_k\>$ in a
relatively simple way, without having to work in the sophisticated space.
\xam
\label{xam:3prisb}
Consider the three-prisoners puzzle in more detail. Here
the naive space is $W = \{w_a, w_b, w_c\}$, where $w_x$ is the world
where $x$ is not executed. We are only interested in runs of length 1,
so  $n =1$. 
The set $O$ of observations (what agent can be told) is $\{
\{w_a,w_b\},\{w_a,w_c\} \}$. Here ``$\{w_a,w_b\}$''
corresponds to the observation that either $a$ or $b$ will {\em
  not\/} be executed (i.e., the jailer saying ``$c$ will be executed'');
similarly, $\{w_a,w_c\}$ corresponds to the jailer saying ``$b$ will be
executed''. 
The sophisticated space consists of the four runs 
$$
(w_a,\<\{w_a,w_b\} \>),
(w_a,\<\{w_a,w_c\} \>),
(w_b,\<\{w_a,w_b\} \>),
(w_c,\<\{w_a,w_c\} \>).
$$
Note that there is no run with observation $\<\{w_b,w_c\}\>$, since the
jailer will not tell $a$ that he will be executed.

According to the story, the prior $\Pr_W$ in the naive space has
$\Pr_W(w) = 1/3$ for $w \in W$. The full distribution $\Pr$ on the runs is not
completely specified by the story. 
In particular, we are not told the probability with which the jailer
will say $b$ and $c$ if $a$ will not be executed.  
We return to this point in Example~\ref{xam:3prisc}.
\exam

\section{The CAR Condition}\label{sec:CAR}
A particularly simple setting is where the agent observes or learns that
the external world is in some set $U \subseteq W$.  
For simplicity, we assume 
throughout this paper
that the agent makes only one observation, and
makes it at the first step of the run.
Thus, the set $O$ of possible observations consists of 
nonempty
subsets of $W$.
Thus, any run $r$ can be written as $r = (w,\< U\>)$ where  $w$ is the
actual world and $U$ is a
nonempty subset of $W$.
However, $O$ does not necessarily consist of {\em all\/} the nonempty
subsets of $W$. Some subsets may
never be 
observed.  For example, in Example~\ref{xam:3prisb}, $a$ is never told
that he will 
be executed, so $\{w_b,w_c\}$ is not observed.
We assume that the agent's
observations are accurate, in that if the agent observes $U$ in a run
$r$, then the actual world in $r$ 
is in $U$. That is, we assume that all runs are of the form $r = (w,\<
U \>)$ where $w \in U$.
In Example~\ref{xam:3prisb}, accuracy is enforced by the requirement
that
runs have the form 
$(w_x,\<\{w_x,w_y\} \>)$.

The observation or information obtained does not have to be exactly of
the form ``the actual world is in $U$''.  It suffices that it is
equivalent to such a statement.  This is the case in both the
Monty Hall puzzle and the three-prisoners puzzle.  For
example, in the three-prisoners puzzle, being told that $b$ will be
executed is essentially equivalent to observing $\{w_a,w_c\}$ (either $a$
or $c$ will not be executed).  

In this setting, we can ask whether, after observing $U$, the agent can
compute her posterior on $W$ by conditioning on $U$.  Roughly speaking,
this amounts to asking whether observing $U$ is the same as discovering
that $U$ is true.  This may not be the case in 
general---observing or being told $U$
may carry more information than just the fact that $U$ is
true.   For example, if for some reason $a$ knows that the jailer would
never say $c$ if he could help it (so that, in particular, if $b$ and
$c$ will be executed, then he will definitely say $b$), then hearing $c$
(i.e., observing $\{w_a, w_b\}$) tells $a$ much more than the fact that
the true world is one of $w_a$ or $w_b$.  It says that the true world
must be $w_b$ (for if the true world were $w_a$, the jailer would have
said $b$).

In
the remainder of this paper we 
assume
that $W$ 
is finite. For every scenario we consider we define a 
set of possible observations $O$, consisting of nonempty subsets of
$W$. For given $W$ and $O$, 
the set of runs $\R$ is then defined to be the set $$
\R = \{ (w,\< U \> \mid  \mbox{\ $U \in O, w \in U$}\}.
$$
Given our assumptions that the state does not change over time and  that
the agent makes only one observation,
the set $\R$ of runs can be viewed as a subset of $W \times O$.
While just taking $\R$ to be a subset of $W \times O$ would slightly
simplify the presentation here, in general, we certainly want to allow
sequences of observations.  (Consider, for
example, an $n$-door version of the Monty Hall problem, where Monty
opens a sequence of doors.)  This framework extends naturally to that
setting.  

Whenever we speak of a distribution $\Pr$ 
on
$\R$, we 
implicitly assume that 
the probability of any set on which we condition is strictly greater than 0.
Let $X_W$ and $X_O$ be two random
variables on $\R$, where $X_W$ is the actual world and $X_O$ is the
observed event.  Thus, for $r = (w,\<U\>)$, $X_W(r) = w$ and $X_O(r) = U$.
Given a distribution $\Pr$ on runs $\R$, 
we denote by $\Pr_W$ the marginal distribution of $X_W$, and by
$\Pr_O$ the marginal distribution of $X_O$. For example, for $V,U
\subseteq W$, $\Pr_W(V)$ is short for $\Pr(X_W \in V)$ and $\Pr_W(V
\mid U)$ is short for $\Pr(X_W \in V \mid X_W \in U)$.

Let $\Pr$ be a prior on $\R$ and let $\Pr' = \Pr(\cdot \mid X_O = U)$
be the posterior after observing $U$. The main question we ask in this
paper is under what conditions we have
\begin{equation}
\label{eq:finalizing}
\mbox{$\Pr$}'_W(V) = \mbox{$\Pr$}_W(V|U)
\end{equation} 
for all $V \subseteq W$.
That is, we want to know under what conditions the posterior
$W$ induced by $\Pr'$ can be computed from the prior on $W$ by
conditioning on the observation. (Example~\ref{xam:3prisc} below gives a concrete case.)
We stress that $\Pr$ and $\Pr'$ are distributions on $\R$, while $\Pr_W$ and
$\Pr_W'$ are distributions on $W$ (obtained by marginalization from $\Pr$ and $\Pr'$,
respectively). Note that (\ref{eq:finalizing}) is equivalently stated
as
\begin{equation}
\label{eq:finalizingb}
\Pr(X_W = w \mid X_O = U) = \Pr(X_W = w \mid X_W \in U) \mbox{\ \ for all $w
\in U$}. 
\end{equation}  
(\ref{eq:finalizing}) (equivalently, (\ref{eq:finalizingb})) is called
the ``CAR condition''. It can be characterized as follows:
\commentout{The CAR condition characterizes 
Let $\R[U]$ consist of all runs in $\R$ where the true
world is in $U$ (i.e., $r_W(0) \in U$).  As before, let $\R[\<U\>]$
consist of all runs where the agent observes $U$ at the first step.
Let $\Pr$ be a prior
on $\R$ and let $\Pr' = \Pr(\cdot \mid \R[\<U\>])$ be the posterior
after observing $U$.  
Thus, we are interested in knowing whether
$\Pr'_W(V) = \Pr_W(V \mid U)$; that is, whether the posterior on $W$ induced
by $\Pr'$ can be computed from the prior on $W$ by conditioning on the
observation.  (Example~\ref{xam:3prisc} below gives a concrete case.)
We stress that $\Pr$ and $\Pr'$ are distributions on $\R$, while $\Pr_W$ and
$\Pr_W'$ are distributions on $W$ (obtained by marginalization from $\Pr$ and $\Pr'$,
respectively).

The following simple proposition, 
whose proof we leave to the reader,
says that this can be done iff
conditioning on $U$ is equivalent to conditioning on observing $U$.
\pro\label{pro:conditioningOK} Let $\Pr' = \Pr(\cdot \mid 
\R[\<U\>])$.
Then $\Pr_W' = \Pr_W(\cdot \mid U)$ iff
$\Pr(\R[V] \mid \R[U]) = \Pr(\R[V] \mid \R[\<U\>])$ for all $V \subseteq
W$. \epro

Now the obvious question is when $\Pr(\R[V] \mid \R[U]) = \Pr(\R[V] \mid
\R[\< U\>])$.  
The CAR condition characterizes this.  It is best stated
in terms of random variables.  

(that is, $\R[\<U \>] = \{ r: X_O(r) = U\})$).
}
\thm\label{thm:CAR} \mbox{\cite{GillLR97}}
Fix 
a probability $\Pr$ on $\R$ and a set
$U \subseteq W$.
The following 
are equivalent:
\begin{itemize}
\item[(a)] 
If $\Pr(X_O = U) > 0$, then 
$\Pr(X_W =w \mid X_O = U) = \Pr(X_W = w \mid X_W \in U)$ for all
$w \in U$.
\item[(b)]
The event
$X_W = w$ is independent of 
the event
$X_O = U$ 
given $X_W \in U$, 
for all $w \in U$.
\item[(c)] $\Pr(X_O = U \mid X_W = w) = \Pr(X_O = U \mid X_W \in U)$ for all
$w
\in U$ such that $\Pr(X_W = w) > 0$.
\item[(d)] $\Pr(X_O = U \mid X_W = w) = \Pr(X_O = U \mid X_W = w')$ for all
$w, w' \in U$ such that $\Pr(X_W = w) > 0$ and $\Pr(X_W = w') > 0$.
\end{itemize}
\ethm
For completeness (and because it is useful for our later 
Theorem~\ref{thm:JeffreyOK}), we
provide a proof of Theorem~\ref{thm:CAR}
in the appendix.

The first condition in Theorem~\ref{thm:CAR} 
is just (\ref{eq:finalizingb}).
The third and fourth
conditions justify the name ``coarsening at random''.  Intuitively,
first some world $w \in W$ is realized, and then some ``coarsening
mechanism''
decides which event $U \subseteq W$ such that $w \in U$ is revealed to
the agent.  The event $U$ is called a ``coarsening'' of $w$.  The third
and fourth conditions effectively say that the probability that $w$ is
coarsened to $U$
is the same for all $w \in U$.  This means that the ``coarsening
mechanism'' is such that the probability of observing $U$ is not
affected by the specific value of $w \in U$ that was realized.
In the remainder of this paper, when we say ``$\Pr$ satisfies CAR'', we
mean that $\Pr$ satisfies condition (a) of Theorem~\ref{thm:CAR} (or,
equivalently, any of the other three conditions) {\em for all $U \in
  O$}. Thus, ``$\Pr$ satisfies CAR'' means that conditioning in the naive
space $W$ coincides with 
conditioning in the sophisticated space $\R$ with probability
$1$. 
The CAR condition explains why conditioning in the naive space is not
appropriate in the Monty Hall puzzle or the three-prisoners puzzle.
We consider the three-prisoners
puzzle in detail; a similar analysis applies to Monty Hall.
\xam
\label{xam:3prisc}
In the three-prisoners puzzle,
what is $a$'s prior distribution $\Pr$ on $\R$?  In
Example~\ref{xam:3prisb} we assumed that the marginal distribution $\Pr_W$
on $W$ is uniform. 
Apart from this, $\Pr$ is unspecified. Now suppose that $a$ observes
$\{w_a,w_c\}$ (``the jailer says $b$'').  Naive conditioning would 
lead $a$ to adopt the distribution $\Pr_W(\cdot \mid  \{w_a,w_c\})$. This
distribution satisfies $\Pr_W(w_a \mid  \{w_a,w_c\}) = 1/2$. Sophisticated
conditioning leads $a$ to adopt the distribution $\Pr' = \Pr(\cdot \mid 
X_O = \{w_a,w_c\})$.
By part (d) of Theorem~\ref{thm:CAR}, naive
conditioning is appropriate 
(i.e., $\Pr'_W = \Pr_W(\cdot \mid \{w_a,w_c\})$) only
if the jailer is equally likely to say $b$ in both worlds $w_a$ and $w_c$.  
Since the jailer must say that $b$ will be executed in world $w_c$, it
follows that $\Pr(X_O = \{w_a,w_c\} \mid  X_W = w_c) = 1$.
Thus, conditioning is
appropriate only if the jailer's protocol is such that he definitely
says $b$ in $w_a$, i.e., even if both $b$ and $c$ are executed. But if
this is the case, when the jailer says $c$, conditioning $\Pr_W$ on
$\{w_a,w_b\}$ is {\em not\/} appropriate, since then $a$ knows that he
will be executed.  The world cannot be $w_a$, for then the jailer
would have said $b$.  
Therefore, {\em no matter what the jailer's protocol is}, conditioning
in the naive space cannot coincide with conditioning in the
sophisticated space for both of his responses.
\exam 
The following example shows that in general, in settings of the type arising in
the Monty Hall and the three-prisoners puzzle, the CAR condition can
only be satisfied in very special cases:
\xam\label{xam:CARhard1} Suppose that 
$O = \{U_1, U_2\}$, 
and both $U_1$ and $U_2$ are observed with positive probability.
(This is the case for both Monty Hall and the
three-prisoners puzzle.)  Then the CAR condition
(Theorem~\ref{thm:CAR}(c)) cannot hold for both
$U_1$ and $U_2$ 
unless $\Pr(X_W \in U_1 \inter U_2)$ is either 0 or 1.
For suppose that $\Pr(X_O = U_1) > 0$, $\Pr(X_O = U_2) > 0$, and $0 <
\Pr(X_W \in U_1 \inter U_2) < 1$.  Without loss of generality, there is some 
$w_1 \in U_1 - U_2$ and $w_2 \in U_1
\inter U_2$ such that $\Pr(X_W = w_1) > 0$ and $\Pr(X_W = w_2) > 0$.
Since observations are accurate, we must have $\Pr(X_O = U_1
\mid X_W = w_1) = 1$.  If CAR holds for $U_1$, then we must have
$\Pr(X_O = U_1 \mid X_W = w_2) = 1$.  But then
$\Pr(X_O = U_2 \mid X_W = w_2) = 0$.  But since $\Pr(X_O = U_2) > 0$, it
follows that there is some $w_3 \in U_2$ such that $\Pr(X_W = w_3) > 0$ and
$\Pr(X_O = U_2 \mid X_W = w_3) > 0$.  This contradicts the CAR condition.
\exam
So when does CAR hold? The previous example exhibited a
combination of $O$ and $W$ for which 
CAR can only be satisfied in ``degenerate'' cases. 
In the next section, we shall study this question for arbitrary
combinations of $O$ and $W$. 

\section{Characterizing CAR}
\label{sec:char}
In this section, we provide some characterizations of when the CAR
condition holds, for finite $O$ and $W$.  Our results extend earlier results
of Gill, van der Laan, and Robins
\citeyear{GillLR97}. We first exhibit a simple situation in which CAR is
guaranteed to hold, and we show that this is the only situation in
which it is guaranteed to hold. We then show that, for arbitrary  $O$ and
$W$, 
we can construct a 0-1--valued matrix 
from which a strong necessary condition for CAR to hold can be obtained.
It turns out that, in 
some
cases of interest, CAR is (roughly
speaking) guaranteed {\em not\/} to hold except in ``degenerate'' 
situations.
Finally, we introduce a new ``procedural'' characterization of CAR: we
provide a mechanism such that a distribution $\Pr$ can be thought of
as arising from the mechanism if and only if $\Pr$ satisfies CAR.
\subsection{When CAR is guaranteed to hold}
We first consider the only situation where CAR is guaranteed to
hold:~if the sets in $O$ are pairwise disjoint.
\commentout{
There is only
one simple situation
where it is guaranteed to hold.
Roughly speaking, this is when the observations are pairwise disjoint.
Given a system $\R$, let $O = \{U_1, \ldots, U_n\}$ be the set of
observations made in $\R$.  

For case (b) let $V_i$ be the set of worlds where $U_i$ is
observed; that is $V_i = \{X_W(r): X_O(r) = U_i, r \in \R\}$, for $i = 1,
\ldots, n$.  Since we have assumed that observations are accurate, we
must have that $V_i \subseteq U_i$.  
Let $O^{\R} = \{V_1, \ldots, V_n\}$.  If the sets in $O^{\R}$ are
pairwise disjoint, 
then for each probability distribution $\Pr$ on $\R$
and each world $w \in V_j$ such that $\Pr(X_W = w) > 0$, it must
be the case that $\Pr(X_O = U_j \mid X_W = w) = 1$.  Thus, part (d) of
Theorem~\ref{thm:CAR} applies.
Note that, if the sets in $O$ are pairwise disjoint, then the sets in
$O^{\R}$ must also be pairwise disjoint.
Whenever the set $O^{\R}$ does {\em not\/} consist of pairwise disjoint
subsets of $W$, one can
construct distributions $\Pr$ on $\R$ such that the CAR condition
does not hold. 
 Summarizing:
\pro
\label{pro:CARmusthold}
The CAR condition holds for all distributions $\Pr$ on $\R$ if
and only if $O^{\R}$ consists of pairwise disjoint subsets of $W$.  \epro 
}
\pro
\label{pro:CARmusthold}
The CAR condition holds for all distributions $\Pr$ on $\R$ if
and only if $O$ consists of pairwise disjoint subsets of $W$. 
\epro 
\commentout{
Note that the sets in $O$ are pairwise disjoint iff 
$X_O$ can be viewed as a function on $W$ (i.e., its value in a run $r$
is completely determined by $r_W(0)$).
If $O$ is a partition of $W$, then the sets in $O$ must be pairwise disjoint.
If all $w \in W$ have positive probability $\Pr(X_W = w) > 0$, then we
also have the converse:~pairwise disjointness of the sets in $O$
implies that $O$ is a partition of $W$.
}
What happens if the sets in $O$ are not pairwise disjoint?  
Are there still cases
(combinations of $O$, $W$, and distributions on $\R$)
when CAR holds?  There are, but they are 
quite
special.  
\subsection{When CAR may hold}
We now present a lemma that provides a new characterization of CAR in terms of a
simple $0/1$-matrix. The lemma allows us to determine
for many combinations of $O$ and $W$, whether a distribution on $\R$
exists that satisfies CAR and gives certain worlds positive
probability.

Fix a set $\R$ of runs, whose worlds are in 
some finite set $W$ and whose observations come from some finite set $O
= \{ U_1, \ldots, U_n\}$. We say 
that $A \subseteq W$ is an {\em $\R$-atom\/}  relative to $W$ and $O$ if 
$A$ has the form $V_1 \inter \ldots \inter V_n$, where each $V_i$ is
either $U_i$ or $\overline{U}_i$, and 
$\{r \in \R: X_W(r) \in A \} \ne \emptyset$.  
Let $\A = \{A_1, \ldots, A_m\}$ be the set of $\R$-atoms relative to $W$ and
$O$. 
We can think
  of ${\cal A}$  as a partition of the worlds according to 
  what can be observed. 
Two worlds $w_1$ and $w_2$ are in the same set $A_i \in \A$ if there are
no observations that distinguish them; that is, 
there is no observation $U \in O$ such that $w_1 \in U$ and $w_2 \not \in U$.
Define the $m \times n$ matrix $S$ with entries $s_{ij}$ as follows:
\begin{equation}
s_{ij} =  \left\{ \begin{array}{ll}  1 & \mbox{if $A_i \subseteq U_j$} \\ 
0 &  
 \mbox{otherwise.} \end{array}
\right.
\end{equation}
We call $S$ the {\em CARacterizing matrix\/} (for $O$ and $W$).
Note that each row $i$ in $S$ corresponds to a unique atom in $\A$; we call
this the atom {\em corresponding\/} to row $i$.
This  matrix (actually, its transpose) was first introduced (but for a
different purpose) in \cite{GillLR97}.
\xam
\label{xam:extra}
Returning to Example~\ref{xam:CARhard1}, the CARacterizing
matrix is given by
$$
\left( \begin{array}{cc}
1 & 0 \\
1 & 1 \\
0 & 1
\end{array} \right),
$$
where the columns correspond to $U_1$ and $U_2$ and the rows
correspond to the three atoms $U_1 - U_2, U_1 \cap U_2$ and $U_2 - U_1$.
For example, the fact that entry $s_{31}$ of this matrix is $0$
indicates that $U_1$ cannot be observed if the actual world $w$ is in
$U_2 - U_1$. 
\exam
In the following lemma, 
$\vec{\gamma}^T$ denotes the transpose of the (row) vector $\vec{\gamma}$,
and $\vec{1}$ denotes the row vector consisting of all $1$s.  

\lem
\label{lem:precharCAR}
Let $\R$ be the set of runs over observations $O$ and worlds $W$, and
let $S$ be the CARacterizing matrix for $O$ and $W$.
\begin{itemize}
\item[(a)]
Let $\Pr$ be any distribution over $\R$ and let 
$S'$ be the matrix obtained by deleting from $S$ all rows
corresponding to an atom $A$ with $\Pr(X_W \in A) = 0$. Define 
the vector
$\vec{\gamma} = (\gamma_1, \ldots, \gamma_n)$ by setting
$\gamma_j = \Pr(X_O = U_j \mid X_W \in U_j)$
if $\Pr(X_W \in U_j) > 0$, and $\gamma_j = 0$ otherwise,
for $j = 1 \ldots n$.
If $\Pr$
satisfies CAR, then 
$S' \cdot \vec{\gamma}^T = \vec{1}^T$.
\item[(b)]
Let $S'$ be a matrix consisting of 
a subset of the rows
of $S$, and let ${\cal P}_{W,S'}$ be the set of distributions over $W$
with support corresponding to $S'$; i.e., 
$${\cal P}_{W,S'} = \{ P_W \mid P_W(A) > 0 \ \mbox{iff $A$
  corresponds to a row in $S'$} \}.
$$ 
If there exists a vector $\vec{\gamma} \ge \vec{0}$
such that $S' \cdot \vec{\gamma}^T = \vec{1}^T$,
then, for {\em all\/} $P_W \in {\cal P}_{W,S'}$, there exists a
distribution $\Pr$ 
over $\R$ with $\Pr_W = P_W$ (i.e., the marginal of $\Pr$ on $W$ is
$P_W$) such that (a) $\Pr$ satisfies CAR and  (b)
$\Pr(X_O = U_j \mid X_W \in U_j) = \gamma_j$ for all $j$ with
$\Pr(X_W \in U_j) > 0$.
\end{itemize}
\elem
Note that (b) is essentially a converse of (a).
A natural question to ask is
whether (b) would still hold if we replaced ``for all  $P_W \in {\cal
  P}_{W,S'}$ there exists $\Pr$ satisfying CAR with $\Pr_W = P_W$''
by ``for all distributions $P_O$ over $O$  there exists $\Pr$
satisfying CAR with $\Pr_O = P_O$.'' 
The answer is no; see 
Example~\ref{xam:CARhard2}(b)(ii).

Lemma~\ref{lem:precharCAR} says that a distribution $\Pr$ that
satisfies CAR and at the same time has $\Pr(X_W \in A) > 0$ for $m$
different atoms $A$ can exist if and only if a certain set of $m$
linear equations in $n$ unknowns has a solution. In many situations of
interest, $m \geq n$ (note that $m$ may be as large as $2^n-1$). Not
surprisingly then, in such situations there often can be {\em no\/}
distribution $\Pr$ that satisfies CAR, as we show in the next
subsection.  On the other hand, if the set of equations $S'
\vec{\gamma}^T = \vec{1}$ does have a solution in $\vec{\gamma}$, then
the set of all solutions forms the intersection of an affine subspace
(i.e. a hyperplane) of ${\bf R}^n$ and the positive orthant
$[0,\infty)^n$. These solutions are just the conditional probabilities
$\Pr(X_O = U_j \mid X_W \in U_j)$ for all distributions for which CAR
holds that have support corresponding to $S'$. These conditional
probabilities may then be extended to a distribution over $\R$ by
setting $\Pr_W = P_W$ for an arbitrary distribution $P_W$ over the
worlds in atoms corresponding to $S'$; all $\Pr$ constructed in this
way satisfy CAR.

Summarizing, we have the remarkable fact that for any given set of atoms
$\A$ there are only two possibilities:
either {\em no\/} distribution exists which has $\Pr(X_W \in A) > 0$
for all $A \in \A$ and satisfies CAR, or {\em for all\/} distributions
$P_W$ over worlds corresponding to atoms in $A$, there exists a
distribution satisfying CAR with marginal distribution over worlds
equal to $P_W$.

\subsection{When CAR is guaranteed {\em not\/} to hold}
We now present a theorem that gives two explicit and easy-to-check
sufficient conditions under which CAR cannot hold unless the
probabilities of some atoms and/or observations are $0$. The theorem
is proved by showing that the condition of 
Lemma~\ref{lem:precharCAR}(a) cannot hold under the stated conditions.
\commentout{
We now develop a characterization of CAR that allows us to determine,
for many combinations of $O$ and $W$, whether a distribution on $\R$
exists that satisfies CAR
for which
and gives certain worlds positive probability.
Fix a set $\R$ of runs, whose worlds are in 
some finite set $W$ and whose observations come from some finite set $O
= \{ U_1, \ldots, U_n\}$. We say 
that $A \subseteq W$ is an {\em $\R$-atom\/}  relative to $W$ and $O$ if 
$A$ has the form $V_1 \inter \ldots \inter V_n$, where each $V_i$ is
either $U_i$ or $\overline{U}_i$, and $\{r: X_W(r) \in 
A
 \} \ne
\emptyset$.  

Let $\A = \{A_1, \ldots, A_m\}$ be the set of $\R$-atoms relative to $W$ and
$O$. 
Define the $m \times n$ matrix $S$ with entries $s_{ij}$ as follows:
\begin{equation}
s_{ij} =  \left\{ \begin{array}{ll}  1 & \mbox{if $A_i \subseteq U_j$} \\ 
0 &  
 \mbox{otherwise.} \end{array}
\right.
\end{equation}
We call $S$ the {\em CARacterizing matrix\/} (for $O$ and $W$).
Note that each row $i$ in $S$ corresponds to a unique atom in $\A$; we call
this the atom {\em corresponding\/} to row $i$.
This  matrix (actually, its transpose) was first introduced (but for a
different purpose) in \cite{GillLR97}.
}
We briefly recall some standard definitions from linear algebra.
A set of vectors $\vec{v}_1, \ldots, \vec{v}_m$ is called {\em linearly
dependent\/} 
if
there exist coefficients $\lambda_1, \ldots, \lambda_m$ 
(not all zero)
such that
$\sum_{i=1}^m \lambda_i \vec{v}_i = \vec{0}$; 
the vectors are {\em affinely dependent\/} if
there exist coefficients $\lambda_1, \ldots, \lambda_m$ 
(not all zero)
such that
$\sum_{i=1}^m \lambda_i \vec{v}_i = \vec{0}$ and $\sum_{i = 1}^m
\lambda_i = 0$. %
A vector $\vec{u}$ is called an {\em affine combination\/} of 
 $\vec{v}_1, \ldots, \vec{v}_m$ if there exist coefficients $\lambda_1, \ldots, \lambda_m$ such that
$\sum_{i=1}^m \lambda_i \vec{v}_i = \vec{u}$ and $\sum_{i = 1}^m
\lambda_i = 0$.
\thm
\label{thm:charCAR}
Let $\R$ be a set of runs over observations $O 
= \{ U_1, \ldots, U_n \}
$ and worlds $W$, and
let $S$ be the CARacterizing matrix for $O$ and $W$.
\begin{itemize}
\item[(a)] 
Suppose 
that
there exists a subset $R$ of the rows in $S$ and 
a vector $\vec{u} = (u_1, \ldots, u_n)$ that is an affine combination of 
the rows of $R$ such that $u_j \geq 0$ for all $j \in \{1, \ldots, n\}$ and
$u_{j^*} > 0$ for some $j^* \in \{1, \ldots, n\}$.
Then there is no distribution
$\Pr$ on $\R$ that satisfies CAR such that 
$\Pr(X_O = U_{j^*}) > 0$ and
$\Pr(X_W \in A) > 0$ for each
$\R$-atom $A$ corresponding to a row in $R$.
\item[(b)] If there exists a subset $R$ of the rows of $S$ that is
linearly dependent but 
not affinely dependent, then there is no distribution
$\Pr$ on $\R$ that satisfies CAR such that $\Pr(X_W \in A) > 0$ for each
$\R$-atom $A$ corresponding to a row in $R$. 
\item[(c)] Given a set $R$ consisting of $n$ linearly
independent rows of $S$ and a distribution $P_W$ on $W$ such
that $P_W(A) > 0$
for all $A$ corresponding to a row in $R$,
there is a unique distribution $P_O$ on $O$ such that
if $\Pr$ is a distribution on $\R$ satisfying
CAR and $\Pr(X_W \in A) = P_W(A)$ for each atom $A$ corresponding to a
row in $R$, then $\Pr(X_O = U) = P_O(U)$.  
\end{itemize}
\commentout{
Then there is {\em no\/} distribution on $\R$ that satisfies CAR and
that has
$\Pr(X_W \in A) > 0$ for all $A \in \A$.
\item[2. The rows of $S'$ are linearly independent,  $m' = n$:]
Then for each distribution $\Pr_O$ on $O$, there can be {\em at most one\/}
distribution $\Pr$ on $\R$ that satisfies CAR and that has (1) $\Pr(X_0
= U) = \Pr_O(U)$ for all $U 
\in O$, and (2), 
$\Pr(X_A = a) > 0$ for all $a \in A$. If such a distribution $\Pr$
exists, it is uniquely determined by the set of equations (one for
each $U \in O$):
\begin{equation}
\label{eq:charCAR}
\Pr(X_O = U \mid X_W \in U) = \frac{\Pr(X_O = U)}{\sum_{a \subset
    U} \Pr(X_A = a)}.
\end{equation}
\end{itemize}
}
\ethm

\commentout{
\paragraph{Remarks}
\begin{enumerate}
\item  For many
  combinations of $W$ and $O$ the number of
potential atoms is much larger than the number of possible
observations, and then typically $m' > n$. But if $m' > n$ then we
must be in item 1. above must be the case, and CAR cannot hold. It follows that in
many realistic situations, it is simply impossible for CAR to hold---%
the structure of $O$ and $W$ is such that {\em no\/} distribution 
on
$\R$ can satisfy CAR.
\item Our theorem says nothing about the case where the rows of
  $||S'||$ are linearly independent, but $m' < n$. In such cases,
  further analysis is needed to determine whether CAR holds or not. 
\end{enumerate}
}
It is well known that in an $m \times n$ matrix, at most $n$ rows can be
linearly independent.  
In many cases of interest (cf.~Example~\ref{xam:CARwash} below), the
number of atoms $m$ is larger than the number of observations $n$,
so that there must exist subsets $R$ of rows of $S$ that are linearly dependent. 
Thus, part (b) of Theorem~\ref{thm:charCAR}
puts nontrivial constraints on the distributions that satisfy CAR. 

The requirement 
in part (a) 
may seem somewhat obscure but 
it can be easily checked and applied in a number of situations, as
illustrated in Example~\ref{xam:CARwash} 
and \ref{xam:CARhard2} below.
Part (c) says that in many other cases 
of interest
where neither part (a) nor (b) applies,
even if a distribution on $\R$ exists satisfying CAR, the
probabilities of making the observations 
are
completely determined by the
probability of various events in the world occurring, which seems rather
unreasonable.

\xam
\label{xam:CARwash}
Consider the CARacterizing matrix of Example~\ref{xam:extra}.
Notice there exists an affine combination of the first two rows that is
not $\vec{0}$ and has no negative components:
$$
-1 \cdot  \left( \begin{array}{c}
1 \\ 0 
\end{array} \right) + 1 \cdot \left( \begin{array}{c}
1 \\ 1 
\end{array} \right) =  \left( \begin{array}{c}
0 \\ 1 
\end{array} \right).
$$
Similarly, there exists an affine combination of the last two rows that
is not $\vec{0}$ and has no negative components.  It follows from
Theorem~\ref{thm:charCAR}(a) that there is no distribution satisfying
CAR that gives 
both of the observations $X_O = U_1$ and $X_O =
U_2$ positive probability and either (a) gives both
$X_W \in U_1 - U_2$ and $X_W \in U_1 \inter U_2$
positive probability or (b) gives 
both $X_W \in U_2 - U_1$ and $X_W \in U_1 \inter U_2$
positive probability.
If both observations have positive probability, then CAR can hold only
if the probability of $U_1 \cap U_2$ is 
either $0$ or $1$. 
(Example~\ref{xam:CARhard1} already shows this using a more
direct argument.)
\exam 

The next example further illustrates that in 
general, it can be very difficult to
satisfy CAR.

\xam\label{xam:CARhard2} Suppose that 
$O = \{U_1, U_2, U_3\}$, and
all three observations
can be made with positive probability.
It turns out that in this situation the CAR condition can hold, but only if
(a) $\Pr(X_W \in U_1 \inter U_2 \inter U_3) = 1$ (i.e., all of $U_1$, $U_2$, and
$U_3$ must hold),
(b)  $\Pr(X_W \in ((U_1 \inter U_2) - U_3) \union ((U_2 \inter U_3) - U_1) \union
((U_1 \inter U_3) - U_2)) = 1$ (i.e., exactly two of $U_1$, $U_2$, and
$U_3$ must hold), (c) $\Pr(X_W \in (U_1 - (U_2 \union
U_3)) \union (U_2 - (U_1 \union U_3)) \union (U_3 - (U_2 \union U_1))) =
1$ (i.e., exactly one of $U_1$, $U_2$, or $U_3$ must hold), or
(d) one
of
$(U_1 - (U_2 \cup U_3)) \union (U_2 \inter U_3)$, $(U_2  - (U_1 \cup
U_3)) \union (U_1 \inter
U_3)$ or $(U_3  - (U_1 \cup U_2)) \union (U_1 \inter U_2)$ has probability 1 (either exactly
one of $U_1$, $U_2$, or $U_3$ holds, or the remaining two both hold).

We first check that CAR can hold in all these cases.  It should be clear
that CAR can hold in case (a).  Moreover,
there are no constraints on $\Pr(X_O = U_i \mid X_W = w)$
for $w \in U_1 \inter U_2 \inter U_3$ (except, by the CAR condition, for
each fixed $i$, the probability must be the same for all $w \in U_1
\inter U_2 \inter U_3$, and the three probabilities must sum to 1).
For case (b), let $A_i$ be the atom where exactly two of 
$U_1$, $U_2$, and $U_3$ hold, and $U_i$ does
not hold, for $i = 1, 2, 3$.  Suppose that $\Pr(X_W \in A_1 \union A_2 \union
A_3) = 1$.  Note that, since all three observations can be made with
positive probability, at least two of $A_1$, $A_2$, and $A_3$ must have
positive probability. 
Hence we can distinguish between two subcases: 
(i) only two of
them have positive probability, and (ii) all three have positive
probability.

For subcase (i), suppose without loss of generality that only $A_1$ and
$A_2$ have positive probability.
Then it immediately follows from the CAR condition
that there must be some $\alpha$ with $0 < \alpha < 1$ such
that $\Pr(X_O = U_3 \mid X_W = w) = \alpha$, for all $w \in A_1 \union A_2$
such that $\Pr(X_W = w) > 0$.  Thus, $\Pr(X_O = U_1 \mid X_W = w) = 1-\alpha$ for 
all $w \in A_2$ such that $\Pr(X_W = w) > 0$, and 
$\Pr(X_O = U_2 \mid X_W = w) = 1-\alpha$ for 
all $w \in A_1$ such that $\Pr(X_W = w) > 0$.

Subcase (ii) is more interesting. 
The rows of the CARacterizing matrix 
$S$
corresponding to $A_1$, $A_2$, and $A_3$ are
$(0\ 1\  1)$, 
$(1 \ 0 \ 1)$, and
$(1 \ 1 \ 0)$, respectively.
Now Lemma~\ref{lem:precharCAR}(a) tells us that if $\Pr$ satisfies
CAR, then we must have $S \cdot \vec{\gamma}^T = \vec{1}^T$ for some
$\vec{\gamma} = 
(\gamma_1, \gamma_2, \gamma_3)$ 
with $\gamma_i = \Pr(X_O = U_i \mid X_W \in
U_i)$. These three linear equations have solution 
$$\gamma_1
= \gamma_2 = \gamma_3 = \frac{1}{2}.$$
Since this solution is unique, it 
follows by Lemma~\ref{lem:precharCAR}(b) that {\em all\/}
distributions that satisfy CAR must have conditional probabilities
$\Pr(X_O = U_i \mid X_W \in 
U_i) = 1/2$,  and that their marginal
distributions on $W$ can be arbitrary.  This fully
characterizes the set of distributions $\Pr$ for which CAR holds in
this case. Note that for $i = 1,2, 3$, since we can write $\gamma_i = \Pr(X_O = U_i)/
\Pr(X_W \in U_i)$ we have $\Pr(X_O = U_i) = \Pr(X_w \in U_i) \gamma_i
\leq 1/2$ so that, in contrast to the marginal distribution over $W$,
the marginal distribution over $O$ cannot be chosen arbitrarily.

\commentout{
Since the rows are linearly independent and there are 3 possible
observations, by Theorem~\ref{thm:charCAR}(b), 
given a probability $P_W$ on the atoms corresponding to these three rows
which gives all three rows positive probability, 
there is a unique probability $P_O$ on the observations such that if
there is a distribution $\Pr$ on the runs satisfying CAR with $\Pr_W =
P_W$, then $\Pr_O = P_O$.
In fact, an even stronger result holds
in this case.  The probability $P_O$ is independent of $P_W$.  Indeed,
the 
probability of observing $U_i$ must be $1/2$ in all worlds in $U_i$, for
$i = 1, 2, 3$.  
For let
$w_1 \in U_1 \inter U_3$, $w_2 \in U_1 \inter U_2$, 
and $w_3 \in U_2 \inter U_3$,
and 
\begin{equation}
\label{eq:hithere}
\Pr(X_O = U_1 \mid X_W = w_1) = \alpha. 
\end{equation}
Conditioned on $X_W = w_1$, $X_O \in \{ U_1,U_3\}$. Thus
(\ref{eq:hithere}) implies that
$\Pr(X_O = U_3 \mid X_W = w_1) = 1 - \alpha$. 
Further, by the CAR condition, item (c), 
\begin{equation}
\label{eq:hello}
\Pr(X_O = U_3 \mid X_W = w_3) = 1-
\alpha.
\end{equation}
Also,
since $\{ w_1, w_2 \} \subseteq U_1$,  
by Theorem~\ref{thm:CAR}(c) again  
and (\ref{eq:hithere}), we have $\Pr(X_O = U_1 \mid X_W = w_2) =
\alpha$. Since conditioned on $X_W = w_2$, $X_O \in \{
U_1,U_2\}$, this further implies that $\Pr(X_O = U_2 \mid X_W = w_2) =
1 - \alpha$. Since $\{ w_2, w_3 \} \subseteq U_2$, by 
the CAR condition
we have $\Pr(X_O = U_2 \mid X_W = w_3) = 1- \alpha$. 
Conditioned on $X_W = w_3$, $X_O \in \{ U_2,U_3\}$, 
so $\Pr(X_O = U_3 \mid X_W = w_3) = \alpha$. Together
with (\ref{eq:hello}), this gives $\alpha = 1/2$.

}
In case (c), it should also be clear that CAR can hold.  Moreover,
$\Pr(X_0 = U_i \mid X_W = w)$ is either 0 or 1, depending on whether $w
\in U_i$.  
Finally, for case (d), suppose that
$\Pr(X_W \in U_1 \union (U_2 \inter U_3)) = 1$.  CAR holds iff there exists $\alpha$
such that $\Pr(X_O = U_2 \mid X_W = w) = \alpha$ and 
$\Pr(X_O = U_3 \mid X_W = w) = 1 - \alpha$ 
for all $w \in U_2 \inter U_3$ such that $\Pr(X_W = w) > 0$.
(Of course, $\Pr(X_O = U_1 \mid X_W = w) = 1$ for all $w \in U_1$ such
that $\Pr(X_W = w) > 0$.)

Now we show that CAR cannot hold in any other case.  First suppose
that $0 < \Pr(X_W \in U_1 \inter U_2 \inter U_3) < 1$.  
Thus, there must
be at least one other atom $A$ such that $\Pr(X_W \in A) > 0$.
\commentout{
This is a case in which Theorem~\ref{thm:charCAR} is of no help since
the CARacterizing matrix has $m = 2$ linearly independent rows and $n
= 3 >2$ columns. 
Nevertheless, we can show that CAR cannot hold by direct means.
Namely, choose $w \in U_1 \inter
U_2 \inter U_3$ such that $\Pr(X_W = w) > 0$, and let $\Pr(X_O = U_i \mid X_W
= w) = \alpha_i$, for $i = 1, 2, 3$.  Note that $\alpha_1 + \alpha_2 +
\alpha_3 = 1$.
Suppose  $w' \notin U_1 \inter U_2
\inter U_3$ and $\Pr(X_W = w') > 0$.  
By the CAR condition, $\Pr(X_O =  U_i \mid X_W = w')$ is either 
$\alpha_i$ or 0, depending on whether $w' \in U_i$ or not.
Since $\Pr(X_O = U_1 \mid X_W = w') + \Pr(X_O = U_2 \mid X_W = w') + 
\Pr(X_O = U_3 \mid X_W = w') = 1$, and at least one of these terms is 0,
we get the desired contradiction.
}
The row corresponding to the atom $U_1 \inter U_2 \inter U_3$ is $(1 \ 1
\ 1)$.  Suppose $r$ is the row corresponding to the other atom $A$.
Since $S$ is a 0-1 matrix, the vector $(1 \ 1 \ 1) - r$ gives is an
affine combination of $(1 \ 1 \ 1)$ and $r$ that is nonzero and has
nonnegative components.  
It now follows by Theorem~\ref{thm:charCAR} that CAR cannot hold in this
case. 
 
Similar arguments give a contradiction in all the other cases; we leave
details to the reader. 
\exam

\commentout{
Part (a) of Theorem~\ref{thm:charCAR} applies even more generally than
these examples suggest.  
It is well known that in an $m \times n$ matrix, at most $n$ rows can be
linearly independent.  Since the number of atoms is typically much
larger than the number of worlds (that is, $m$ is typically much larger
than $n$), there will typically be many subsets $R$ of rows of $S$ that
are linearly dependent.  As the following result shows, if the rows that
are linearly dependent are not affinely independent, it follows from
Theorem~\ref{thm:charCAR}(a) that there is no distribution $\Pr$ on $\R$
that satisfies CAR.  This shows that 

\pro\label{pro:bimpliesa}
If there exists a subset $R$ of rows of $S$
that is linearly dependent but not affinely dependent, 
then
for all $\R$-atoms $A$ corresponding to a row in $R$, for all $j^* \in
\{1, \ldots, n\}$, we have: if $A \in U_{j^*}$, then
the antecedent of
Theorem~\ref{thm:charCAR}(a) holds
for %
\epro
}
\commentout{
that CAR holds. If $U_1 - (U_2 \union U_3)$ is nonempty, then the
same arguments as in Example~\ref{xam:CARhard1} show that the
probability of observing $U_2$ is 0, a contradiction.  
In the same way 
we can show 
that $U_2 - (U_1 \union U_3)= U_3 - (U_1
\union U_2) = \emptyset$.

It is also not hard to show that $U_1 \inter U_2
\inter U_3 = \emptyset$.  For suppose that $w_0 \in U_1 \inter U_2
\inter U_3$ and
$w_1 \in U_1 - U_2$.  Since $U_1 - (U_2 \union U_3) = \emptyset$, it
must be the case that $w_1 \in U_1 \inter
U_3$.  Let $\alpha = \Pr(X_O = U_1 \mid X_W = w_1)$.  Then $\Pr(X_O =
U_3 \mid X_W = w_1) = 1 - \alpha$, since in runs where the world is
$w_1$, the agent must observe either $U_1$ or $U_3$.  By CAR, it follows
that
$\Pr(X_O = U_1 \mid X_W = w_0) = \alpha$ and
$\Pr(X_O = U_3 \mid X_W = w_0) = 1- \alpha$.  Thus,
$\Pr(X_O = U_2 \mid X_W = w_0) = 0$.  Using CAR, it follows that the
probability of observing $U_2$ is 0, contradicting our initial
assumption.  Thus, $U_1 \inter U_2 \inter U_3 =
\emptyset$.

We have shown that, assuming CAR holds, every world in $W$ must be in exactly one of
$U_1 \inter U_2$, $U_1 \inter U_3$, or $U_2 \inter U_3$.
It is then easy to see that the only way to satisfy CAR is to have the
probability of observing $U_i$ be $1/2$ in all worlds in $U_i$.  For let
$w_1 \in U_1 \inter U_3$, $w_2 \in U_1 \inter U_2$, 
and $w_3 \in U_2 \inter U_3$,
and 
\begin{equation}
\label{eq:hithere}
\Pr(X_O = U_1 \mid X_W = w_1) = \alpha. 
\end{equation}
Conditioned on $X_W = w_1$, $X_O \in \{ U_1,U_3\}$. Thus
(\ref{eq:hithere}) implies that
$\Pr(X_O = U_3 \mid X_W = w_1) = 1 - \alpha$. 
Further, by the CAR condition, item (c), 
\begin{equation}
\label{eq:hello}
\Pr(X_O = U_3 \mid X_W = w_3) = 1-
\alpha.
\end{equation}
Also,
since $\{ w_1, w_2 \} \subseteq U_1$,  
by Theorem~\ref{thm:CAR}(c) again  
and (\ref{eq:hithere}), we have $\Pr(X_O = U_1 \mid X_W = w_2) =
\alpha$. Since conditioned on $X_W = w_2$, $X_O \in \{
U_1,U_2\}$, this further implies that $\Pr(X_O = U_2 \mid X_W = w_2) =
1 - \alpha$. Since $\{ w_2, w_3 \} \subseteq U_2$, by 
the CAR condition
we have $\Pr(X_O = U_2 \mid X_W = w_3) = 1- \alpha$. 
Conditioned on $X_W = w_3$, $X_O \in \{ U_2,U_3\}$, 
so $\Pr(X_O = U_3 \mid X_W = w_3) = \alpha$. Together
with (\ref{eq:hello}), this gives $\alpha = 1/2$.
\exam
}
\commentout{
\subsection{Discussion: ``CAR is everything'' or ``often CAR is nothing''?}
\label{sec:CARnothing}
In one of their main theorems,
Gill, van der Laan, and Robins \citeyear[Section 2]{GillLR97} show that
for every finite set $W$ of worlds,
every set $O$ of observations, and every distribution 
$P_O$
on $O$, there is a distribution 
$\Pr^*$
on $\R$ such that 
$\Pr_O^*$ (the marginal of 
$\Pr^*$
on $O$)
is equal to $P_O$ and $\Pr^*$ satisfies CAR.

The authors summarize this as 
``CAR is   everything''.
We regard this interpretation of the theorem as somewhat misleading. 
Theorem~\ref{thm:charCAR} shows that for many combinations of $O$
and $W$, CAR can hold {\em only\/} for distributions $\Pr$ with
$\Pr(X_W \in A) = 0$ for some atoms $A$.
(In the previous sections, we called such distributions ``degenerate''.)
In our view, this says that for such $O$ and $W$, CAR
effectively cannot hold: 
if we chose to formalize the problem using a
particular $W$ then we do not {\em a priori\/} want to assume that
a particular atom $A \subseteq W$ will never occur; if we had been
willing to assume that, 
we would have formalized
the problem using $W - A$ rather than $W$. 
But CAR forces us to make just such an assumption.

Its misleading name ``CAR is everything'' notwithstanding,
Gill, van der Laan, and Robins' result is quite strong; to get an
idea about the constraints that CAR imposes on $\Pr$, our
Theorem~\ref{thm:charCAR} should certainly be 
contrasted with their theorem and 
Theorem~\ref{thm:CARresolved} of the 
next subsection.
Given that CAR is so difficult to satisfy, 
the reader may wonder why there is so much study of
  the CAR condition in the statistics literature. The reason is
  that some of the special situations in which 
CAR holds often arise in missing data and survival analysis problems.
Here is an example.
Suppose that the set of observations can be written as 
$O= \cup_{i=1}^k \Pi_i$, where each 
$\Pi_i$ is a partition of $W$
(that is, a set of pairwise disjoint subsets of $W$ whose union is $W$).
Further suppose that observations are
generated by the following process,
which we call {\sc CARgen}. 
Some $i$ between 1 and $k$ is chosen according to some
arbitrary distribution $P_0$; independently, $w \in W$ is chosen 
according to $P_W$. 
The agent then observes the 
unique $U \in \Pi_i$ such that
  $w \in U$. 
Intuitively, the partitions $\Pi_i$ represent the observations that can be
made with a particular sensor.  Thus, $P_0$ determines the probability
that a particular sensor is chosen; $P_W$ determines the probability
that a particular world is chosen.  
The sensor and the world together determine the observation that is made.
It is easy to see that this mechanism induces a distribution on
  $\R$ for which CAR holds. 

The 
special case with $O = \Pi_1
\cup \Pi_2$, $\Pi_1 = \{W\}$, and 
$\Pi_2
= \{ \{ w \} \mid  w \in W\}$ 
corresponds to a simple missing data problem.
Intuitively, either complete information is given, or
there is no data at all.
In this context, CAR is often called {\em MAR:  missing at random\/}. 
In more realistic MAR problems, 
we may observe 
a vector with some of its components missing.  In such cases 
the CAR condition often still holds. 

}
\subsection{Discussion: ``CAR is everything'' vs.~``sometimes CAR is
nothing''}
\label{sec:CARnothing}

In one of their main theorems,
  Gill, van der Laan, and Robins \citeyear[Section 2]{GillLR97} show
  that the CAR assumption is untestable from observations of $X_O$
  alone, in the sense that the assumption ``$\Pr$ satisfies CAR''
  imposes no restrictions at all on the marginal distribution $\Pr_O$
  on $X_O$. More precisely, they show that for every finite set $W$ of
  worlds, every set $O$ of observations, and every distribution $P_O$
  on $O$, there is a distribution $\Pr^*$ on $\R$ such that $\Pr_O^*$
  (the marginal of $\Pr^*$ on $O$) is equal to $P_O$ and $\Pr^*$
  satisfies CAR.  The authors summarize this as ``CAR is everything''.

We must be careful in interpreting this result.
Theorem~\ref{thm:charCAR} shows that, for many combinations of $O$
and $W$, CAR can hold {\em only\/} for distributions $\Pr$ with
$\Pr(X_W \in A) = 0$ for some atoms $A$.
(In the previous sections, we called such distributions ``degenerate''.)
In our view, this says that 
in some cases, CAR effectively cannot hold. To see why, first 
suppose we are given a set $W$ of worlds and a set $O$ of observations.
Now we may feel confident {\em a priori\/} that
some $U_0 \in O$ and some $w_0 \in W$ cannot occur in practice.  In
this case, we are willing to consider only distributions $\Pr$ on $O
\times W$ that have $\Pr(X_O = U_0) = 0$, $\Pr(X_W= w_0) = 0$. 
(For example, $W$ may be a product space $W= W_a \times W_b$ and it is
known that some combination $w_a \in W_a$ and $w_b$ in $W_b$ can never
occur together; then $\Pr(X_w = (w_a,w_b)) = 0$.)
Define $O^*$ to be the subset of $O$ consisting of all $U$
that we cannot a priori rule out; similarly, $W^*$ is the subset of
$W$ consisting of all $w$ that we cannot a priori rule out. By
Theorem~\ref{thm:charCAR}, it is still possible that $O^*$ and $W^*$
are such that, even if we restrict to runs where only observations in
$O^*$ are made, CAR can only hold if $\Pr(X_W \in A) = 0$ for some
atoms (nonempty subsets) $A \subseteq W^*$. 
This means that CAR may force us to assign probability 0 to some events 
that, {\em a priori}, were considered possible.
Examples~\ref{xam:CARhard1} and~\ref{xam:CARhard2} illustrate this
phenomeonon. 
We may summarize this as ``sometimes CAR is nothing''.
Given therefore that CAR imposes such strong conditions, 
the reader may wonder why there is so much study of
  the CAR condition in the statistics literature. The reason is
  that some of the special situations in which 
CAR holds often arise in missing data and survival analysis problems.
Here is an example:
Suppose that the set of observations can be written as 
$O= \cup_{i=1}^k \Pi_i$, where each 
$\Pi_i$ is a partition of $W$
(that is, a set of pairwise disjoint subsets of $W$ whose union is $W$).
Further suppose that observations are
generated by the following process,
which we call {\sc CARgen}. 
Some $i$ between 1 and $k$ is chosen according to some
arbitrary distribution $P_0$; independently, $w \in W$ is chosen 
according to $P_W$. 
The agent then observes the 
unique $U \in \Pi_i$ such that
  $w \in U$. 
Intuitively, the partitions $\Pi_i$ may represent the observations that can be
made with a particular sensor.  Thus, $P_0$ determines the probability
that a particular sensor is chosen; $P_W$ determines the probability
that a particular world is chosen.  
The sensor and the world together determine the observation that is made.
It is easy to see that this mechanism induces a distribution on
  $\R$ for which CAR holds. 

The  special case with $O = \Pi_1
\cup \Pi_2$, $\Pi_1 = \{W\}$, and 
$\Pi_2
= \{ \{ w \} \mid  w \in W\}$ 
corresponds to a simple missing data problem (Example~\ref{xam:mar} below).
Intuitively, either complete information is given, or
there is no data at all. In this context, CAR is often called {\em MAR:  missing at random\/}. 
In more realistic MAR problems, 
we may observe a vector with some of its components missing.  In such cases 
the CAR condition sometimes still holds. 
In practical missing data problems, the goal is often
to infer the distribution $\Pr$ on runs $\R$ from
successive observations of $X_O$. That is, one observes a sample
$U_{(1)}, U_{(2)}, \ldots,U_{(n)}$, where $U_{(i)} \in O$. Typically, the
$U_{(i)}$ are assumed to be an i.i.d. (independently identically
distributed) sample of outcomes of $X_O$. 
The corresponding ``worlds'' $w_1, w_2, \ldots$
(outcomes of $X_W$) are {\em not\/} observed. Depending on the
situation, $\Pr$ may be completely unknown or is assumed to be a
member of some parametric family of distributions. If the number of
observations $n$ is large, then clearly the sample $U_{(1)}, U_{(2)},
\ldots,U_{(n)}$ can be used to obtain a reasonable estimate of
$\Pr_O$, the marginal distribution on $X_O$. But one is interested in
the full distribution $\Pr$. That distribution usually cannot be
inferred without making additional assumptions, such as the CAR assumption.
\xam\label{xam:mar}
(adapted from \cite{ScharfsteinDR02})
Suppose that a medical study is conducted to test the effect of a new
drug. The drug is administered to a group of patients on a weekly
basis. Before the experiment is started and after it is finished, 
some characteristic (say, the blood pressure) 
of the patients is measured. The data are thus differences in blood
pressure for individual
patients before and after the treatment. In practical studies of this kind, often several of the patients drop out of the
experiment. For such patients there is then no data. We model this as follows:
$W$ is the set of possible values of the characteristic we are
interested in (e.g., blood pressure difference). $O = \Pi_1
\cup \Pi_2$ with $\Pi_1 = \{W\}$, and 
$\Pi_2
= \{ \{ w \} \mid  w \in W\}$ as above. For ``compliers'' (patients that
did not drop out), we observe $X_O = \{w \}$, where $w$ is the value of
the characteristic we want to measure. For dropouts, we
observe $X_O = W $ (that is, we observe nothing at all).
We thus have, for example, a sequence of observations $U_1 = \{ w_1 \}, U_2 =
\{w_2\}, U_3 = W, U_4 = \{w_4 \}, U_5 = W, \ldots, U_n = \{w_n
\}$. 
If this sample
is large enough, we can use it to obtain a reasonable estimate of the probability
that a patient drops out (the ratio of outcomes with $U_i = W$ to
the total number of outcomes). We can also get a reasonable estimate
of the distribution of $X_W$ for the complying patients. 
Together these two distributions determine the distribution 
of $X_O$.

We are interested in the effect of the drug in the general
population. Unfortunately, it may be the case that the effect on
dropouts is different from the effect on compliers.
(Scharfstein, Daniels, and Robins \citeyear{ScharfsteinDR02} discuss an
actual medical study in 
which physicians judged the effect on dropouts to be very  different
from the effect of compliers.)  
Then we cannot infer the distribution on $W$ from the observations
$U_1, U_2, \ldots$ alone without making additional assumptions about
how the distribution for dropouts is related to the distribution for
compliers. Perhaps the simplest such assumption that one can make is that
the distribution of $X_W$ for dropouts {\em is\/} in fact the same as
the distribution of $X_W$ for compliers: the data are ``missing at
random''. Of course, this assumption is just the CAR
assumption.  By Theorem~\ref{thm:CAR}(a), CAR holds iff
for all $w \in W$  
$$\Pr(X_W = w \mid X_O = W ) = \Pr(X_W = w \mid X_W \in W) = \Pr(X_W
= w),$$
which means just that the distribution of $W$ is
independent of whether a patient drops out ($X_O =
W$) or not.
Thus, {\em if\/} CAR can be assumed, then we can infer the distribution on $W$
(which is what we are really interested in).
\exam 
Many problems in missing data and survival analysis
are of the kind illustrated above:  The analysis would be greatly
simplified if CAR 
holds, but whether or not this is so is not clear. It is therefore of
obvious interest to investigate whether, from observing the ``coarsened''
data $U_{(1)}, U_{(2)}, \ldots, U_{(n)}$ alone, it may already be
  possible to test the assumption that CAR holds. For example, one
  might imagine that there are distributions on $X_O$ for which CAR
  simply cannot hold. If the empirical distribution of the $U_{i}$
were ``close'' (in the appropriate sense) to a distribution that
  rules out CAR, the statistician might infer that $\Pr$ does not
  satisfy CAR. 
Unfortunately, if $O$ is   finite, then the result of 
  Gill, van der Laan, and Robins \citeyear[Section 2]{GillLR97} 
referred to at the beginning of this section shows that we can never
rule out CAR in this way.
We are interested in the question
of  whether CAR can hold in a
``nondegenerate'' sense, given $O$ and $W$. From this point of
view, the slogan 
``sometimes CAR is nothing''
makes sense. In
contrast,
\cite{GillLR97} were interested in the question whether
CAR can be tested from observations of $X_O$ alone. From that point
of view, the slogan ``CAR is everything'' makes perfect sense.
In fact, Gill, van der Laan, and Robins were quite aware, and explicitly
stated,  that CAR imposes very strong assumptions on the
 distribution $\Pr$. In a later paper,
it  was even implicitly stated that in
some cases CAR forces $\Pr(X_W \in  A) =0$ for some atoms $A$
\cite[Section 9.1]{RobinsRS99}.   
Our contribution is to provide the precise 
 conditions (Lemma~\ref{lem:precharCAR} and Theorem~\ref{thm:charCAR})
 under which this happens. 

Robins, Rodnitzky, and Scharfstein \citeyear{RobinsRS99} also introduced
a Bayesian 
method (later extended in \cite{ScharfsteinDR02}) that allows one to specify a prior distribution
over a parameter $\alpha$ which indicates 
in a precise sense, how much $\Pr$ deviates from CAR. 
For example, $\alpha = 0$ corresponds to the set of distributions
$\Pr$ satisfying CAR. 
The precise connection between this work and ours
needs further investigation.

\subsection{A mechanism for generating distributions satisfying CAR}
\label{sec:CARmechanics}
In Theorem~\ref{thm:CAR}
and Lemma~\ref{lem:precharCAR}
we described CAR in an algebraic way, as a collection of
probabilities satisfying certain equalities. Is there 
a more ``procedural'' way of representing CAR?
In particular,
does there exist a single mechanism that gives rise to
CAR such that {\em every\/} case of CAR can be viewed as a special
case of this mechanism? 

Before we can answer this question, we have to make clear what counts as
a mechanism.  Without any constrainst, there is clearly a trivial
solution to the problem, as already noted by Gill, van der Laan, and Robins
\citeyear{GillLR97}:  Given a distribution $\Pr$ satisfying CAR, we
simply draw a world $w$ according to $\Pr_W$, and then draw $U$ such $w
\in U$ according to the distribution $\Pr(X_O = U \mid X_W = w)$.  This
is obviously cheating in some sense.  Intuitively, the problem here is
that we cannot ``choose'' $U$ according to a certain distribution.  We
do not have that kind of control over the observations that are made.

So what can we do?  Intuitively, the mechanism should be able to control
only what can be controlled in an experimental setup.  While it is fair
to assume that we are given some sensor, it is not fair to assume that
we can control their output (or exactly what they can sense).  
Assume that we are given a world $w \in W$, generated
according to some distribution $P_W$.  Intuitively, we do not have
control over $P_W$. Given $P_W$, our goal is to find a procedure that
generates all and only the distributions $\Pr$ satisfying CAR such that
$\Pr_W = P_W$.  One approach is to
assume that the agent gets to make observations, using
possibly different sensors.  While the agent can choose which
sensor to observe, it cannot choose what the sensor
observes.  Indeed, given a world $w$, then observation returned by the
sensor is determined.  This is exactly what is done in the {\sc CARgen}
scheme discussed in Section~\ref{sec:CARnothing}.

Gill, van der Laan, and Robins \citeyear{GillLR97}
consider another approach.  They 
show that in several
problems of survival analysis, 
observations are generated according to 
what they call
a {\em randomized monotone
coarsening scheme}. 
They also show that their randomized scheme generates only distributions
that satisfy CAR.
In fact, the randomized monotone coarsening scheme turns out to be a
special case of 
{\sc CARgen},
although we do not prove this here.
Gill, van der Laan, and Robins show by example that the randomized
coarsening schemes do not suffice to generate all CAR distributions.  We
now use essentially the same example to show that {\sc CARgen} does not
either.

\xam
\label{xam:CARwithoutpartition}
Consider subcase (ii) of Example~\ref{xam:CARhard2} again.  Let 
$U_1, U_2$, $U_3$ and $A_1$, $A_2$ and $A_3$ be as in that example,
and assume for simplicity that $W = A_1 \cup A_2 \cup A_3$. The
example showed that there exists distributions $\Pr$ satisfying CAR in
this case with $\Pr(A_i) > 0$ for $i \in \{1, 2, 3\}$, all having conditional
probabilities $\Pr(X_O = U_i \mid X_W = w) = 1/2$ for all $w \in U_i$. 
Clearly, $U_1, U_2$ and $U_3$ cannot be grouped together to form a set
of partitions of $W$. So, even though CAR holds for $\Pr$, {\sc
CARgen} cannot be used to simulate $\Pr$. 
\exam

The problem of finding a natural mechanism that generates all and only
distributions that satisfy CAR  
seems to be one of the goal of Gill, van der Laan, and Robins' work
(see, in particular, \citeyear[Section 3]{GillLR97}), although they do
not formulate the problem precisely.  While we also do not give a
precise formulation of what counts as a reasonable mechanism (although
it can be done in the runs framework---essentially, each step of the
algorithm can depend only on information available to the experimenter,
where the ``information'' is encoded in the observations made by the
experimenter in the course of running the algorithm), we do give
an argument that the mechanism we propose is in fact reasonable.
We call the procedure ${\sc CARgen}^*$, since it extends {\sc CARgen}.
Just like {\sc CARgen}, ${\sc CARgen}^*$ assumes that there is a
collection of sensors, and it consults a given sensor with a certain
predetermined probability. However, unlike {\sc CARgen}, ${\sc
CARgen}^*$ may ignore a sensor reading. 

\paragraph{Procedure {\sc CARgen$^{*}$}}
\begin{enumerate}
\item Preparation:
\begin{itemize} 
\item Fix an arbitrary distribution $P_W$ on $W$.
\item Fix a set $\P$ of partitions of $W$, and fix an arbitrary
  distribution $P_{\P}$ on $\P$.
\item Choose numbers $q \in [0,1)$ and $q_{U|\Pi} \in [0,1]$ for each
pair $(U,\Pi)$ such that $\Pi \in \P$ and $U \in \Pi$ satisfying the
following constraint, for each $w \in W$ 
such that $P_W(w) > 0$:
\begin{equation}\label{CARconstraint}
q = \sum_{\{(U, \Pi): \; w \in U , \,  U \in \Pi\}} P_\P(\Pi)q_{U|\Pi}.
\end{equation}
\end{itemize}
\item Generation:
\begin{itemize}
\item[2.1] Choose $w \in W$ according to $P_W$.
\item[2.2] Choose $\Pi \in \P$ according to $P_{\P}$. Let $U$ be the
unique set in $\Pi$ such that $w \in U$.
\item[2.3] With probability $1- q_{U|\Pi}$, return $(w,U)$ and halt.
With probability $q_{U|\Pi}$,
 go to step 2.2.
\end{itemize}
\end{enumerate}
\commentout{
${\sc CARgen}^{**}$ is an extension of the procedure
{\sc CARgen} described previously. Each partition $\Pi \in\P$ may be interpreted as
a measuring device or sensor. Just as in {\sc CARgen},
a sensor $s$ is chosen independently of a world $w \in W$. This gives rise
probability $q_{U|\Pi}$, and starts the procedure of generating a world
and a sensor all over again. We will refer to $q_{U|\Pi}$ as the {\em
rejection probability of $U$ given sensor $s$}.

Each set of distributions/numbers $P_W$, $P_\P$, $q_{U|\Pi}$ determines a
distribution on $\R$ (More precisely, it determines a distribution on $\R \cup \{
\uparrow \}$, where $\uparrow$ is the event that ${\sc CARgen}^{**}$
runs forever). 
We denote this distribution by $\Pr^*$.
${\sc CARgen}^{**}$ in itself does {\em  not\/} guarantee that $\Pr^*$ satisfies CAR. To achieve this, we need to
  constrain the rejection probabilities $q_{U|\Pi}$ in the following
  manner. 
For $w
  \in W$, define the {\em  marginal rejection probability\/} $q_w$ as
$$
q_w := \sum_{U: \; w \in U \; ; \; s: U \in s} P_\P(s)q_{U|\Pi}.
$$
Intuitively, $q_w$ is the probability that a world $w$ selected in
step 2 will be rejected in step 3 of the algorithm. Because the
rejection process is independent of the $w \in W$ that has actually
been selected, $q_w$ may also be viewed as the {\em conditional\/}
probability that $w$ is rejected given that $w$ has been selected. The
next proposition shows that if the rejection probabilities $q_w$ are
identical for all $w \in W$, then ${\sc CARgen}^{**}$ gives rise to a
distribution satisfying CAR:
\pro
\label{pro:CARsensor}
${\sc CARgen}^{**}$ halts with probability 1 {\em \bf and\/} for all $w \in
W$, 
$\Pr^*(X_W = w) =
P_W(w)$ {\em \bf and\/} the distribution $\Pr^*$ satisfies CAR \\
\begin{centering}  
if and only if \\
\end{centering}
there  exists some $0 \leq  q < 1$ with for all $w \in W$, $q_w
= q$.
\epro
Henceforth we use the name ${\sc CARgen}^*$ to denote instances of
${\sc CARgen}^{**}$ with for all $w \in W$, $q_w
= q$ for some $0 \leq q < 1$.
Proposition~\ref{pro:CARsensor} shows that the distributions $\Pr^*$
corresponding to ${\sc CARgen}^*$ are
guaranteed to satisfy CAR. Theorem~\ref{thm:CARresolved} shows that {\em all\/} CAR distributions
can be represented as instances of ${\sc CARgen}^*$:
\thm
\label{thm:CARresolved}
Given any $O$ and $W$, let $\Pr$ be a distribution over $\R$ such that
CAR holds. Then $\Pr = \Pr^*$ where $\Pr^*$ arises from an instantiation of ${\sc
  sensor}^*$. 
\ethm
Together, Proposition~\ref{pro:CARsensor} and Theorem~\ref{thm:CARresolved}
show that ${\sc CARgen}^{*}$ precisely captures the CAR condition.

\xam (continuation of Example~\ref{xam:CARwithoutpartition})
Let $U_1, U_2, U_3, A_1, A_2$ and $A_3$ be as in 
Example~\ref{xam:CARwithoutpartition}. Consider the set of partitions 
$\P = \{\Pi_1,
\Pi_2, \Pi_3\}$ with $\Pi_i = \{ U_i, A_i \}$. Let $\Pr^*$ be determined by
an instance of ${\sc CARgen}^{**}$ with $P_\P(\Pi_1) = P_\P(\Pi_2) = P_\P(\Pi_3)
= 1/3$ and rejection probabilities 
$q_{U_i|\Pi_i} = 0$ and $q_{A_i|\Pi_i} = 1$. Then, for all $w \in A_i$, $q_w =
\sum_{\{U,s : w \in U, U \in s\}} P_\P(s) q_{U|\Pi}  = 1/3$. Therefore, we are
actually dealing with an instance of ${\sc CARgen}^*$ and by
Proposition~\ref{pro:CARsensor}, CAR must hold for $\Pr^*$. Indeed, direct
calculation of $\Pr^*$ shows that $\Pr^*(X_O \in O^*) = 1$ with $O^* =  \{U_1,U_2, U_3\}$
and, for $j=1..3$,  $\Pr^*(X_O = U_j \mid X_W = w) = 1/2$ for all $w \in U_j$. We see
that ${\sc CARgen}^*$ can account for such $\Pr^*$, whereas in
Example~\ref{xam:CARwithoutpartition} we showed that ${\sc CARgen}$ cannot.  
\exam
}
It is easy to see that {\sc CARgen} is the special case of {\sc
CARgen$^*$} where $q_{U|\Pi} = 0$ for all $(U,\Pi)$. 
Allowing $q_{U|\Pi} > 0$ gives us a little more flexibility.
To understand the role of the 
constraint (\ref{CARconstraint}), note that $q_{U|\Pi}$ is the
probability that the algorithm does not terminate at step 2.3, given
that $U$ and $\Pi$ are chosen at step 2.2.
It follows that the probability $q_w$ that a pair $(w,U)$ is not output
at step 2.3 for some $U$ is 
$$q_w  = \sum_{\{(U, \Pi): \; w \in U , \,  U \in \Pi\}} P_\P(\Pi)q_{U|\Pi}.
$$
Thus,  (\ref{CARconstraint}) says that the
probability $q_w$ that a pair whose first component is $w$ is not output
at step 2.3 is the same for all $w \in W$. 

{\sc CARgen$^*$} can  generate the CAR distribution in
Example~\ref{xam:CARwithoutpartition}, which could not be generated by
{\sc CARgen}.  To see this, using
the same notation as in the example, consider the set of partitions 
$\P = \{\Pi_1, \Pi_2, \Pi_3\}$ with $\Pi_i = \{ U_i, A_i \}$. Let
$P_\P(\Pi_1) = P_\P(\Pi_2) = P_\P(\Pi_3) = 1/3$, 
$q_{U_i|\Pi_i} = 0$, and $q_{A_i|\Pi_i} = 1$. 
It is easy to verify that for all $w \in W$, we have that 
$\sum_{\{U,\Pi : w \in U, U \in \Pi\}} P_\P(\Pi) q_{U|\Pi}  = 1/3$,
so that the constraint (\ref{CARconstraint}) is satisfied.
Moreover, direct calculation shows that,
for arbitrary $P_W$,
the distribution $\Pr^*$ on runs
generated by {\sc CARgen$^*$} with this choice of parameters is precisely
the unique distribution satisfying CAR in this case.

So why is {\sc CARgen$^*$} a legitimate mechanism?  The key point is
that all the relevant steps in the algorithm can be carried out by an
experimenter.  The
parameters $q$ and $q_{U|\Pi}$ for $\Pi \in \P$ and $U \in \Pi$ are
chosen before the algorithm begins; this can certainly be done by an
experimenter.  Similarly, it is straightforward to check that
the equation (\ref{CARconstraint}) holds for each $w \in W$.  As for the
algorithm itself, the
experimenter has no control over the choice of $w$; this is chosen
by nature according its distribution, $P_W$.  However, the experimenter
can perform steps 2.2 and 2.3, that is choosing $\Pi \in \P$ according
to the probability distribution $P_{\P}$, and rejecting 
the observation $U$ with probability $q_{U|\Pi}$ (since the experimenter
knows both the sensor chosen (i.e., $\Pi$) and the observation ($U$).

The following theorem shows that {\sc CARgen$^*$} does exactly what we want.
\thm\label{thm:CARresolved}
Given a set $\R$ of runs over a set
$W$ of worlds and a set $O$ of observations, $\Pr$ is a distribution on 
$\R$ that satisfies CAR if and only if there is a setting of the parameters in 
(step 1 of)
{\sc CARgen$^*$} such that, for all $w \in W$ and $U \in O$, 
$\Pr(\{r: X_W(r) = w, \, X_O(r) = U\})$ is
the probability that {\sc CARgen$^*$} returns $(w,U)$.
\ethm

\section{Beyond Observations of Events}\label{sec:MRE}
\subsection{Jeffrey Conditioning}
\label{sec:Jeffrey}
In the previous section, we assumed that the information received is of
the form ``the actual world is in $U$''.  But information does not
always come in such nice packages.  Perhaps the simplest generalization
of this is to assume that there is a partition $\{U_1, \ldots, U_n\}$ of
$W$ and the agent observes $\alpha_1 U_1; \ldots; \alpha_n
U_n$, where $\alpha_1 + \cdots + \cdots \alpha_n = 1$.  This is to be
interpreted as an observation that leads the agent to believe $U_j$ with
probability $\alpha_j$, for $j = 1, \ldots,n$.   According to Jeffrey
conditioning,
given a distribution $P_W$ on $W$, 
$$\begin{array}{ll}
&P_W(V \mid \alpha_1 U_1; \ldots; \alpha_n U_n) \\
= &\alpha_1 P_W(V\mid U_1) +
\cdots + \alpha_n P_W(V\mid U_n).
\end{array}$$
Jeffrey conditioning is defined only if $\alpha_i > 0$ implies that
$P_W(U_i) > 0$; if
$\alpha_i = 0$ and $P_W(U_i) = 0$, then $\alpha_i P_W(V \mid U_i)$ is
taken to be 0.
Clearly ordinary conditioning
is the special case of Jeffrey conditioning where $\alpha_i = 1$ for
some $i$ so, as is standard, we deliberately use the same notation for
updating using Jeffrey conditioning and ordinary conditioning.

We now want to determine when updating in the naive space using Jeffrey
conditioning is appropriate.  Thus, we assume that the agent's
observations now have the form of $\alpha_1 U_1; \ldots; \alpha_n U_n$
for some partition $\{U_1, \ldots, U_n\}$ of $W$.  (Different
observations may, in general, use different partitions.)  
Just as we did for the case that observations are events
(Section~\ref{sec:CAR}, first paragraph), we 
once again assume that the
agent's observations are accurate. What does that mean in the present
context?
We simply require that, conditional on
making the observation, the probability of $U_i$ really is $\alpha_i$
for $i = 1, \ldots, n$.  That is,
for $i = 1, \ldots, n$, we have
\begin{equation}\label{eq:accurate}
\Pr(X_W \in U_i \mid X_O = \alpha_1 U_1 ; \ldots; \alpha_n U_n) =
\alpha_i. 
\end{equation}
This clearly generalizes the requirement of accuracy given in the case
that the observations are events.

Not surprisingly, there is a generalization of the CAR condition that is
needed to guarantee that Jeffrey conditioning can be applied to the
naive space.

\thm\label{thm:JeffreyOK}
Fix 
a probability $\Pr$ on $\R$,
a partition $\{U_1, \ldots, U_n\}$ of $W$,
and probabilities $\alpha_1, \ldots, \alpha_n$ such that $\alpha_1 +
\cdots + \alpha_n = 1$.
Let $C$ be the observation $\alpha_1 U_1; \ldots ; \alpha_n U_n$.
Fix some $i \in \{1,\ldots, n\}$.
Then the following are equivalent:
\begin{itemize}
\item[(a)] If $\Pr(X_O = C) > 0$, then 
$\Pr(X_W = w \mid X_O = C) = \Pr_W(w \mid \alpha_1 U_1; \ldots;
\alpha_n U_n)$
for all $w \in U_i$.
\item[(b)] $\Pr(X_O = C \mid X_W = w) =
\Pr(X_O = C \mid X_W \in U_i)$ 
for all
$w \in U_i$ such that $\Pr(X_W = w) > 0$.
\end{itemize}
\ethm
Part~(b) of Theorem~\ref{thm:JeffreyOK} is analogous to part (c) of
Theorem~\ref{thm:CAR}.  There are a number of conditions equivalent to
(b) that we could have stated, similar in spirit to the conditions in 
Theorem~\ref{thm:CAR}. 
Note that these
are even more
stringent conditions than are required for 
ordinary
conditioning to be
appropriate.  

Examples~\ref{xam:CARhard1} and~\ref{xam:CARhard2} already suggest that
there are not too many nontrivial scenarios where
applying Jeffrey conditioning to the naive space is appropriate.
However, just as for the original CAR condition, there do
exist special situations in which generalized CAR is a realistic
assumption. For ordinary CAR, we mentioned 
the {\sc CARgen} mechanism (Section~\ref{sec:CARmechanics}).
For Jeffrey conditioning, a
similar mechanism may be a realistic model in some situations where
all observations refer to the same partition
$\{U_1, \ldots, U_n\}$ of $W$. We now describe a scenario for such a
situation.
Suppose $O$ consists of $k>1$
observations $C_1, \ldots, C_k$ with $C_i = \alpha_{i1} U_1 ; \ldots
; \alpha_{in} U_n$ such that all $\alpha_{ij} > 0$. Now, fix $n$
(arbitrary) conditional distributions 
$\Pr_j$, $j = 1, \ldots, n$, on $W$.  Intuitively, $\Pr_j$ is $\Pr_W(\cdot
\mid U_j)$. 
Consider the following mechanism: first an
observation $C_i$ is chosen (according to some distribution $P_O$ 
on $O$); then a
set $U_j$ is chosen 
with probability $\alpha_{ij}$ (i.e., 
according to the distribution induced by $C_i$);
finally, a world $w \in U_j$ is chosen
according to $\Pr_j$.   

If the observation $C_i$
and world $w$ are generated this way, then the generalized CAR
condition holds, that is, conditioning in the sophisticated space
coincides with Jeffrey conditioning: 

\pro
\label{pro:GCARcanhold}
Consider a partition $\{U_1, \ldots, U_n\}$ of $W$ and a set of 
$k > 1$
observations $O$ as above. 
For every distribution $P_O$ on $O$ with $P_O(C_i) > 0$ for all
$i \in \{1,\ldots, k\}$, 
there exists a distribution $\Pr$ on $\R$ such that 
$P_O = \Pr_O$ (i.e. $P_O$ is the marginal of $\Pr$ on $O$)
and $\Pr$ satisfies the generalized
CAR condition 
(Theorem~\ref{thm:JeffreyOK}(b))
for $U_1, \ldots, U_n$.
\epro
\commentout{
First, for $C \in O$ 
set $\Pr(X_O= C) = \Pr_O(C)$. Then set $\Pr(U_j  \mid  C_i) =
\alpha_{ij}$. This  
induces the marginal distribution $\Pr_U$ on $\{U_1, \ldots,
U_n\}$ satisfying
$$
\mbox{$\Pr_U$}(U_j) = \sum_{i=1..k} \mbox{$\Pr_O$}(C_i) \alpha_{ij}.
$$
Define, for $j=1..n, w \in U_j$,
$$\mbox{$\Pr_W$}(w) := \sum_i P(w \mid U_i)\Pr(U_i) = P(w \mid U_j)\Pr(U_j).$$
By construction, 
the generalized CAR condition
(Theorem~\ref{thm:JeffreyOK}(a)) holds for $\Pr$.  
}

Proposition \ref{pro:GCARcanhold} demonstrates that, even though the
analogue of the CAR condition expressed in Theorem~\ref{thm:JeffreyOK}
is hard to satisfy in general, at least if the set $\{U_1, \ldots,
U_n\}$ is the same for all observations, then 
for every such set of observations
there exist {\em some\/}
priors $\Pr$ on $\R$ for which the CAR-analogue {\em is\/} satisfied
for all observations.  
As we show next, for MRE updating, this is no longer the case.

\subsection{Minimum Relative Entropy Updating}
What about cases where the constraints are not in the special form where
Jeffrey's conditioning can be applied?  Perhaps the most common approach
in this case is to use MRE.  Given a constraint (where a constraint is
simply a set of probability distributions---intuitively, the
distributions satisfying the constraint) and a prior distribution 
$P_W$ on $W$,
the idea is to pick, among all  distributions satisfying the
constraint, the one that is ``closest'' to the prior distribution, where
the ``closeness'' of $P_W'$ to $P_W$ is measured using relative entropy.
The {\em relative entropy between $P_W'$ and $P_W$\/}
\cite{X.Entropy,CoverThomas}
is defined as
$$\sum_{w \in W} P_W'(w) \log\left(\frac{P_W'(w)}{P_W(w)}\right).$$
(The logarithm here is taken to the base 2; if $P_W'(w) = 0$ then
$P_W'(w) \log(P_W'(w)/P_W(w))$ is taken to be 0.  This is reasonable
since $\lim_{x \rightarrow 0} x \log(x/c) = 0$ if $c > 0$.) 
The relative entropy is finite provided that $P_W'$ is
{\em absolutely continuous\/} with respect to $P_W$, in that if 
$P_W(w) = 0$, then $P_W'(w) = 0$,
for all $w \in W$. Otherwise, it is defined to be infinite. 

The constraints we consider here are all closed and
convex sets of probability measures. In this case, it is known that
there is a unique distribution that satisfies the constraints and
minimizes the relative entropy.  Given 
a nonempty constraint
$C$ and a 
probability distribution $P_W$ on $W$, let $P_W(\cdot \mid C)$
denote the distribution that minimizes relative entropy with 
respect to $P_W$.

If the constraints have the form to
which Jeffrey's Rule is applicable, that is, if they have the form
$\{P_W': P_W'(U_i) = \alpha_i, i = 1, \ldots, n\}$ for some partition
$\{U_1, \ldots, U_n\}$, then it is well known that the distribution that
minimizes entropy relative to a prior $P_W$ is $P_W(\cdot \mid \alpha_1
U_1 ; \ldots; \alpha_n U_n)$
(see, e.g., \cite{DZ}).
Thus, MRE updating generalizes Jeffrey
conditioning (and hence also standard conditioning).

To study MRE updating in our framework, we assume that the
observations are now arbitrary closed convex 
constraints on the probability measure.  Again, we assume that the observations are accurate in that,
conditional on making the observation, the constraints hold.
For now, we  focus on the simplest 
possible case that  cannot be handled by Jeffrey updating. In this case,
constraints (observations) still have the
form $\alpha_1 U_1; \ldots; \alpha_n U_n$, but now the $U_i$'s do not
have to form a partition (they may overlap and/or not cover $W$)
and the $\alpha_i$ do not have to sum to 1.
Such an observation is accurate if it satisfies
(\ref{eq:accurate}), just as before.
We can now ask the same questions that we asked before about ordinary
conditioning and Jeffrey conditioning in the naive space.
\begin{enumerate}
\item Is there an alternative characterization of the conditions
  under which MRE updating coincides with conditioning in the
  sophisticated space? That is, are there analogues of
  Theorem~\ref{thm:CAR} and Theorem~\ref{thm:JeffreyOK}
for MRE updating?

\item Are there combinations of $O$ and $W$ for which it is
  not even possible that MRE can coincide with conditioning in the
  sophisticated space?
\end{enumerate}
With regard to 
question 1, it is easy to provide a counterexample showing
that there is no obvious analogue to Theorem~\ref{thm:JeffreyOK} for
MRE.  There is
a constraint $C$ such that the condition of part (a) of
Theorem~\ref{thm:JeffreyOK} holds for MRE updating whereas part (b) does
not hold.
(We omit the details here.)
Of course, it is possible that there are some quite different conditions
that characterize when MRE
updating coincides with conditioning in the sophisticated
space. However, even if they
exist, such conditions may be uninteresting in that they may hardly
ever apply. Indeed, as a partial answer to question 2, we now
introduce a very simple setting in which
MRE updating necessarily
leads to a result different from conditioning in the sophisticated
space. 

Let $U_1$ and $U_2$ be two subsets of $W$ such that 
$V_1 = U_1 - U_2$,  $V_2 = U_2 - U_1$,
$V_3 = U_1 \inter U_2$,  and $V_4 = W - (U_1 \cup U_2)$ are all
nonempty. 
Consider a constraint of the form
$C= \alpha_1 U_1 ; \alpha_2 U_2$,
where 
$\alpha_{1},\alpha_{2}$
are both 
in $(0,1)$. 
We investigate what happens if we use MRE
updating on $C$. Since $U_1$ and $U_2$ overlap and do not cover the space,
in general Jeffrey conditioning cannot be applied to update on $C$. 
There are some situations where, despite the overlap, Jeffrey
conditioning can essentially be applied.  
We say that  observation $C= \alpha_1 U_1 ; \alpha_2 U_2$ is 
{\em
Jeffrey-like\/}
iff, after MRE updating on 
one of the constraints $\alpha_1 U_1$ or $\alpha_2 U_2$, the other
constraint holds as well.  That is, $C$ is Jeffrey-like
(with respect to $P_W$)
if either $P_W(U_2\mid  \alpha_1 U_1) = \alpha_2$  
or $P_W(U_1 \mid  \alpha_2 U_2) = \alpha_1$. 
Suppose that $P_W(U_2\mid  \alpha_1 U_1) = \alpha_2$; then it is easy
to show that 
$P_W(\cdot \mid \alpha_1 U_1) = P_W( \cdot \mid  \alpha_1 U_1;
\alpha_2 U_2)$.

Intuitively, if the ``closest'' distribution  
$P_W'$
to $P_W$ that satisfies $P_W'(U_1) = \alpha_1$ also satisfies $P_W'(U_2)
= \alpha_2$, then $P_W'$ is the closest distribution to $P_W$ that
satisfies the constraint $C = \alpha_1 U_1 ; \alpha_2 U_2$.
Note that MRE updating on $\alpha U$ is equivalent to Jeffrey
conditioning on 
$\alpha U ; (1 - \alpha) (W - U)$.
Thus, if $C$ is 
Jeffrey-like,
then updating with $C$ is equivalent to Jeffrey updating.

\commentout{
\xam
\label{xam:noMRECAR}
We  now show that even in the simple setting introduced above,
an analogue to Theorem~\ref{thm:JeffreyOK} no longer holds.
Consider any distribution $P_W$ on $W$ with 
$P_W(w) > 0$ for all $w \in W$, and any constraint $C = \alpha_1
U_1 ; \alpha_2 U_2$ as above. 
We want a distribution $\Pr$ on runs such that
$\Pr_W(X_W = w) = P_W(w)$ and $\Pr_W(\cdot
| X_O =
C) = P_W( \cdot \mid C)$. 
To complete the definition of $\Pr$, we need
to define $\Pr_W(\cdot
\mid X_O =
C')$ for $C' \in O, C' \neq C$. 
It is clear that no matter how this is done, condition (a) of
Theorem~\ref{thm:JeffreyOK} holds for constraint $C$ and both $U_1$
and $U_2$. However, Proposition~\ref{pro:noMREchar} below shows that, unless
constraint $C$ is Jeffrey-like, condition (b) of
Theorem~\ref{thm:JeffreyOK} holds for neither $U_1$ nor
$U_2$. Therefore, Theorem~\ref{thm:JeffreyOK} is violated.
\exam

\pro
\label{pro:noMREchar}
Let $W = \{w_1, w_2, w_3\}, 
$V_i = \{ w_i \}$, and $U_i = W - V_i$, for $i = 1, 2, 3$.
Suppose that $\Pr_W(w) > 0$ for all $w \in W$ and let $\Pr_W(\cdot \mid X_O
= C)$ be constructed as above.  
If $C$ is not Jeffrey-like, then $\Pr(X_O = C \mid
X_W = w_1) \neq  \Pr(X_O = C \mid
X_W = w_3)$ and $\Pr(X_O = C \mid
X_W = w_2) \neq  \Pr(X_O = C \mid
X_W = w_3)$.
\epro

We now concentrate on a special case of the previous setting. Let
$U_1, U_2$, $V_1, V_2, V_3$ and $V_4$ be as before, and let $O = \{
C_1, C_2\}$, 
where 
\begin{eqnarray}
C_1 & = & \alpha_{11} U_1 ; \alpha_{12} U_2 \nonumber \\
C_2 & = & \alpha_{21} U_1 ; \alpha_{22} U_2. \nonumber
\end{eqnarray}
We assume that  
$\alpha_{11},\alpha_{12},\alpha_{21},\alpha_{22}$
are all 
in $(0,1)$ and are such that for both $i = 1,2$, some distribution on $W$
satisfying  $C_i$ exists.
Note that both $C_1$ and $C_2$ refer to the same
events $U_1$ and $U_2$. 
\xam
\label{xam:noMRECARb}
In Example \ref{xam:noMRECAR} we showed that
Theorem~\ref{thm:JeffreyOK} does not hold for MRE updating. But 
what about a modified version of
Theorem~\ref{thm:JeffreyOK}, in which all occurrences of sets $U_i$ are replaced
by atoms $V_i$? Since atoms by definition do not overlap, we may
suspect that such a modified version 
still holds for the type of constraints 
considered here. However, this is not the case. To see this, consider
the setting introduced above and let
$W = \{ w_1, \ldots, w_4 \}$ such that $V_i = \{ w_i \}$. In this
case, we trivially have for  $i=1..4$ and  $j = 1..2$ that
$$
\Pr(X_O = C_j \mid  X_W =w) = \Pr(X_O = C_j \mid  X_W \in V_i)
$$
for all $w \in V_i$.
So, no matter what the marginal distribution $\Pr_O$ over observations induced
by $\Pr$ is, the analogue of condition (b) of
Theorem~\ref{thm:JeffreyOK} holds for all combinations of 
$C_1, C_2$ and $U_1, U_2$. However,
Theorem~\ref{thm:MREnotOK} below implies that we can choose $C_1$ and
$C_2$ and $\Pr$ such that condition (a) of Theorem~\ref{thm:JeffreyOK}
does not hold. Therefore, the modified version of
Theorem~\ref{thm:MREnotOK} cannot hold either.
\exam
Together, Example~\ref{xam:noMRECAR} and~\ref{xam:noMRECARb} indicate 
that the condition $\Pr(X_W = w \mid X_O = C) =
\Pr_W(w \mid C)$ 
does not have a clear alternative characterization if
$C$ cannot be written as a Jeffrey constraint. Nevertheless, it could
still be possible that, in analogy to the mechanism {\sc CARgen} of
Section~\ref{sec:CARnothing} and the mechanism discussed before
Proposition~\ref{pro:GCARcanhold}, MRE coincides with conditioning in the
sophisticated space in some interesting cases corresponding to some
physical mechanisms. 
Theorem~\ref{thm:MREnotOK} below strongly suggests that this
is not so: it shows that even in the very simple setting
introduced above,  if $C$
is not 
Jeffrey-like, then there may be no distribution $\Pr$ on $\R$
such that MRE updating coincides with conditioning in the
sophisticated space; thus, there can be no equivalent to the CAR condition.
}
\commentout{

We say that  observation $C= \alpha_1 U_1 ; \alpha_2 U_2$ is 
{\em
Jeffrey-like\/}
iff, when MRE updating on 
one of the constraints $\alpha_1 U_1$ or $\alpha_2 U_2$, the other
constraint holds as well.  That is, $C$ is Jeffrey-like
(with respect to $\Pr_W$)
if either $\Pr_W(U_2\mid  \alpha_1 U_1) = \alpha_2$  
or $\Pr_W(U_1 \mid  \alpha_2 U_2) = \alpha_1$. 
Suppose that $\Pr_W(U_2\mid  \alpha_1 U_1) = \alpha_2$; then it is easy
to show that 
$\Pr_W(\cdot \mid \alpha_1 U_1) = \Pr_W( \cdot \mid  \alpha_1 U_1;
\alpha_2 U_2)$.

Intuitively, if the ``closest'' distribution  
$\Pr$
to $\Pr_W$ that satisfies $\Pr(U_1) = \alpha_1$ also satisfies $\Pr(U_2)
= \alpha_2$, then $\Pr$ is the closest distribution to $\Pr_W$ that
satisfies the constraint $C = \alpha_1 U_1 ; \alpha_2 U_2$.
Note that MRE updating on $\alpha U$ is equivalent to Jeffrey
conditioning on 
$\alpha U ; (1 - \alpha) (W - U)$.
Thus, if $C$ is 
Jeffrey-like,
then updating with $C$ is equivalent to Jeffrey updating.

To demonstrate this, we focus on the simplest possible case. 
Let $O$ consist of two observations (constraints),
$C_i = \alpha_{i1} U_1; \alpha_{i2} U_2$, $i = 1,2$, 
where $U_1 - U_2$, 
$U_1 \inter U_2$,  $U_2 - U_1$ and $W - (U_1 \cup U_2)$ are all
nonempty. We further assume that 
$\alpha_{11},\alpha_{12},\alpha_{21},\alpha_{22}$
are all 
in $(0,1)$
, and are such that for both $i = 1,2$, some distribution on $W$
satisfying  $C_i$ exists.
Note that both $C_1$ and $C_2$ refer to the same
events $U_1$ and $U_2$. 

We say that  observation $C= \alpha_1 U_1 ; \alpha_2 U_2$ is 
{\em
Jeffrey-like\/}
iff, when MRE updating on 
one of the constraints $\alpha_1 U_1$ or $\alpha_2 U_2$, the other
constraint holds as well.  That is, $C$ is Jeffrey-like
(with respect to $\Pr_W$)
if either $\Pr_W(U_2\mid  \alpha_1 U_1) = \alpha_2$  
or $\Pr_W(U_1 \mid  \alpha_2 U_2) = \alpha_1$. 
Suppose that $\Pr_W(U_2\mid  \alpha_1 U_1) = \alpha_2$; then it is easy
to show that 
$\Pr_W(\cdot \mid \alpha_1 U_1) = \Pr_W( \cdot \mid  \alpha_1 U_1;
\alpha_2 U_2)$.

Intuitively, if the ``closest'' distribution  
$\Pr$
to $\Pr_W$ that satisfies $\Pr(U_1) = \alpha_1$ also satisfies $\Pr(U_2)
= \alpha_2$, then $\Pr$ is the closest distribution to $\Pr_W$ that
satisfies the constraint $C = \alpha_1 U_1 ; \alpha_2 U_2$.
Note that MRE updating on $\alpha U$ is equivalent to Jeffrey
conditioning on 
$\alpha U ; (1 - \alpha) (W - U)$.
Thus, if $C$ is 
Jeffrey-like,
then updating with $C$ is equivalent to Jeffrey updating.

The following theorem
shows that, in general, if $C$
is not 
Jeffrey-like, then there may be no distribution $\Pr$ on $\R$
such that MRE updating coincides with conditioning in the
sophisticated space; thus, there can be no equivalent to the CAR condition.
}
\thm
\label{thm:MREnotOK}
Given a set $\R$ of runs and a set $O = \{C_1, C_2\}$ of observations,
where $C_i = \alpha_{i1} U_1; \alpha_{i2} U_2$, for $i = 1,2$,
let $\Pr$ be a distribution on $\R$ such that
$\Pr(X_O = C_1)$, $\Pr(X_O = C_2) > 0$,
and $\Pr_W(w) = \Pr(X_W = w) > 0$ for all $w \in W$. 
Let $\Pr^i = \Pr(\cdot \mid X_O = C_i )$, and let 
$\Pr^i_W$ be the marginal of $\Pr^i$ on $W$. 
If either $C_1$ or $C_2$ is not 
Jeffrey-like,
then we cannot have $\Pr^i_W = \Pr_W(\cdot \mid  C_i)$,
for  both $i = 1, 2$. 
\commentout{
Given a set $\R$ of runs and a set $O = \{C_1, C_2\}$ of observations,
where $C_i = \alpha_{i1} U_1; \alpha_{i2} U_2$, for $i = 1,2$,
let $\Pr$ be a distribution on $\R$ such that
$\Pr(X_O = C_1)$, $\Pr(X_O = C_2) > 0$,
and $\Pr_W(w) > 0$ for all $w \in W$. 
Let $\Pr^i = \Pr(\cdot \mid \R[\<C_i\>])$, and let 
$\Pr^i_W$ be the marginal of $\Pr^i$ on $W$. 
If either $C_1$ or $C_2$ is not 
Jeffrey-like,
then we cannot have $\Pr^i_W = \Pr_W(\cdot \mid  C_i)$,
for  both $i = 1, 2$. 
}
\ethm
\commentout{To get a better feel for Theorem~\ref{thm:MREnotOK},
note that it follows from standard properties of MRE \cite{Csiszar91} that 
(in the notation of Theorem~\ref{thm:MREnotOK}, and using 
$\Pr_{C_i}$
as an abbreviation for the distribution
$\Pr_W(\cdot \mid C_i)$), conditional  
on each element of the set ${\cal V} = \{U_1 - U_2, U_1 \inter U_2, U_2
- U_1, W - (U_1 \cup U_2) \}$, 
$\Pr_{C_i}$ and $\Pr^i_W$ agree; that is,
$\Pr_{C_i}(\cdot \mid V) = \Pr^i_W(\cdot 
\mid V)$ for $V \in {\cal V}$. This
means that for $\Pr_{C_i}$ to be equal to $\Pr^i_W$ we need a ``CAR-like''
condition to hold for each $V \in {\cal V}$. That is, we need to have
$\Pr(X_O = C_i  \mid X_W = w) = \Pr(X_O = C_i \mid X_W \in V)$ for all $w
\in V$ such that $\Pr(X_W = w) > 0$, for all $V \in {\cal V}$. Theorem
\ref{thm:MREnotOK} says that, except for 
 ``singular'' cases, such a condition cannot hold.
}
For fixed $U_1$ and $U_2$, we can identify an observation $\alpha_1 U_1
; \alpha_2 U_2$ with the pair $(\alpha_1,\alpha_2) \in (0,1)^2$.
Under our conditions on 
$U_1$ and $U_2$,
the set of all 
Jeffrey-like
observations is a subset of $0$ (Lebesgue) measure of
this set.
Thus, the set of observations for which MRE conditioning corresponds to
conditioning in the sophisticated space is a (Lebesgue) measure 0 set in
the space of possible observations.
Note however, that this set depends on the prior $P_W$ over $W$.

A result similar to Theorem~\ref{thm:MREnotOK} 
was proved by Seidenfeld~\citeyear{Seidenfeld86} (and
considerably generalized in \cite{Dawid01}). Seidenfeld 
shows that, under very weak conditions,
MRE updating cannot coincide with sophisticated conditioning 
 if the observations have the form ``the
conditional probability of $U$ given $V$ is $\alpha$'' (as is the case
in the Judy Benjamin problem).  
Theorem~\ref{thm:MREnotOK} shows that this
is impossible even for observations of
the much simpler form $\alpha_1 U_1; \alpha_2 U_2$, unless 
we can reduce the problem to Jeffrey conditioning (in which case
Theorem~\ref{thm:JeffreyOK} applies).
\commentout{
Theorem~\ref{thm:MREnotOK} stands in marked contrast to Proposition~\ref{pro:GCARcanhold} below. 
\pro
\label{pro:GCARcanhold}
Fix a partition $U_1, \ldots, U_n$ of $W$. Let the set of observations $O$ consist of $k > 1$ observations
$C_1, \ldots, C_k$ with $C_i = \alpha_{i1} U_1 ; \ldots ; \alpha_{in}
U_n$ such that all $\alpha_{ij} > 0$. 
For every distribution $\Pr_O$ on $O$ with $\Pr_O(C_i) > 0$ for all
$i \in \{1,\ldots, k\}$, 
there exists a distribution $\Pr$ on $\R$ such that 
$\Pr_O$ is the marginal of $\Pr$ on $O$
and $\Pr$ satisfies the generalized
CAR condition (i.e., $(\Pr(\cdot \mid  \R[\< C_i\>]))_W = \Pr_W(\cdot \mid C_i )$) for all $i \in \{1,\ldots,k\}$.
\epro
The proposition expresses the following. 
Even though the analogue of the CAR condition expressed in
Theorem~\ref{thm:JeffreyOK} is hard to satisfy,
at least for all finite sets $O$ all referring to 
an (arbitrary)
fixed partition $U_1, \ldots, U_n$, there exists some prior $\Pr$ on $\R$
for which the CAR-analogue {\em is\/} satisfied for all observations. 
In case that the $U_i$ do not form a partition, Theorem~\ref{thm:MREnotOK} shows
that in general, no such priors need exist; in such cases, MRE coincides with conditioning in the sophisticated
space for no observation $C \in O$ under no prior 
$\Pr$ whatsoever. Therefore, in such cases there can be no equivalent to 
the CAR condition.
Theorem~\ref{thm:MREnotOK} shows that, in the case where only two
observations are possible,
MRE cannot coincide with conditioning in the sophisticated
space unless both observations are Jeffrey-like.
}
\commentout{Below we extend this theorem to a more general case, involving an
arbitrary number of $k$ observations and an arbitrary number of $n$
subsets $U_1, \ldots, U_n$. For this we need to define an extension
of the concept of ``Jeffrey-like'' that we will call ``singular''; the
reasons for adapting different technology will become apparent later. 
In the general case
there may be  some very special 
non-singular
combinations of priors $\Pr$ and
observations such that
MRE updating corresponds to conditioning in the sophisticated
space. However, in marked contrast to the case for Jeffrey
conditioning, these remain isolated cases. More specifically,
Proposition~\ref{pro:GCARcanhold} 
shows that, given
an arbitrary set $O$ of observations to which Jeffrey conditioning can
apply, where all the observations in $O$ refer to the same events, 
and a distribution $\Pr_O$ on $O$, 
we can {\em always\/} construct {\em some\/}
distribution $\Pr$ on $\R$ such that $\Pr(X_O = C) =
\Pr_O(C)$ for all $C \in O$ and 
$\Pr$ satisfies the generalized CAR condition. 
Theorem~\ref{thm:CARalmostnever}
shows that, if the $U_i$ are allowed to overlap 
and/or not cover the set of worlds $W$,
this is 
typically not possible. Here `typically not possible' means that for almost all
distributions $\Pr_O$ over observations, no non-singular distribution over $\R$
exists such that (a) its marginal over $O$ coincides with $\Pr_O$ and
(b) MRE updating on $C$ coincides with conditioning on $X_O = C$ for
all $C \in O$.

We first need some terminology. Let $W$ be given.
Let $C = \alpha_1 U_1 ; \ldots; \alpha_n U_n$ be a constraint. We say
that $C$ is {\em singular\/} (relative to 
some given $\Pr_W$) if there exists $i \in \{1, \ldots, n \}$ such
that $\Pr_W(U_i \mid \alpha_1 U_1 ; \ldots; \alpha_{i-1} U_{i-1} ;
\alpha_{i+1} U_{i+1} ; \ldots; \alpha_n U_n) = \alpha_i$. Note that,
if $i=2$, then `singular' means exactly `Jeffrey-like'. For $n > 2$,
MRE updating on a `singular' constraint need not coincide with
successive Jeffrey updating; it only means that MRE updating on the
constraint  is equivalent to MRE updating on a smaller constraint and
then Jeffrey updating.

For given $O = \{C_1,\ldots,C_k\}$, we 
say $(\lambda_1,\ldots, \lambda_k)$ is {\em  CAR-compatible\/} iff there
exists a distribution $\Pr$ on $\R$ with $\Pr(X_O = C_i) = 
\lambda_i$ such that the generalized
CAR condition holds.
We let $\Delta^k$ stand for the unit simplex in ${\bf R}^k$. 
For given
sets $U_1, \ldots, U_n$, $U_i \subseteq W$ and $w \in W$, we define 
$$
{\bf U}(w) = \{ U_i \mid w \in U_i \}. $$
\thm
\label{thm:CARalmostnever}
Fix any set $\{U_1, \ldots, U_n\}$ of
subsets of $W$ such that for some $w_0,w_1, U_j$, 
${\bf U}(w_1) = {\bf U}(w_0) \cup \{ U_j \}$. 
Suppose that there are $k> 1$ possible observations,
$C_1, \ldots, C_k$, with $C_i = \alpha_{i1} U_1 ; \ldots ; \alpha_{in}
U_n$, such that all 
$\alpha_{ij} \in (0,1)$, and for all $i$ a distribution satisfying
$C_i$ exists. 
Define $$S = \{ (\lambda_1, \ldots, \lambda_k) \in \Delta^k :
\mbox{$\lambda$ is CAR-compatible.} \}.$$
Then either all of  the $C_i$ are singular or
$S$ must be a subset of $\Delta^k$ of Lebesgue measure $0$.
\ethm

\commentout{
\prf 
As in the proof of Theorem~\ref{thm:MREnotOK}, it follows from
\cite[Theorem 3]{Csiszar75} that for each $C_i$, for all $w \in W$,
\begin{equation}
\label{eq:hidehoc}
\mbox{$\Pr_W$}(w \mid C_i) = \frac{1}{Z_i}\exp(\sum_{j=1}^n \beta_{ij}
{\bf 1}_{w \in U_j})
 \mbox{$\Pr_W$}(w).
\end{equation}
Without loss of generality, let $U_1$ be such that  ${\bf U}(w_1) =
{\bf U}(w_0) \cup \{ U_1 \}$.
In order to have
 $\Pr^i_W = \Pr_W(\cdot \mid  C_i)$ for $i = 1..k$, we must have 
\begin{equation}
\label{eq:ciao}
\sum_{i=1}^k \lambda_i \mbox{$\Pr_W$}(w \mid C_i) =
 \mbox{$\Pr_W$}(w)
\end{equation} 
for all $w \in  W$, in particular for $w \in \{w_0,w_1 \}$. By
(\ref{eq:hidehoc}) the corresponding instances of (\ref{eq:ciao})
become
\begin{eqnarray}
\label{eq:hullob}
\sum_{i=1}^k \lambda_i f_i 
 \mbox{$\Pr_W$}(w_0)  & = & \mbox{$\Pr_W$}(w_0) \\
\label{eq:hulloa}
 \sum_{i=1}^k \lambda_i f_i  \exp(\beta_{i1})
 \mbox{$\Pr_W$}(w_1)  & = & \mbox{$\Pr_W$}(w_1) 
\end{eqnarray}
where $f_i = Z_i^{-1} \exp(\sum_{j: w_0 \in U_j} \beta_j)$.
We must also have
\begin{equation}
\label{eq:hulloc}
\sum_{i=1}^k \lambda_i = 1.
\end{equation}
If the $f_i$'s are not all identical, then (\ref{eq:hullob}) and
(\ref{eq:hulloc}) are linearly independent. Then the set of
$(\lambda_1,\ldots, \lambda_k)$ satisfying both these equations must have
dimension $k-2$, and we are done. If the $f_i$ are all identical and
not equal to $1$, there is no $(\lambda_1,\ldots, \lambda_k)$
satisfying both equations and we are done. The only remaining
possibility is that all $f_i$ are equal to $1$. In that case, if
furthermore not all $\beta_{i1}$ are identical,   (\ref{eq:hulloa}) and
(\ref{eq:hullob}) are linearly independent and  we are done. If all
$\beta_{i1}$ are identical and not equal to $0$, there is no  $(\lambda_1,\ldots, \lambda_k)$
satisfying both equations and we are done. If all 
$\beta_{i1}$ are identical and equal to $0$, then all $C_i$ must be
Jeffrey-like by the Claim inside the proof of 
Theorem~\ref{thm:MREnotOK}.
\eprf
}

The condition ${\bf U}(w_1) = {\bf U}(w_0) \cup \{ U_j \}$ may apply
in two different cases. If ${\bf U}(w_0) = \emptyset$, 
then we must have that the $U_1,
\ldots, U_n$  do not cover the space. If ${\bf U}(w_0)$ is nonempty,
then we must have that the $U_1,
\ldots, U_n$ overlap, since $w_1$ is contained in at least two sets:
$w_1$ is in  all
$U \in {\bf U}(w_0)$ and in $U_j$.

\commentout{Proposition~\ref{pro:CARalmostnever} says that, as long
Proposition~\ref{pro:CARalmostnever} says that for all sets of
observations $O$ of a certain kind, {\em almost all\/} 
(that is, a measure 1 subset of)
priors on observations are such
that CAR cannot hold. 
To compare this with 
Theorem~\ref{thm:MREnotOK}, note first that
whether a constraint $C_i = \alpha_{i1} U_1 ; \alpha_{i2} U_2$ 
is Jeffrey-like depends not only on
the $\alpha_{ij}$ but also on the marginal prior distribution
$\Pr_W$ on $W$. Theorem~\ref{thm:MREnotOK} says
that for all sets of observations of a certain kind,  
{\em all\/} priors on worlds are such that for {\em almost all
observations}, CAR cannot hold.
}
}
\section{Discussion}\label{sec:discussion}
We have studied the circumstances under which
ordinary 
conditioning, Jeffrey conditioning, and
MRE updating 
in a naive space
can be justified, 
where ``justified'' for us means 
``agrees with conditioning in the
sophisticated space''.
The main message of this paper is that, except for quite special cases,
the three methods cannot be justified. 
Figure~\ref{fig:overview} summarizes the main insights of this paper in
more detail. 
As we mentioned in the introduction,
the idea of comparing an update rule in a ``naive space'' with
conditioning in a ``sophisticated space'' 
is not new; it 
appears in the CAR literature and the MRE
literature
(as well as in papers such as \cite{HT} 
and \cite{DawidD77}).
In addition to bringing these two strands of research together, our own
contributions are the following:  
(a) we show that the CAR framework can be used as a general tool to
clarify many of the well-known paradoxes of conditional  
probability;
(b) we give a general characterization of CAR in terms of a
binary-valued matrix, showing that in many realistic scenarios, the
CAR condition {\em cannot\/} hold (Theorem~\ref{thm:charCAR});
(c) we define a mechanism ${\sc CARgen}^{*}$ that
generates all and only distributions satisfying CAR
(Theorem~\ref{thm:CARresolved});
(d) we show that the CAR condition has a natural extension to cases
where Jeffrey conditioning can be applied
(Theorem~\ref{thm:JeffreyOK}); 
and (e) we show that no CAR-like condition
can 
hold
in general for cases where only MRE (and not Jeffrey)
updating can be applied (Theorem~\ref{thm:MREnotOK}).
\begin{figure*}
\begin{center}
\begin{tabular}{|p{2. cm}|p{3.5 cm}|p{3.0 cm}|p{4.7 cm}|}
\hline
{\em observation type \/} & {\em set of observations $O$ \/} & {\em
  simplest applicable update rule\/} & 
{\em when it coincides with sophisticated conditioning \/}\\
\hline
event & pairwise disjoint & naive conditioning & 
always (Proposition~\ref{pro:CARmusthold})
\\ \hline
event & arbitrary set of events  & naive conditioning & 
\mbox{iff CAR holds}
(Theorem~\ref{thm:CAR})\\
\hline 
probability vector & probabilities of partition  & Jeffrey conditioning
& iff generalization of CAR holds (Theorem~\ref{thm:JeffreyOK})  
\\ \hline
probability vector & probabilities of 
two
overlapping sets  & MRE & if both observations Jeffrey-like
(Theorem~\ref{thm:MREnotOK})
\\ \hline
\end{tabular}
\end{center}
\caption{\label{fig:overview} Conditions under which updating in the naive space 
coincides with
  conditioning in the sophisticated space.}
\end{figure*}

Our results suggest that working in the naive space is rather
problematic.  On the other hand, as we observed in the introduction,
working in the sophisticated space (even assuming it can be constructed)
is problematic too.   So what are the alternatives?

For one thing, it is worth observing that MRE updating is not always so
bad.  In many successful practical applications, the ``constraint'' on
which to update is of the form
$
\frac{1}{n}\sum_{i=1}^{n} X_i = t
$
for some large $n$, where $X_i$ is the $i$th outcome of a random
variable $X$ on $W$.
That is, we observe an empirical average of outcomes of $X$.
In such a case, the MRE distribution is ``close'' (in the
appropriate distance measure) to the distribution we arrive at by
sophisticated conditioning. That is, if $\Pr'' = \Pr_W(\cdot \mid E(X) =
t)$,  
$\Pr' = \Pr( \cdot \mid X_O = <\frac{1}{n}\sum_{i=1}^{n} X_i =
t\>)$, 
and $Q^n$ denotes the $n$-fold product of a
probability distribution $Q$, then for sufficiently large $n$, we have
that
$(\Pr'')^n \approx (\Pr'_W)^n$
\cite{CampenhoutC81,Grunwald01a,Skyrms85,Uffink96}.
Thus, in such cases MRE (almost) coincides with sophisticated
conditioning after all.
(See \cite{Dawid01} for a discussion of how this result can be
reconciled with the results of Section~\ref{sec:MRE}.)

But when this special situation does not apply, it is worth asking whether
there exists an approach for updating in the naive space that can be
easily applied in practical situations, yet leads to {\em better}, in
some formally provable sense, updated distributions than the methods we
have considered?  A very interesting
candidate, often informally applied by human agents, is to simply {\em
ignore\/} the available extra information. It turns out that in many
situations this update rule behaves better, in a precise sense, than the
three
methods we have considered.  This will be explored in future work.

Our discussion here has focused completely on the probabilistic case.
However, these questions also make sense for other representations of
uncertainty.  Interestingly, in \cite{FrH2full}, it is shown that 
AGM-style belief revision \cite{agm:85} can be represented in terms of
conditioning using a qualitative representation of uncertainty called a
{\em plausibility measure}; to do this, the plausibility measure must
satisfy the analogue of Theorem~\ref{thm:CAR}(a), so that observations
carry no more information than the fact that they are true.  No CAR-like
condition is given to guarantee that this condition holds for
plausibility measures though.  It would be interesting to know if there
are analogues to CAR for other representations of uncertainty, such as
{\em possibility measures\/} \cite{DuboisPrade88} or {\em belief
functions\/} \cite{Shaf}.

\subsubsection*{Acknowledgments} 
A preliminary  version of this paper 
appears in {\em Uncertainty in Artificial Intelligence, Proceedings of
the Eighteenth Conference}, 2002.
We thank the referees of 
the UAI submission 
and the {\em JAIR\/} submission
for their perceptive comments.  
The first author was supported by a travel grant awarded by the
Netherlands Organization for Scientific Research (NWO).
The second author
was supported in part by NSF under grants
IIS-0090145
and CTC-0208535,
by ONR under grants N00014-00-1-0341,
N00014-01-1-0511,  and N00014-02-1-0455,
by the DoD Multidisciplinary University Research
Initiative (MURI) program administered by the ONR under
grants N00014-97-0505 and N00014-01-1-0795, and by a 
Guggenheim Fellowship, a Fulbright Fellowship, and a grant from the NWO.
Sabbatical support from CWI and
the Hebrew 
University of Jerusalem is also gratefully acknowledged.

\appendix
\section{Proofs}
In this section, we provide the proofs of all the results in the paper.
For convenience, we restate the results here.

\othm{thm:CAR} 
Fix a probability $\Pr$ on $\R$ and a set
$U \subseteq W$.
The following 
are equivalent:
\begin{itemize}
\item[(a)] 
If $\Pr(X_O = U) > 0$, then 
$\Pr(X_W =w \mid X_O = U) = \Pr(X_W = w \mid X_W \in U)$ for all
$w \in U$.
\item[(b)]
The event
$X_W = w$ is independent of 
the event
$X_O = U$ 
given $X_W \in U$, 
for all $w \in U$.
\item[(c)] $\Pr(X_O = U \mid X_W = w) = \Pr(X_O = U \mid X_W \in U)$ for all
$w \in U$ such that $\Pr(X_W = w) > 0$.
\item[(d)] $\Pr(X_O = U \mid X_W = w) = \Pr(X_O = U \mid X_W = w')$ for all
$w, w' \in U$ such that $\Pr(X_W = w) > 0$ and $\Pr(X_W = w') > 0$.
\end{itemize}
\eothm
\prf 
Suppose (a) holds.  We want to show that $X_W = w$ and $X_O = U$
are independent, for all $w \in U$.  Fix $w \in U$.  If $\Pr(X_O = U) =
0$ then the events are trivially independent.  So suppose that $\Pr(X_O
= U) > 0$.  Clearly 
$$\Pr(X_W = w \mid X_O = U \inter X_W \in U) = 
\Pr(X_W = w \mid X_O = U)$$ (since observing $U$ implies that the true
world is in U).  
By part (a), 
$$\Pr(X_W = w \mid X_O = U) = \Pr(X_W = w \mid
X_W \in U).$$   Thus, 
$$\Pr(X_W = w \mid X_U = U \inter X_W \in U) = 
\Pr(X_W = w \mid X_W \in U),$$ showing that $X_W = w$ is independent of
$X_O = U$, given $X_W \in U$.

Next suppose that (b) holds, and $w \in U$ is such that $\Pr(X_W = w) >
0$.  From part (b) it is immediate that  $\Pr(X_O = U
\mid X_W = w \inter X_W \in U) = \Pr(X_O = U \mid X_W \in U)$. 
Moreover, since $w \in U$, clearly $\Pr(X_O = U
\mid X_W = w \inter X_W \in U) = \Pr(X_O = U \mid X_W = w)$.  Part (c)
now follows.

Clearly (d) follows immediately from (c).  Thus, it remains to show that
(a) follows from (d).  We do this by showing that (d) implies (c) and
that (c) implies (a).  So suppose that (d) holds.  Suppose that
$\Pr(X_O = U \mid X_W = w) = a$ for all $w \in U$ such that $\Pr(X_W =
w) > 0$.  From the definition of conditional probability  
$$\begin{array}{ll}
&\Pr(X_O = U \mid X_W \in U) \\
= &\sum_{\{w \in U: \Pr(X_W = w) > 0\}} \Pr(X_O = U \inter X_W =
w)/\Pr(X_W \in U)\\
= &\sum_{\{w \in U: \Pr(X_W = w) > 0\}} \Pr(X_O = U \mid X_W = w)
\Pr(X_W = w)/\Pr(X_W \in U)\\
= &\sum_{\{w \in U: \Pr(X_W = w) > 0\}} a \Pr(X_W = w)/\Pr(X_W \in U)\\
= &a
\end{array}
$$
Thus, (c) follows from (d).

Finally, to see that (a) follows from (c), suppose that (c) holds.
If $w \in U$ is such
that $\Pr(X_W = w) = 0$, then (a) is immediate, so suppose that $\Pr(X_W
= w) > 0$.  
Then, using (c) and the fact that $X_O = U \subseteq X_W \in U$, we have that
$$\begin{array}{ll}
&\Pr(X_W = w \mid X_O = U) \\
= & \Pr(X_O = U \mid X_W = w) \Pr(X_W = w) /\Pr(X_O = U)\\
= & \Pr(X_O = U \mid X_W \in U) \Pr(X_W = w)/ \Pr(X_O = U)\\
= & \Pr(X_O = U \inter X_W \in U) \Pr(X_W = w)/ \Pr(X_W \in U) \Pr(X_O
= U)\\ 
 = & \Pr(X_O = U ) \Pr(X_W = w)/ \Pr(X_W \in U) \Pr(X_O = U)\\
= & \Pr(X_W = w)/ \Pr(X_W \in U)\\
= & \Pr(X= w \mid X_W \in U),
\end{array}
$$
as desired. \eprf
\medskip

\opro{pro:CARmusthold}
The CAR condition holds for all distributions $\Pr$ on $\R$ if
and only if $O$ consists of pairwise disjoint subsets of $W$. 
\eopro

\prf 
First suppose that the sets in $O$ are
pairwise disjoint. 
Then for each probability distribution $\Pr$ on $\R$, each $U \in O$,
and each world $w \in U$ such that $\Pr(X_W = w) > 0$, it must
be the case that $\Pr(X_O = U \mid X_W = w) = 1$.  Thus, part (d) of
Theorem~\ref{thm:CAR} applies.

For the converse, suppose that the sets in $O$ are not pairwise disjoint. Then
there exist sets $U, U' \in O$ such that both $U - U'$ and $ U
\cap U'$ are nonempty. Let $w_0 \in U \cap U'$. Clearly there 
exists a distribution $\Pr$ on $\R$ such that $\Pr(X_O = U) > 0$,
$\Pr(X_O = U') > 
0$, $\Pr(X_W = w_0 \mid X_O = U) = 0, \Pr(X_W = w_0 \mid X_O = U') > 0$.
But then $\Pr(X_W= w_0 \mid X_W \in U) > 0$.
Thus
$$\Pr(X_W = w_0 \mid X_O = U) \neq \Pr(X_W= w_0 \mid X_W \in U),$$
and the CAR condition (part (a) of Theorem~\ref{thm:CAR}) is violated.
\eprf

\medskip
\olem{lem:precharCAR}
Let $\R$ be the set of runs over observations $O$ and worlds $W$, and
let $S$ be the CARacterizing matrix for $O$ and $W$.
\begin{itemize}
\item[(a)]
Let $\Pr$ be any distribution over $\R$ and let 
$S'$ be the matrix obtained by deleting from $S$ all rows
corresponding to an atom $A$ with $\Pr(X_W \in A) = 0$. Define 
the vector
$\vec{\gamma} = (\gamma_1, \ldots, \gamma_n)$ by setting 
$\gamma_j = \Pr(X_O = U_j \mid X_W \in U_j)$
if $\Pr(X_W \in U_j) > 0$, and $\gamma_j = 0$ otherwise,
for $j = 1, \ldots, n$.
If $\Pr$
satisfies CAR, then $S' \cdot \vec{\gamma}^T = \vec{1}^T$.
\item[(b)]
Let $S'$ be a matrix consisting of 
a subset of the rows
of $S$, and let ${\cal P}_{W,S'}$ be the set of distributions over $W$
with support corresponding to $S'$; i.e., 
$${\cal P}_{W,S'} = \{ P_W \mid P_W(A) > 0 \ \mbox{iff $A$
  corresponds to a row in $S'$} \}.
$$ 
If there exists a vector $\vec{\gamma} \ge \vec{0}$
such that $S' \cdot \vec{\gamma}^T = \vec{1}^T$,
then, for {\em all\/} $P_W \in {\cal P}_{W,S'}$, there exists a
distribution $\Pr$ over $\R$ with $\Pr_W = P_W$ (i.e., the marginal of
$\Pr$ on $W$ is 
$P_W$) such that (a) $\Pr$ satisfies CAR and  (b)
$\Pr(X_O = U_j \mid X_W \in U_j) = \gamma_j$ for all $j$ with
$\Pr(X_W \in U_j) > 0$.
\end{itemize}
\eolem

\prf
For part (a), suppose that $\Pr$ is a distribution on $\R$
that satisfies CAR. 
Let $k$ be the number of rows in $S'$, and
let $\alpha_i = \Pr(X_W \in A_i)$, for $i = 1, \ldots, k$, where $A_i$
is the atom corresponding to the $i$th row of $S'$. 
\commentout{
Let
$\beta_j = \Pr(X_O = U_j)$, for $j = 1, \ldots, n$.
Note that if $\Pr(X_W \in U_j) > 0$, then
\begin{equation}
\label{eq:betachar}
\beta_j = \Pr(X_O = U_j) = \Pr(X_O = U_j \mid X_W \in U_j) \Pr(X_W \in
U_j).
\end{equation} 
Define $\vec{\gamma}$ as in the statement of part (a).
If  $\Pr(X_W \in U_j) > 0$, then
$\Pr(X_W \in U_j) = \sum_{\{i: A_i \subseteq U_j\}} \alpha_i$,
so that, by (\ref{eq:betachar}),
\begin{equation}
\label{eq:gammab}
\gamma_j = \Pr(X_O = U_j \mid X_W \in U_j) = \frac{\beta_j}{\sum_{\{i: A_i \subseteq
    U_j\}} \alpha_i}.
\end{equation}
(We remark that (\ref{eq:gammab}) was also derived in
\cite{GillLR97}, but the  further developments in the proof are
original.)
}
Note that 
$\alpha_i > 0$ for $i = 1, \ldots, k$. Clearly,
\begin{equation}
\label{eq:gammaf}
\sum_{\{j: A_i \subseteq U_j\}} \Pr(X_O = U_j \mid X_W \in A_i ) = 1.
\end{equation}
It easily follows from the CAR condition that 
$$\Pr(X_O = U_j \mid X_W \in A_i ) = \Pr(X_O = U_j \mid X_W \in U_j)$$ for
all $A_i \subseteq U_j$, so 
(\ref{eq:gammaf}) is equivalent to 
\begin{equation}
\label{eq:gammac}
\sum_{\{j: A_i \subseteq U_j\}} \Pr(X_O = U_j \mid X_W \in U_j ) = 1.
\end{equation}
(\ref{eq:gammac}) implies that
$
\sum_{\{j: A_i \subseteq U_j\}} \gamma_j =   1
$
for $i = 1, \ldots, k$.
Let
$\vec{s}_i$ be the row in  $S'$  corresponding to
$A_i$.
Since $\vec{s}_i$ has a $1$ as its $j$th component if $A_i
\subseteq U_j$ and a $0$ otherwise,
it follows that $\vec{s}_i \cdot \vec{\gamma}^T = 1$ and hence 
$S' \cdot \vec{\gamma}^T = \vec{1}^T$. 

For part (b), let $k$ be the number of rows in $S'$, let $\vec{s}_1,
\ldots, \vec{s}_k$ be the rows of $S'$, and let 
$A_1, \ldots, A_k$ be the corresponding atoms.
Fix $P_W \in {\cal P}_{W,S}$, and set $\alpha_i = P_W(A_i)$ for $i = 
1, \ldots, k$.  Let $\Pr$ be the unique distribution on $\R$ such that
\begin{equation}
\label{eq:snavel}
\begin{array}{llll}
\Pr(X_W \in A_i) &= &    \alpha_i,
\mbox{for $i = 1, \ldots, k$},
\\
\Pr(X_W \in A) &= &    0 \mbox{\ if $A \in {\cal A} - \{A_1,
  \ldots, A_k \}$}, \\
\Pr(X_O = U_j \mid X_W \in A_i) & = 
&\left \{
\begin{array}{ll}
\gamma_j &\mbox{if $A_i \in U_j$,}\\
0 &\mbox{otherwise.}
\end{array}
\right.
\end{array}
\end{equation}
Note that $\Pr$ is indeed a probability distribution on $\R$, since
$\sum_{A \in {\cal A}} \Pr(X_W \in A) = 1$, 
$\Pr(X_W \in A_i) > 0$ for $i = 1, \ldots, k$, 
and,  since we are assuming that $S'\cdot
\vec{\gamma}^T = \vec{1}^T$, 
$$\sum_{j= 1}^n \Pr(X_O = U_j \mid X_W \in A_i) = \vec{s}_i \cdot
\vec{\gamma}^T = 1,$$
for $i = 1, \ldots, k$.
Clearly $\Pr_W = P_W$. 
It remains to show that $\Pr$ satisfies CAR and that
$\gamma_j = \Pr(X_O = U_j \mid X_W \in U_j)$.
Given $j \in \{1, \ldots, n\}$, suppose that there exist atoms  $A_i, A_{i'} $ 
corresponding to rows $\vec{s}_i$ and $\vec{s}_{i'}$ of $S'$ such that
$A_i, A_{i'} \in U_j$. Then
$$\Pr(X_O = U_j| X_W \in A_i) =  \Pr(X_O = U_j| X_W \in A_{i'}) =
\gamma_j.$$
It now follows by Theorem~\ref{thm:CAR}(c) that $\Pr$
satisfies the CAR condition for $U_1, \ldots, U_n$. 
Moreover, Theorem~\ref{thm:CAR}(d), it must be the case that 
$\Pr(X_O = U_j \mid X_W \in U_j) = \gamma_j$.
\eprf

\bigskip
The proof of Theorem~\ref{thm:charCAR} builds on
Lemma~\ref{lem:precharCAR} and the following proposition, which shows
that the condition of part (b) of  Theorem~\ref{thm:charCAR} is
actually stronger than the condition of part (a). It is therefore not
surprising that it leads to a stronger conclusion.
\pro
\label{pro:bimpliesa} 
If there exists a subset $R$ of rows of $S$
that is linearly dependent but not affinely dependent, 
then
for all $\R$-atoms $A$ corresponding to a row in $R$ and all $j^* \in
\{1, \ldots, n\}$, 
if $A \subseteq U_{j^*}$, 
there exists a vector $\vec{u}$ that is an affine combination of the
rows in $R$ such that $u_j \geq 0$ for all $j \in  \{1, \ldots, n \}$
and $u_{j^*} > 0$. 
\epro

\prf
Suppose that there exists a subset $R$ of rows of $S$ that is linearly
dependent but not affinely dependent.
Without loss of generality, let $\vec{v}_1, \ldots, \vec{v}_k$ be the
rows in $R$.   
There exist $\lambda_1, \ldots, \lambda_k$ such that
$\kappa = \sum_{i=1}^k \lambda_i \neq 0$ and $\sum_{i=1}^k \lambda_i \vec{v}_i
= 0$. 
We first show that in fact every row $\vec{v}$ in $R$ is an affine combination
of the other rows.
Fix some $j \in \{1, \ldots, k\}$. 
Let $\mu_j =  (\lambda_j - \sum_{i=1}^k \lambda_i) =   - \sum_{i \ne j}
\lambda_i$ and  
let $\mu_i =  \lambda_i$ for $i \neq j$. Then $\sum_{i=1}^k \mu_i =
0$ and 
$$\sum_{i=1}^k \mu_i \vec{v}_i = \sum_{i=1}^k \lambda_i 
\vec{v}_i - \sum_{i=1}^k \lambda_i \vec{v}_j = - \kappa \vec{v}_j. 
$$ 
For $i = 1, \ldots, k$, let $\mu'_i = - \mu_i /
\kappa$. Then  $\sum_{i=1}^k \mu'_i = 0$ and $\sum_{i=1}^k \mu'_i
\vec{v}_i = 
\vec{v}_j$.
Now if $A_i  \subseteq U_{j^*}$ 
for some $i = 1, \ldots, k$ and some $j^* =
1, \ldots, n$, then $\vec{v}_i$ has a $1$ as its $j^*$th
component. Also, $\vec{v}_i$ is an affine combination of the rows of
$R$ with no negative components, 
so $\vec{v}_i$ is the desired vector.
\eprf

\othm{thm:charCAR}
Let $\R$ be a set of runs over observations $O 
= \{ U_1, \ldots, U_n \}$ and worlds $W$, and
let $S$ be the CARacterizing matrix for $O$ and $W$.
\begin{itemize}
\item[(a)] 
Suppose that
there exists a subset $R$ of the rows in $S$ and 
a vector $\vec{u} = (u_1, \ldots, u_n)$ that is an affine combination of 
the rows of $R$ such that $u_j \geq 0$ for all $j \in \{1, \ldots, n\}$ and
$u_{j^*} > 0$ for some $j^* \in \{1, \ldots, n\}$.
Then there is no distribution
$\Pr$ on $\R$ that satisfies CAR such that 
$\Pr(X_O = U_{j^*}) > 0$ and
$\Pr(X_W \in A) > 0$ for each
$\R$-atom $A$ corresponding to a row in $R$.
\item[(b)] If there exists a subset $R$ of the rows of $S$ that is
linearly dependent but 
not affinely dependent, then there is no distribution
$\Pr$ on $\R$ that satisfies CAR such that $\Pr(X_W \in A) > 0$ for each
$\R$-atom $A$ corresponding to a row in $R$. 
\item[(c)] Given a set $R$ consisting of $n$ linearly
independent rows of $S$ and a distribution $P_W$ on $W$ such
that $P_W(A) > 0$
for all $A$ corresponding to a row in $R$,
there is a unique distribution $P_O$ on $O$ such that
if $\Pr$ is a distribution on $\R$ satisfying
CAR and $\Pr(X_W \in A) = P_W(A)$ for each atom $A$ corresponding to a
row in $R$, then $\Pr(X_O = U) = P_O(U)$.  
\end{itemize}

\commentout{
Let $\R$ be a set of runs over observations $O$ and worlds $W$, and
let $S$ be the CARacterizing matrix for $O$ and $W$.
\begin{itemize}
\item[(a)] 
Suppose 
that
there exists a subset $R$ of the rows in $S$ and 
a vector $\vec{u} = (u_1, \ldots, u_n)$ that is an affine combination of 
the rows of $R$ such that $u_j \geq 0$ for all $j \in \{1, \ldots, n\}$ and
$u_{j^*} > 0$ for some $j^* \in \{1, \ldots, n\}$.
Then there is no distribution
$\Pr$ on $\R$ that satisfies CAR such that $\Pr(X_W \in A) > 0$ for each
$\R$-atom $A$ corresponding to a row in $R$ 
and $\Pr(X_O = U_{j^*}) > 0$.
\item[(b)] Given a set $R$ consisting of $n$ linearly
independent rows of $S$ and a distribution $P_W$ on $W$ such
that $P_W(A) > 0$, there is a unique distribution $P_O$ on $O$ such that
if $\Pr$ is a distribution on $\R$ satisfying
CAR and $\Pr(X_W \in A) = P_W(A)$ for each atom $A$ corresponding to a
row in $R$, then $\Pr(X_O = U) = P_O(U)$.  
\end{itemize}}
\commentout{
\begin{itemize}
\item[(a)] If there exists a subset $R$ of the rows of $S$ that is
linearly dependent but 
not affinely dependent, then there is no distribution
$\Pr$ on $\R$ that satisfies CAR such that $\Pr(X_W \in A) > 0$ for each
$\R$-atom $A$ corresponding to a row in $R$. 
\item[(b)] If there exists a subset $R$ of the rows in $S$ and 
a vector $\vec{u}$ that is an affine combination of 
the rows of $R$ such that $\vec{u} \neq \vec{0}$ and
every component of the vector $\vec{u}$ is nonnegative,
then there is no distribution
$\Pr$ on $\R$ that satisfies CAR such that $\Pr(X_W \in A) > 0$ for each
$\R$-atom corresponding to a row in $R$ 
and $\Pr(X_O = U) > 0$ for each $U \in O$.
\item[(c)] Given a set $R$ consisting of $n$ linearly
independent rows of $S$ and a distribution $P_W$ on $W$ such
that $P_W(A) > 0$, there is a unique distribution $P_O$ on $O$ such that
if $\Pr$ is a distribution on $\R$ satisfying
CAR and $\Pr(X_W \in A) = P_W(A)$ for each atom $A$ corresponding to a
row in $R$, then $\Pr(X_O = U) = P_O(U)$.  
\end{itemize}
}
\eothm

\prf
\commentout{
Suppose that $\Pr$ is a distribution on $\R$ that satisfies CAR.
Let $\alpha_i = \Pr(X_W \in A_i)$, for $i = 1, \ldots, m$, and let
$\beta_j = \Pr(X_O = U_j)$, for $j = 1, \ldots, n$.
Note that for $j = 1, \ldots, m$, we have
$$\beta_j = \Pr(X_O = U_j) = \Pr(X_O = U_j \mid X_W \in U_j) \Pr(X_W \in
U_j).$$ 
If 
$\Pr(X_W \in U_j) = \sum_{\{i: A_i \subseteq U_j\}} \alpha_i > 0$,
then
it follows that 
\begin{equation}
\label{eq:gammab}
\Pr(X_O = U_j \mid X_W \in U_j) = \frac{\beta_j}{\sum_{\{i: A_i \subseteq
    U_j\}} \alpha_i}.
\end{equation}
(We remark that (\ref{eq:gammab}) was also derived in
\cite{GillLR97}, but the  further developments in the proof are
original.)
Let
$$\gamma_j = \frac{\beta_j}{\sum_{\{i: A_i \subseteq
    U_j\}} \alpha_i},$$
for all $j$ with $\sum_{\{i: A_i \subseteq  U_j\}} \alpha_i > 0$.
If  $\sum_{\{i: A_i \subseteq
    U_j\}} \alpha_i = 0$, we define  $\gamma_j = 0$.
Note that $\gamma_j > 0$ if $\beta_j > 0$.

Next note that if $\alpha_i > 0$, then
\begin{equation}
\label{eq:gammaf}
\sum_{\{j: A_i \subseteq U_j\}} \Pr(X_O = U_j \mid X_W \in A_i ) = 1.
\end{equation}
It easily follows from the CAR condition that 
$\Pr(X_O = U_j \mid X_W \in A_i ) = \Pr(X_O = U_j \mid X_W \in U_j)$ for
all $A_i \subseteq U_j$, so 
(\ref{eq:gammaf}) is equivalent to 
\begin{equation}
\label{eq:gammac}
\sum_{\{j: A_i \subseteq U_j\}} \Pr(X_O = U_j \mid X_W \in U_j ) = 1.
\end{equation}
Combining (\ref{eq:gammab}) and (\ref{eq:gammac}), it follows that 
for each $i$ such that $\alpha_i > 0$, we have the following equation:
\begin{eqnarray}
\label{eq:lincar}
\sum_{\{j: A_i \subseteq U_j\}} \gamma_j =   1.
\end{eqnarray}

For part (a) 
suppose, by way of contradiction, that the rows
in $R$ are linearly dependent but not affinely dependent.  Without loss
of generality, suppose that $R$ consists of the rows $\vec{v}_1, \ldots,
\vec{v}_k$, which correspond to the atoms $A_1, \ldots, A_k$. 
By assumption, there exist coefficients $\lambda_1, \ldots, \lambda_k$
such that $\sum_{i=1}^k \lambda_i \ne 0$, but 
$\sum_{i=1}^k \lambda_i \vec{v}_i = \vec{0}$.
Let $\vec{\gamma}$ be
the column vector of length $n$ whose $j$th component is $\gamma_j$.
Suppose, by means of contradiction, that  $\alpha_i = \Pr(X_W \in A_i) > 0$ for
$i \in \{1, \ldots, k \}$. Then (\ref{eq:lincar}) says that for $i
\in \{ 1, \ldots, k \}$, we have 
$\vec{v}_i \cdot \vec{\gamma} = 1$, where $\cdot$ represents inner
product. 
Thus, 
$$
0 = \left(\sum_{i=1}^k \lambda_i
\vec{v}_i \right) \cdot \vec{\gamma} = \sum_{i=1}^k \lambda_i
(\vec{v}_i \cdot \vec{\gamma}) = \sum_{i=1}^k \lambda_i \ne 0,
$$
giving us the desired contradiction.

Now for part (b). 

}
For part (a), suppose that $R$ consists of $\vec{ v}_1, \ldots,
\vec{v}_k$, corresponding to atoms $A_1, \ldots, A_k$. 
By assumption, there exist coefficients $\lambda_1, \ldots, \lambda_k$
such that $\sum_{i=1}^k \lambda_i = 0$,
and a vector $\vec{u} = \sum_{i=1}^k \lambda_i \vec{v}_i$ such that 
every component of $\vec{u}$ is nonnegative.
Suppose, by way of contradiction, 
that 
$\Pr$ satisfies CAR and
that
$\alpha_i = \Pr(X_W \in A_i) > 0$ for 
$i \in \{1, \ldots, k \}$. 
By Lemma~\ref{lem:precharCAR}(a), we have
\begin{equation}\label{eq:contra}
\vec{u} \cdot \vec{\gamma} = 
\left(\sum_{i=1}^k \lambda_i
\vec{v}_i \right) \cdot \vec{\gamma} = \sum_{i=1}^k \lambda_i
(\vec{v}_i \cdot \vec{\gamma}) = \sum_{i=1}^k \lambda_i = 0,
\end{equation}
where $\vec{\gamma}$ is defined as in Lemma~\ref{lem:precharCAR}.
For $j = 1, \ldots, n$, 
if $\Pr(X_O = U_j) > 0$ then $\Pr(X_O = U_j \inter X_W \in U_j ) = \Pr(X_O
= U_j) > 0$ and $\Pr(X_W \in U_j) > 0$, so $\gamma_j > 0$. By
assumption, all the components of $\vec{u}$ and $\vec{\gamma}$ are 
nonnegative. Therefore, if there exists $j^*$ such that $\Pr(X_O =
U_{j^*}) > 0$ and
$u_{j^*} > 0$, then $\vec{u} \cdot \vec{\gamma} > 0$.
This contradicts (\ref{eq:contra}), and part (a) is proved.
For part (b), suppose 
that 
there exists a subset $R$ of rows of $S$ that is
linearly dependent but not affinely dependent. Suppose, 
by way of contradiction, that $\Pr$ satisfies CAR and that $\Pr(X_W \in
A) > 0$ for all atoms $A$ corresponding 
to a row in $R$. Pick an atom $A^*$ corresponding to such a row. By
Proposition~\ref{pro:bimpliesa} and Theorem~\ref{thm:charCAR}(a),  
we have that $\Pr(X_O = U_{j^*}) = 0$ for all $j^*$ such
that $A^* \in U_{j^*}$. But then $\Pr(X_W \in A^*) = 0$, and we have
arrived at a contradiction.
\commentout{For part (b), suppose that there is a set $R$ of rows of $S$ 
and a vector $\vec{u}$ that is an affine combination of 
the rows of $R$ such that $\vec{u} \neq \vec{0}$ and
every component of the vector $\vec{u}$ is nonnegative.
Using the same notation as for part (a), we get that 
\begin{equation}\label{eq:contra}
\vec{u} \cdot \vec{\gamma} = 
\left(\sum_{i=1}^k \lambda_i
\vec{v}_i \right) \cdot \vec{\gamma} = \sum_{i=1}^k \lambda_i
(\vec{v}_i \cdot \vec{\gamma}) = \sum_{i=1}^k \lambda_i = 0.
\end{equation}
Since, as we observed earlier, the fact that $\Pr(X_O = U_j) = 
\beta_j > 0$ for $j = 1,
\ldots, n$ implies that $\gamma_j > 0$ for $j = 1, \ldots, n$, and, by
assumption, all the components of $\vec{u}$ are nonnegative and at least
one is positive, it must be the case that $\vec{u} \cdot \vec{\gamma} > 0$.
This gives a contradiction to (\ref{eq:contra}).
}

For part (c), suppose that $R$ consists of the rows
$\vec{v}_1, \ldots, \vec{v}_n$.  Let $S'$ be the $n \times n$ submatrix
of $S$ consisting of the rows of $R$.  Since these rows are linearly
independent, a standard result of linear algebra says that $S'$ is
invertible.  
Let $\Pr$ be a distribution on $\R$ satisfying CAR. By
Lemma~\ref{lem:precharCAR}(a),
$S' \vec{\gamma} = \vec{1}^T$.
Thus, $\vec{\gamma} = (S')^{-1} \vec{1}$.  
For $j = 1, \ldots, n$ we must have 
$\gamma_j = \beta_j/\Pr(X_W \in U_j)$, where $\beta_j = \Pr(X_O =  U_j)$. 
Given 
$\Pr_W(A)$ for each atom $A$, 
we can clearly solve for the $\beta_j$'s.
\eprf

\medskip

\commentout{\pro\label{pro:bimpliesa} If there exists a subset $R$ of rows of $S$
that is linearly dependent but not affinely dependent, then there is
some subset $R'$ of 
the rows in $R$ such that an affine combination of the rows in $R'$
gives a nonzero vector all of whose components are non-negative.
}

\commentout{
\opro{pro:CARsensor}
${\sc CARgen}^{**}$ halts with probability 1 {\em \bf and\/} for all $w \in
W$, 
$\Pr^*(X_W = w) =
P_W(w)$ {\em \bf and\/} the distribution $\Pr^*$ satisfies CAR \\
\begin{centering}  
if and only if \\
\end{centering}
there  exists some $0 \leq  q < 1$ with for all $w \in W$, $q_w
= q$.
\eopro

\prf 
The proof consists of two stages:
\paragraph{Stage 1} Here we show that 
${\sc CARgen}^{**}$ halts with probability 1 {\em \bf and\/} for all $w \in
W$, 
$\Pr^*(X_W = w) =
P_W(w)$ {\em 
if and only if}
there  exists some $0 \leq  q < 1$ with for all $w \in W$, $q_w
= q$.

To see this, note that for all $w \in W$ we have the recursive relation:
\begin{equation}
  \label{eq:recursion}
  \Pr(X_W = w) = P_W(w) (1 - q_w) +\sum_{w' \in W} P_W(w') q_{w'} \Pr(X_W = w)
\end{equation}
Let $q = \sum_{w' \in W} P_W(w') q_{w'}$. If $q < 1$, then
(\ref{eq:recursion}) can be rewritten as 
\begin{equation}
  \label{eq:recursionb}
  \Pr(X_W = w) = P_W(w) \frac{1 - q_w}{1-q}.
\end{equation}
It follows that 
\begin{equation}
\label{eq:statement}
\mbox{if\ } q< 1 \mbox{\ then\ } [ \ \Pr(X_W = w) = P_W(w) 
\mbox{\ for all $w
\in W$ {\em iff\/} $q_w = q$ for all $w \in W$}.\ ]
\end{equation} 
This implies both directions
of the statement we have to prove, since:
\begin{enumerate}
\item If ${\sc CARgen}^{**}$ halts with probability $1$, then $q =
  \sum_{w' \in W} P_W(w') q_{w'} < 1$. Therefore, if  ${\sc
    sensor}^{**}$ halts with probability $1$ and $\Pr(X_W = w) = P_W(w)$ for all $w
\in W$, then by (\ref{eq:statement}), $q_w = q$ for all $w \in W$, and $0 \leq q < 1$.
\item If for all $w \in W$, $q_w = q'$ for some $0 \leq q' < 1$ then
${\sc CARgen}^{**}$ halts with probability $1$ and we must have $q' =
q = \sum_{w' \in W} P_W(w') q_{w'} <1$, so by (\ref{eq:statement}),  
$\Pr(X_W = w) = P_W(w)$ for all $w
\in W$.
\end{enumerate}
\paragraph{Stage 2}
We now show that if for all $w \in W$, $q_w = q =   \sum_{w' \in W} P_W(w')
q_{w'} < 1$, then $\Pr^*$ satisfies CAR. First note that
\begin{eqnarray}
\Pr(X_O = U \inter X_W = w) & = & P_W(w) \sum_{\{s: U \in s\}} P_S(s) (1 - q_{U|\Pi})
+ \nonumber \\
& & \sum_{w'} P_W(w') q_{w'} 
\Pr(X_O = U \inter X_W = w)
\end{eqnarray}
Since we are assuming $q< 1$, this can be rewritten as
$$
\Pr(X_O = U \inter X_W = w) = P_W(w) \frac{ \sum_{\{s: U \in s\}} P_S(s) (1 - q_{U|\Pi})}{1-q}
$$
whence, by (\ref{eq:recursionb}),
\begin{equation}
\label{eq:recursionc}
\Pr(X_O = U \mid X_W = w)= \frac{\Pr(X_O = U \inter X_W = w)}{\Pr(X_W = w)}
=
\frac{\sum_{\{s: U \in s\}} P_S(s) (1 - q_{U|\Pi})}{1-q_w} 
\end{equation}
Since we are assuming $q_w = q_{w'}$ for all $w, w' \in W$,
(\ref{eq:recursionc})  shows that $\Pr(X_O = U \mid X_W = w) = \Pr(X_O =
U \mid X_W = w')$ for all $w', w \in W$ and hence CAR is satisfied.

\ 
\\
Together, the partial results of Stage 1 and 2 prove the Proposition.
\eprf

\medskip

\othm{thm:CARresolved}
Given any $O$ and $W$, let $\Pr$ be a distribution over $\R$ such that
CAR holds. Then $\Pr = \Pr^*$ where $\Pr^*$ arises from an instantiation of ${\sc
  sensor}^*$. 
\eothm

\prf 
Let $O = \{ U_1, \ldots, U_n \}$ and let $\beta_i = \Pr(X_O =
U_i)$. Without loss of generality we may assume that $\beta_i >
0$ for $i=1..n$.

We first set up the procedure ${\sc CARgen}^{**}$ such that it
simulates $\Pr$. For this, set $P_W(w) = \Pr(X_W = w)$. Let, for $i = 1..n$, $\Pi_i := \{ U_i,
\overline{U_i} \}$. Set $P_S(\Pi_i) := \Pr(X_O = U_i) = \beta_i$ and
$q_{\overline{U}_i|\Pi_i} = 1$. Set 
$$\gamma_i := \epsilon/ \Pr(X_W \in
U_i)$$ 
where $\epsilon := \min_{i = 1..n} \Pr(X_W \in U_i)$. 
Since $\beta_i > 0$ and $\Pr(X_W \in U_i \mid X_O = U_i) = 1$, 
all $\gamma_i$ are well-defined and $0 < \gamma_i
\leq 1$. We now set $q_{U_i|\Pi_i} := 1- \gamma_i$.

Let {\sc accept} be the event that ${\sc CARgen}^{**}$ halts in the
first round, i.e. the very first $U$ that is generated is also
accepted and ${\sc CARgen}^{**}$ never jumps back to step 2.1.
We have 
\begin{eqnarray}
  \label{eq:1}
  \mbox{$\Pr^*$}(X_O = U_i ; {\sc accept}
  ) 
& =  & \beta_i (1- q_{U_i|\Pi_i}) \Pr(X_W \in U_i) \nonumber \\
& = &  \beta_i \gamma_i \Pr(X_W \in U_i) = \beta_i \epsilon
\end{eqnarray}
For $w \in U_i$, we have 
\begin{eqnarray}
  \label{eq:2}
\lefteqn{
  \mbox{$\Pr^*$}(X_W = w \inter X_O = U_i ; {\sc accept}
  )} & & \nonumber \\  
& = &  \beta_i (1- q_{U_i|\Pi_i}) \Pr(X_W \in U_i) \Pr(X_W = w \mid X_W \in
U_i) \nonumber \\
& = &   \beta_i \epsilon
\Pr(X_W = w \mid X_W \in U_i)
\end{eqnarray}
Also,
\begin{equation}
  \label{eq:3}
  \mbox{$\Pr^*$}({\sc accept}
  ) 
= \sum_{i=1}^n \beta_i (1- q_{U_i|\Pi_i}) \Pr(X_W \in U_i)
= \sum_{i=1}^n \beta_i \gamma_i \Pr(X_W \in U_i) = \epsilon > 0
\end{equation}
so that
\begin{eqnarray}
  \label{eq:4}
&   \mbox{$\Pr^*$}(X_O = U_i \mid {\sc accept}
  ) 
= \beta_i = \Pr(X_O = U_i) &  \\
&   \Pr^*(X_W = w \inter X_O = U_i \mid {\sc accept}
  ) = \beta_i 
\Pr(X_W = w \mid X_W \in U_i) & \nonumber
\end{eqnarray}
which gives
\begin{eqnarray}
  \label{eq:5}
  \mbox{$\Pr^*$}(X_W = w \mid X_O = U_i ; {\sc accept}
  ) 
& = & \Pr(X_W = w \mid X_W \in U_i) \nonumber \\
& = &  \Pr(X_W = w \mid X_O = U_i),
\end{eqnarray}
where the last equality follows since $\Pr$ satisfies CAR.
Taken together, (\ref{eq:4}) and (\ref{eq:5}) show that
\begin{equation}
\label{eq:6}
\mbox{$\Pr^*$}(\cdot \mid  {\sc accept}) =
\Pr.
\end{equation}
If ${\sc CARgen}^{**}$ does not halt in the first round, then the
procedure of generating $w \in W$, $\Pi \in \P$ and $U \in s$ is started
from scratch, based on the same distributions $P_W, P_\P,
q_{U|\Pi}$. Therefore we must have 
$$\mbox{$\Pr^*$} = \mbox{$\Pr^*$}(\cdot \mid  {\sc accept}).$$
Together with (\ref{eq:6}) this shows that the distribution $\Pr^*$
generated by ${\sc CARgen}^{**}$ is equal to $\Pr$. Since $\Pr$
satisfies CAR, we must have that $\Pr^*$ satisfies
CAR; it only remains to show that the instantiation of ${\sc
  sensor}^{**}$ that we defined here is actually an instantiation of
${\sc CARgen}^{*}$. But this follows from
Proposition~\ref{pro:CARsensor}, which, since $\Pr^*$ satisfies CAR, 
gives that $P_W, P_\P, q_{U|\Pi}$ must be such that $q_w = q_{w'}$ for all $w, w' \in
W$. The theorem is proved.
\eprf
}

\medskip

\othm{thm:CARresolved}
Given a set $\R$ of runs over a set
$W$ of worlds and a set $O$ of observations, $\Pr$ is a distribution on 
$\R$ that satisfies CAR iff there is a setting of the parameters in 
{\sc CARgen$^*$} such that, for all $w \in W$ and $U \in O$, 
$\Pr(\{r: X_W(r) = w, \, X_O(r) = U\})$ is
the probability that {\sc CARgen$^*$} returns $(w,U)$.
\eothm

\prf 
First we show that if $\Pr$ is a probability on $\R$ such that, for some
setting of the parameters of {\sc CARgen$^*$}, 
$\Pr(\{r: X_W(r) = w, \, X_O(r) = U\})$ is
the probability that {\sc CARgen$^*$} returns $(w,U)$, then $\Pr$
satisfies CAR.  By Theorem~\ref{thm:CAR}, it suffices to show that, for
each set $U \in O$ and worlds $w_1,  w_2 \in U$ such that $\Pr(X_W =
w_1) > 0$ and $\Pr(X_W = w_2) > 0$, we have  
$\Pr(X_O = U \mid X_W = w_1) = \Pr(X_O = U \mid X_W = w_2)$.  
So suppose that $w_1, w_2 \in U$, $\Pr(X_W = w_1) > 0$, and $\Pr(X_W =
w_2) > 0$.   
Let $\alpha_U = \sum_{\{\Pi \in \P: U \in \Pi\}}P_\P(\Pi)(1 -
q_{U|\Pi})$.
Intuitively, $\alpha_U$ is the probability that the algorithm terminates
immediately at step 2.3 with $(w,U)$ conditional on some $w \in U$ being
chosen at step 2.1.
Notice for future reference that, for all $w$, 
\begin{equation}\label{eq:alphaU}
\sum_{\{U: w \in U\}} \alpha_U = \sum_{\{(U,\Pi): U \in
\Pi, w \in U\}}P_\P(\Pi)(1 - q_{U|\Pi}) = 1 - q,
\end{equation}
where $q$ is defined
by (\ref{CARconstraint}).
As explained in the main text, for both $i = 1,2$, 
$q$ is the probability that the algorithm does not
terminate at step 2.3 given that $w_i$ is chosen in step 2.1.
It easily follows that 
the probability that $(w_i,U)$ is output at step 2.3 is 
$$P_W(w_i) \alpha_U (1 + q + q^2 + \cdots) = P_W(w_i)
\alpha_U/(1-q).$$
Thus, $\Pr(X_W = w_i \inter X_O = U) = P_W(w_i) \alpha_U/(1-q)$.  Using
(\ref{eq:alphaU}), we have that 
$$\Pr(X_W = w_i) = \sum_{\{U: w_i \in U\}} \Pr(X_W = w_i \inter X_O = U) = 
\frac{P_W(w_i)}{1-q} \sum_{\{U: w_i \in U\}} \alpha_U = P_W(w_i).$$
Finally, we have that $\Pr(X_O = U \mid X_W = w_i) =
\alpha_U/(1-q)$, for $i = 1,2$.  Thus, $\Pr$ satisfies the CAR condition.

For the converse, suppose that $\Pr$ satisfies the CAR condition.  
Let $O = \{U_1, \ldots, U_n\}$.   We choose the parameters for {\sc
CARgen$^*$} as follows.
Set $P_W(w) = \Pr(X_W = w)$ and let $\beta_i = \Pr(X_O = U_i)$.
Without loss of generality, we assume that $\beta_i > 0$ (otherwise,
take $O'$ to consist of those sets that are observed with positive
probability, and do the proof using $O'$).

For $i = 1, \ldots, n$, let $\Pi_i = \{ U_i,
\overline{U_i} \}$. Set $P_\P(\Pi_i) = \Pr(X_O = U_i) = \beta_i$ and
$q_{\overline{U}_i|\Pi_i} = 1$.  (Thus, the set $\overline{U}_i$ is always
rejected, unless $\overline{U}_i = U_j$.)
Since $\Pr(X_W \in U_j) \ge \Pr(X_O = U_j) > 0$ by assumption, it must
be the case that $
\epsilon = 
\min_{j=1}^n \Pr(X_W \in U_j) > 0$.  
Now set
$q_{U_i|\Pi_i} = 1 - \epsilon/\Pr(X_W \in U_i)$.
We first show that, with these parameter settings, we can choose $q$
such that constraint (\ref{CARconstraint}) is satisfied.
Let
$q_w = \sum_{\{U, \Pi: \; w \in U, \,  U \in \Pi\}}
P_\P(\Pi) q_{U|\Pi}$.
For each $w \in W$ such that $P_W(w)> 0$, we have
$$\begin{array}{llr}
&q_w\\
= &\sum_{\{U, \Pi: \; w \in U , \,  U \in
\Pi\}} P_\P(\Pi)q_{U|\Pi} \\
= &\sum_{i=1}^n \sum_{\{U: \; w \in U , \,  U \in \Pi_i\}} P_\P(\Pi_i)q_{U|\Pi_i}\\
= &\sum_{\{i: w \in U_i\}} P_\P(\Pi_i)q_{U_i|\Pi_i} +
\sum_{\{i: w \in \overline{U}_i\}} P_\P(\Pi_i)q_{\overline{U}_i|\Pi_i}.
\end{array}
$$
The last equality follows because
$\Pi = \{U_i, \overline{U}_i\}$.  Thus, for a fixed $i$, 
$\sum_{\{U: \; w \in U , \,  U \in \Pi_i\}} P_\P(\Pi_i)q_{U|\Pi_i}$ is either
$P_\P(\Pi_i)q_{U_i|\Pi_i}$ if $w \in U_i$,
or 
$P_\P(\Pi_i)q_{\overline{U}_i|\Pi_i}$  if
$w \in \overline{U}_i$.
It follows that
$$\begin{array}{llr}
&q_w  \\
= &\sum_{\{i : w \in U_i\}} \beta_i (1 - \epsilon/\Pr(X_W \in U_i))
+ \sum_{\{i : w \notin U_i\}} \beta_i \cdot 1 \\
= &\sum_{\{i : w \in U_i\}} \Pr(X_O = U_i) (1 - \epsilon/\Pr(X_W \in U_i))
+ \sum_{\{i: w \notin U_i\}} \Pr(X_O = U_i)\\
= &\sum_{i= 1}^n \Pr(X_O = U_i) - \epsilon \sum_{\{i: w \in U_i\}}
\Pr(X_O = U_i \mid X_W \in U_i)\\ 
= &1 - \epsilon \sum_{\{i: w \in U_i\}}
\Pr(X_O = U_i \mid X_W = w)  \ \ \ \ \  \mbox{[since $\Pr$ satisfies CAR]}\\
= &1 - \epsilon.
\end{array}
$$
Thus, $q_w = q_{w'}$ if $P_W(w), P_W(w') > 0$, so these parameter settings
are appropriate for {\sc CARgen$^*$} (taking $q=q_w$ for any $w$ such
that $P_W(w) > 0$). 
Moreover, $\epsilon = 1- q$.

We now show that, with these parameter settings,
$\Pr(X_W = w \inter X_O = U)$ is the probability that {\sc CARgen$^*$}
halts with $(w,U)$, for all $w \in W$ and $U \in O$.
Clearly if $\Pr(X_W = w) = 0$, this is true, since then
$\Pr(X_W = w \inter X_O = U) = 0$, and the probability that {\sc
CARgen$^*$} halts with output $(w,U)$ is at most $P_W(w) = \Pr(X_W = w)
= 0$.
So suppose that $\Pr(X_W = w) > 0$.  Then it suffices to show that
$\Pr(X_O = U_i \mid X_W = w)$ is the probability that $(w,U_i)$ is output,
given that $w$ is chosen at the first step.  But the argument of the
first half of the proof shows that this probability is just
$\frac{\alpha_{U_i}}{1- q}$.  But
$$\begin{array}{lll}
&\frac{\alpha_{U_i}}{1- q}\\ 
= &\frac{\alpha_{U_i}}{\epsilon} &\mbox{[since $\epsilon = 1-q$]}\\ 
= &\frac{\sum_{\{\Pi \in \P: U_i \in
\Pi\}}P_\P(\Pi)(1 - q_{U_i|\Pi})}{\epsilon}\\
= &\frac{\beta_i (\epsilon/\Pr(X_W \in U_i))}{\epsilon} \\
= &\Pr(X_O = U_i)/\Pr(X_W \in U_i) \\
= &\Pr(X_O = U_i \mid X_W = w) &\mbox{[since $\Pr$ satisfies CAR]},
\end{array}
$$
as desired.  
\eprf

\bigskip

\othm{thm:JeffreyOK}
Fix 
a probability $\Pr$ on $\R$,
a partition $\{U_1, \ldots, U_n\}$ of $W$,
and probabilities $\alpha_1, \ldots, \alpha_n$ such that $\alpha_1 +
\cdots + \alpha_n = 1$.
Let $C$ be the observation $\alpha_1 U_1; \ldots ; \alpha_n U_n$.
Fix some $i \in \{1,\ldots, n\}$.
Then the following are equivalent:
\begin{itemize}
\item[(a)] If $\Pr(X_O = C) > 0$, then 
$\Pr(X_W = w \mid X_O = C) = \Pr_W(w \mid \alpha_1 U_1; \ldots;
\alpha_n U_n)$
for all $w \in U_i$.
\item[(b)] $\Pr(X_O = C \mid X_W = w) =
\Pr(X_O = C \mid X_W \in U_i)$ 
for all
$w \in U_i$ such that $\Pr(X_W = w) > 0$.
\end{itemize}
\eothm

\prf 
The proof is similar in spirit to that of Theorem~\ref{thm:CAR}.
Suppose that (a) holds, $w \in U_i$, and $\Pr(X_W = w) > 0$.  
Then 
$$\begin{array}{ll}
&\Pr(X_O = C \mid X_W = w)\\
= &\Pr(X_W = w \mid X_O = C) \Pr(X_O = C)/\Pr(X_W = w)\\
= &\Pr_W(w \mid \alpha_1 U_1; \ldots; \alpha_n U_n) \Pr(X_O = C)/\Pr(X_W
=  w)\\ 
= &\alpha_i \Pr_W(w \mid U_i) \Pr(X_O = C)/ \Pr_W(w)\\
= &\alpha_i \Pr(X_O = C)/\Pr_W(U_i)\\
\end{array}
$$
Similarly,
$$\begin{array}{ll}
&\Pr(X_O = C \mid X_W \in U_i)\\
= &\Pr(X_W \in U_i \mid X_O = C) \Pr(X_O = C)/\Pr(X_W \in U_i)\\
= &\sum_{w' \in U_i} \Pr_W(w' \mid \alpha_1 U_1;
\ldots; \alpha_n U_n) \Pr(X_O = C)/\Pr(X_W \in U_i)\\ 
= &\sum_{w' \in U_i} \alpha_i \Pr_W(w' \mid U_i) \Pr(X_O = C)/ \Pr(X_W
\in U_i)\\
= &\alpha_i \Pr(X_O = C)/\Pr_W(U_i)
\end{array}
$$
Thus, $\Pr(X_O = C \mid X_W = w) = \Pr(X_O = C \mid X_W \in U_i)$
for all $w \in U_i$ such that $\Pr(X_W = w) > 0$.

For the converse, suppose that (b) holds and $\Pr(X_O = C) > 0$.  
Given $w \in U_i$, 
if $\Pr(X_W = w) = 0$, then (a) trivially holds, so
suppose that $\Pr(r(X_W = w) > 0$.  Suppose that $w \in U_i$.
Clearly  $\Pr(w \mid \alpha_1 U_1; \ldots; \alpha_n U_n) = \alpha_i
\Pr_W(w \mid U_i)$.  Now, using (b), we have that
$$\begin{array}{ll}
&\Pr(X_W = w \mid X_O = C) \\
= &\Pr(X_O = C \mid X_W = w) \Pr(X_W = w) / \Pr(X_O = C)\\
= &\Pr(X_O = C \mid X_W \in U_i) \Pr(X_W = w)/ \Pr(X_O = C)\\
= &\Pr(X_W \in U_i \mid X_O = C) \Pr(X_W = w)/ \Pr(X_W \in U_i) \\
= &\alpha_i \Pr_W(w \mid U_i) \mbox{\ \ \ \ [using (\ref{eq:accurate})].}
\end{array}$$
Thus, (a) holds.
\eprf

\medskip

\opro{pro:GCARcanhold}
Consider a partition $\{U_1, \ldots, U_n\}$ of $W$ and a set of 
$k > 1$ observations $O = \{C_1, \ldots, C_k \}$ with $C_i =
\alpha_{i1} U_1 ; \ldots ; \alpha_{in} U_n$ such that all $\alpha_{ij}
> 0$.
For every distribution $P_O$ on $O$ with $P_O(C_i) > 0$ for all
$i \in \{1,\ldots, k\}$, 
there exists a distribution $\Pr$ on $\R$ such that 
$P_O = \Pr_O$ (i.e. $P_O$ is the marginal of $\Pr$ on $O$)
and $\Pr$ satisfies the generalized
CAR condition 
(part (b) of Theorem~\ref{thm:JeffreyOK})
for $U_1, \ldots, U_n$.
\eopro

\prf 
Given a set $W$ of worlds, a set $O =
\{C_1, \ldots, C_k\}$
of observations
with distribution $P_O$ satisfying  $P_O(C_i) > 0$ for $i \in \{1,\ldots,
k\}$, and 
arbitrary distributions $\Pr_j$ on $U_j$, $j = 1, \ldots, n$,
we
explicitly construct a prior
$\Pr$ on $\R$ that satisfies CAR
such that $P_O = \Pr_O$, where $\Pr_O$  
is the marginal of $\Pr$ on $O$ and $\Pr_j = \Pr_W(
 \cdot \mid U_j)$.

Given $w \in U_j$, 
define $$\Pr(\{r \in \R: X_O(r) = C_i, X_W(r) = w\}) = P_O(C_i)
\alpha_{ij} {\Pr}_j(w).$$
(How the probability is split up over all the runs $r$ such that $X_O(r)
= C_i$ and $X_W(r) = w$ is irrelevant.)
It remains to check that $\Pr$ is a distribution on $\R$ and that it
satisfies all the requirements.
It is easy to check that
$$\Pr(X_O = C_i) = \sum_{j = 1}^n \sum_{w \in U_j} P_O(C_i)
\alpha_{ij} {\Pr}_j(w) = P_O(C_i).$$
It follows that $\sum_{i = 1}^k \Pr(X_O = C_i) = 1$, showing that $\Pr$
is a probability measure and $P_O$ is the marginal of $\Pr$ on $O$.
If $w \in U_j$, then 
$$\begin{array}{ll}
&\Pr_W(w \mid U_j) = \Pr_W(w)/ \Pr_W(U_j) \\
= & \frac{\sum_{i = 1}^k \Pr_O(C_i)
\alpha_{ij} \Pr_j(w)}{\sum_{w' \in U_j} \sum_{i = 1}^k \Pr_O(C_i)
\alpha_{ij}\Pr_j(w')}\\
= &\frac{\Pr(w) \sum_{i = 1}^k \Pr_O(C_i) \alpha_{ij}}{(\sum_{w' \in U_j}
\Pr_j(w'))  \sum_{i = 1}^k \Pr_O(C_i) \alpha_{ij}}\\
= &\Pr_j(w).
\end{array}$$
Finally, note that, for 
$j \in \{1, \ldots, n \}$, for all $w \in U_j$ such that $\Pr(X_W = w) > 0$, we
have that 
$$\begin{array}{ll}
&\Pr(X_O = C_i \mid X_W = w)\\
= &\frac{\Pr_O(C_i) \alpha_{ij} \Pr_j(w)}
{\sum_{i = 1}^k \Pr_O(C_i) \alpha_{ij} \Pr_j(w)} \\
= &\frac{\Pr_O(C_i) \alpha_{ij} }
{\sum_{i = 1}^k \Pr_O(C_i) \alpha_{ij} } \\
= & \frac{\Pr_O(C_i) \alpha_{ij}\Pr(X_W \in U_j)  }
{\sum_{i = 1}^k \Pr_O(C_i) \alpha_{ij}\Pr(X_W \in U_j) } \\
= & \frac{\Pr(X_O = C_i \inter X_W \in U_j)}{\Pr(X_W \in U_j)}\\
= & \Pr(X_O = C_i \mid X_W \in U_j) 
\end{array}$$
so the generalized CAR condition holds for $\{U_1, \ldots, U_n \}$.
\eprf

\bigskip

To prove 
Theorem~\ref{thm:MREnotOK},
we first need some background on minimum relative entropy
distributions. 
\commentout{Given a distribution $P_W$ on $W$,
Theorems 2.1 and 3.1 \cite{Csiszar75} together show 
that for each $U \subseteq W$ and $0 < \alpha < 1$, 
there is a 
vector
$\beta \in {\bf R}$ such that 
for all $w \in W$:
\begin{equation}
\label{eq:onevar}
\mbox{$P_W$}(w \mid \alpha U ) = \frac{1}{Z}e^{\beta {\bf 1}_U(w)}
 \mbox{$P_W$}(w).
\end{equation}
Here $Z= \sum_{w \in W} e^{\beta {\bf 1}_{U}(w)} P_W(w) $ is
and ${\bf 1}_U$ is the indicator function, i.e. ${\bf 1}_{U}(w) = 1$
if $w \in U$ and $0$ otherwise.
Conversely, for each $U \subseteq W$, for each $\beta \in {\bf R}$
there exists $0 < \alpha < 1$ such that for all $w \in W$ (\ref{eq:onevar})
holds.
}
Fix some 
space
$W$ and let $U_1, \ldots, U_n$ be subsets of $W$.
Let $\Delta$ be the set of $(\alpha_1, \ldots, \alpha_n)$ for which
there exists some distribution $P_W$ with $P_W(U_i) = \alpha_i$ for $i =
1, \ldots, n$ and $P_W(w) >0$ for all $w \in W$.
Now let $P_W$ be a distribution  
with $P_W(w) > 0$ for all $w \in W$.
Given a 
vector $\vec{\beta} = (\beta_1, \ldots, \beta_n) \in {\bf R}^n$, let
$$
 P_W^{\vec{\beta}}(w) =
\frac{1}{Z}e^{\beta_{1} {\bf 1}_{U_1}(w) + \ldots + \beta_{n} {\bf
 1}_{U_n}(w)} \mbox{$P_W$}(w),
$$
where ${\bf 1}_U$ is the indicator function, i.e. ${\bf 1}_{U}(w) = 1$
if $w \in U$ and $0$ otherwise, and $Z= \sum_{w \in W} e^{{\beta_{1}
{\bf 1}_{U_1}(w) + \ldots +  
\beta_{n} {\bf  1}_{U_n}}(w)} \mbox{$P_W$}(w)$ is a normalization
factor.
Let
$\alpha_i = P_W^{\vec{\beta}}(U_i)$
for $i = 1, \ldots, n$.
By \cite[Theorems 2.1 and 3.1]{Csiszar75}, it follows that
\begin{equation}
\label{eq:hideho}
P_W( \cdot \mid \alpha_1 U_1 ; \ldots ; \alpha_n U_n) =
P_W^{\vec{\beta}};
\end{equation}
Moreover, for
each vector $(\alpha_{1}, \ldots, \alpha_{n})
\in \Delta$,
there is a 
vector
$\vec{\beta} =
(\beta_{1},\ldots, \beta_{n}) \in {\bf R}^n$ such that
(\ref{eq:hideho}) holds.
\commentout{
we have that
for each vector $(\alpha_{1}, \ldots, \alpha_{n})$ with
$\alpha_1,\ldots, \alpha_n \in (0,1)$ such that 
there exists a distribution $P_W'$ with $P_W'(U_1) =
\alpha_1, \ldots, P_W'(U_n) = \alpha_n$, there is a 
vector $(\beta_{1},\ldots, \beta_{n}) \in {\bf R}^n$ such that, for all $w \in
W$,
\begin{equation}
\label{eq:hideho}
\mbox{$P_W$}(w \mid \alpha_1 U_1 ; \ldots ; \alpha_n U_n) =
\frac{1}{Z}e^{\beta_{1} {\bf 1}_{U_1} + \ldots + \beta_{n} {\bf
 1}_{U_n}} \mbox{$P_W$}(w),
\end{equation}
where $Z$ is again a normalization factor.
Conversely, for each $(\beta_{1},\ldots, \beta_{n}) \in {\bf R}^n$
there exists $(\alpha_1, \ldots, \alpha_n)$ with $0 <
\alpha_i < 1$ such that (\ref{eq:hideho}) holds.
}
(For an informal and easy derivation of 
(\ref{eq:hideho}), see
\cite[Chapter 9]{CoverThomas}.)
\lem
\label{clm:csiszar}
Let $C= \alpha_1 U_1; \ldots ; \alpha_n U_n$ 
for some $(\alpha_1, \ldots, \alpha_n) \in 
\Delta$.
Let $(\beta_1,\ldots, \beta_n)$ be a
vector such that (\ref{eq:hideho}) holds 
for $\alpha_1,\ldots, \alpha_n$. 
If 
$\beta_i = 0$ for some $i \in \{1, \ldots, n\}$, then 
$$P_W(U_i  \mid \alpha_1 U_1; \ldots ;
\alpha_{i-1} U_{i-1} ; \alpha_{i+1} U_{i+1} ; \ldots ;
\alpha_n U_n ) = \alpha_i.$$ 
\elem
\prf
Without loss of generality, assume that $\beta_1 = 0$. 
Taking $\alpha_i' = P_W^{\vec{\beta}}(U_i)$ for $i = 2, \ldots, n$, it
follows from (\ref{eq:hideho}) that 
$$\mbox{$P_W$}(w \mid \alpha'_2 U_2 ; \ldots ; \alpha'_n U'_n) =
\frac{1}{Z}e^{\beta_{2} {\bf 1}_{U_2} + \ldots + \beta_{n} {\bf
 1}_{U_n}} \mbox{$P_W$}(w),$$
so that 
$$
P_W(\cdot \mid \alpha_1 U_1 ; \ldots ;\alpha_n U_n)
= P_W(\cdot \mid \alpha'_2 U_2 ; \ldots ; \alpha'_n U_n).$$
Since 
$P_W(U_i \mid \alpha'_2 U_2; \ldots; \alpha'_n U_n) =
\alpha'_i$ and $P_W(U_i \mid \alpha_1 U_1 ; \ldots \alpha_n U_n) =
\alpha_i$
for $i = 2, \ldots, n$,
we have that $\alpha_i = \alpha'_i$
for $i = 2, \ldots, n$.
Thus, $P_W(\cdot
\mid 
\alpha_2 U_2 ; \ldots ; \alpha_n U_n)
= P_W(\cdot \mid \alpha_1 U_1 ; \ldots ; \alpha_n U_n )$ and, in particular,
$$\alpha_1 = P_W(U_1 \mid \alpha_1 U_1 ; \ldots ; \alpha_n U_n) = P_W(U_1 \mid
\alpha_2 U_2 ; \ldots ; \alpha_n U_n).
$$
\eprf

\medskip

\commentout{

\opro{pro:noMREchar}
Let $V_i = \{ w_i \}$ so that $U_1 = \{w_1, w_3 \}$ and $U_2 = \{ w_2,
w_3 \}$, let $\Pr_W(w) > 0$ for all $w \in W$ and let $\Pr_W(\cdot \mid X_O = C)$ be constructed as above. 
If $C$ is not Jeffrey-like, then $\Pr(X_O = C \mid
X_W = w_1) \neq  \Pr(X_O = C \mid
X_W = w_3)$ and $\Pr(X_O = C \mid
X_W = w_2) \neq  \Pr(X_O = C \mid
X_W = w_3)$.
\eopro

\prf 
By construction, for $i = 1,2,3$, $\Pr(X_W = w_i
| X_O = C) = \Pr_W(w_i \mid C)$ and hence, by Bayes' rule, 
\begin{equation}
\label{eq:boink}
\Pr(X_O = C|
X_W = w_i) = \frac{\Pr_W(w_i \mid C) \Pr(X_O = C)}{\Pr(X_W = w_i)}.
\end{equation}
By (\ref{eq:hideho})  
we have $\Pr_W(w_1 \mid C) = Z^{-1} \exp(\beta_1) \Pr_W(w_1), \Pr_W(w_2 \mid C) =
Z^{-1} \exp(\beta_2) \Pr_W(w_2)$, $\Pr_W(w_3 \mid C) = Z^{-1} \exp(\beta_1 +
\beta_2) \Pr_W(w_3)$, and, by (\ref{eq:boink}),
$$
\frac{\Pr(X_O = C \mid X_W = w_3)}{\Pr(X_O = C \mid X_W = w_1)} = \exp(\beta_2)
\ \ ; \ \ 
\frac{\Pr(X_O = C \mid X_W = w_3)}{\Pr(X_O = C \mid X_W = w_2)} = \exp(\beta_1)
$$ 
Thus, using Claim~\ref{clm:csiszar}, we see that if $C$ is not
Jeffrey-like, then $\Pr_W(X_O = C \mid X_W = w_j) \neq \Pr_W(X_O = C |
X_W = w_3)$ for $j = 1,2$.
\eprf
}

\othm{thm:MREnotOK}
Given a set $\R$ of runs and a set $O = \{C_1, C_2\}$ of observations,
where $C_i = \alpha_{i1} U_1; \alpha_{i2} U_2$, for $i = 1,2$,
let $\Pr$ be a distribution on $\R$ such that
$\Pr(X_O = C_1)$, $\Pr(X_O = C_2) > 0$,
and $\Pr_W(w) = \Pr(X_W = w) > 0$ for all $w \in W$. 
Let $\Pr^i = \Pr(\cdot \mid X_O = C_i )$, and let 
$\Pr^i_W$ be the marginal of $\Pr^i$ on $W$. 
If either $C_1$ or $C_2$ is not 
Jeffrey-like,
then we cannot have $\Pr^i_W = \Pr_W(\cdot \mid  C_i)$,
for  both $i = 1, 2$. 
\commentout{
Given a set $\R$ of runs and a set $O = \{C_1, C_2\}$ of observations,
where $C_i = \alpha_{i1} U_1; \alpha_{i2} U_2$, for $i = 1,2$,
let $\Pr$ be a distribution on $\R$ such that
$\Pr(X_O = C_1)$, $\Pr(X_O = C_2) > 0$,
and $\Pr_W(w) > 0$ for all $w \in W$. 
Let $\Pr^i = \Pr(\cdot \mid \R[\<C_i\>])$, and let 
$\Pr^i_W$ be the marginal of $\Pr^i$ on $W$. 
If either $C_1$ or $C_2$ is not 
Jeffrey-like,
then we cannot have $\Pr^i_W = \Pr_W(\cdot \mid  C_i)$,
for  both $i = 1, 2$. 
}
\eothm
\commentout{
Given a set $\R$ of runs and a set $O = \{C_1, C_2\}$ of observations,
let $\Pr$ be a distribution on $\R$ such that
$\Pr(X_O = C_1)$, $\Pr(X_O = C_2) > 0$,
and $\Pr_W(w) > 0$ for all $w \in W$.   
Let $\Pr^i = \Pr(\cdot \mid \R[\<C_i\>])$, and let 
$\Pr^i_W$ be the marginal of $\Pr^i$ on $W$. 
If either $C_1$ or $C_2$ is not 
Jeffrey-like,
then we cannot have $\Pr^i_W = \Pr_W(\cdot \mid  C_i)$,
for  both $i = 1, 2$. 
\eothm}

\prf 
Let $V_1 = U_1 - U_2, V_2 = U_2 - U_1, V_3 = U_1 \cap U_2$, and $V_4 =
W - (U_1 \cup U_2)$. 
Since $V_1, V_2, V_3, V_4$ are all assumed to be nonempty, we have
$\Delta = (0,1)^2$, where $\Delta$ is defined as above, that is, $\Delta$
is the set $(\alpha_1,\alpha_2)$ such that there exists a distribution
$P_W$ with $P_W(U_1) = \alpha_1, P_W(U_2) = \alpha_2$, 
$P_W(w) > 0$ for all $w \in W$.
If
 $\Pr^i_W = \Pr_W(\cdot \mid  C_i)$ for $i = 1,2$, then
\begin{equation}
\label{eq:bonjour}
\lambda  \mbox{$\Pr_W$}(\cdot |C_1) + (1 - \lambda)
\mbox{$\Pr_W$}(\cdot \mid C_2) =   \mbox{$\Pr_W$},
\end{equation} 
where
$\lambda =  \Pr(X_O = C_1)$. 
We prove the theorem by  showing that (\ref{eq:bonjour})
cannot hold if either $C_1$ or $C_2$ is not Jeffrey-like.
Since we have assumed that $(\alpha_{i1},\alpha_{i2}) \in (0,1)^2 =
\Delta$ for $i= 1,2$, we can apply
(\ref{eq:hideho}) to $C_i$ for $i = 1,2$.
Thus, there are vectors
$(\beta_{i1},\beta_{i2}) \in {\bf R}^2$ 
for $i = 1, 2$ 
such that, for all $w \in W$,
\begin{equation}
\label{eq:hidehob}
\mbox{$\Pr_W$}(w |C_i) = \frac{1}{Z_i}e^{\beta_{i1}
{\bf 1}_{U_1} + \beta_{i2} {\bf 1}_{U_2}}
 \mbox{$\Pr_W$}(w).
\end{equation}
(\ref{eq:hidehob}) implies that
 $\Pr_W(V_1| C_i) = Z^{-1}_i
e^{\beta_{i1}}
\Pr_W(V_1)$, $\Pr_W(V_2| C_i) = Z^{-1}_i e^{\beta_{i2}} \Pr_W(V_2)$,
$\Pr_W(V_3| C_i) = Z^{-1}_i e^{\beta_{i1}+ \beta_{i2}} \Pr_W(V_3)$, 
$\Pr_W(V_4| C_i) = Z^{-1}_i  \Pr_W(V_4)$. Plugging this into
(\ref{eq:bonjour}), we obtain the following four equations:
\begin{eqnarray}
\label{eq:adieu}
\mbox{$\Pr_W$}(V_1) & = & \lambda \frac{e^{\beta_{11}}}{Z_1}
\mbox{$\Pr_W$}(V_1) +  (1- \lambda)  
\frac{e^{\beta_{21}}}{Z_2} \mbox{$\Pr_W$}(V_1) \nonumber \\
\mbox{$\Pr_W$}(V_2) & = & \lambda \frac{e^{\beta_{12}}}{Z_1}
\mbox{$\Pr_W$}(V_2) +  (1- \lambda)  
\frac{e^{\beta_{22}}}{Z_2} \mbox{$\Pr_W$}(V_2) \nonumber \\
\mbox{$\Pr_W$}(V_3) & = & 
\lambda \frac{e^{\beta_{11} +\beta_{21}}}{Z_1} \mbox{$\Pr_W$}(V_3) +
(1- \lambda)\frac{e^{\beta_{21} + \beta_{22}}}{Z_2}
\mbox{$\Pr_W$}(V_3) \nonumber \\ 
\mbox{$\Pr_W$}(V_4) & = & \lambda \frac{1}{Z_1}
\mbox{$\Pr_W$}(V_4) +  (1- \lambda) 
\frac{1}{Z_2} \mbox{$\Pr_W$}(V_4).
\end{eqnarray}
Since we have assumed that $\Pr(w) > 0$ for all $w \in W$, it must be
the case that $\Pr_W(V_i) > 0$, for $i = 1, \ldots, 4$.  
Thus, $\Pr(V_i)$ factors out of the $i$th equation above.  
By the change of variables $\mu = \lambda / Z_1$, $1 - \mu   =
(1-\lambda) / Z_2$, $\epsilon_{ij} = e^{\beta_{ij}} -1$ and some
rewriting, we see that  (\ref{eq:adieu}) is equivalent to
\begin{eqnarray}
  \label{eq:simpler}
 0 & = & \mu \epsilon_{11} + (1 - \mu) \epsilon_{21} \nonumber \\
 0 & = & \mu \epsilon_{12} + (1 - \mu) \epsilon_{22} \nonumber \\
 0 & = & \mu (\epsilon_{11} + \epsilon_{12} + \epsilon_{11}
 \epsilon_{12}) + 
(1 - \mu) (\epsilon_{21} + \epsilon_{22} + \epsilon_{21}
 \epsilon_{22}).
\end{eqnarray}
If, for some $i$, both $\epsilon_{i1}$ and $\epsilon_{i2}$ are
nonzero, then the three equations of (\ref{eq:simpler}) have no
solutions for $\mu \in (0,1)$. Equivalently, 
if for some $i$, both $\beta_{i1}$ and $\beta_{i2}$ are
nonzero, then the four equations of (\ref{eq:adieu}) have no
solutions for $\lambda  \in (0,1)$. So it only remains to
show that for some $i$, both $\beta_{i1}$ and $\beta_{i2}$ are
nonzero. To see this, note that by assumption for some $i$, $C_i$ is not Jeffrey-like. But then
it follows from Lemma~\ref{clm:csiszar} above that both  $\beta_{i1}$
and $\beta_{i2}$ are 
nonzero.  Thus, the theorem is proved.
\eprf

\commentout{
\ 
\\
\noindent
\paragraph{Proof (of Theorem~\ref{thm:CARalmostnever})}
By applying (\ref{eq:hideho}), it follows 
that for each $C_i$, for all $w \in W$,
\begin{equation}
\label{eq:hidehoc}
\mbox{$\Pr_W$}(w |C_i) = \frac{1}{Z_i}\exp(\sum_{j=1}^n \beta_{ij}
{\bf 1}_{w \in U_j})
 \mbox{$\Pr_W$}(w).
\end{equation}
Without loss of generality, let $U_1$ be such that  ${\bf U}(w_1) =
{\bf U}(w_0) \cup \{ U_1 \}$.
In order to have
 $\Pr^i_W = \Pr_W(\cdot \mid  C_i)$ for $i = 1..k$, we must have 
\begin{equation}
\label{eq:ciao}
\sum_{i=1}^k \lambda_i \mbox{$\Pr_W$}(w |C_i) =
 \mbox{$\Pr_W$}(w)
\end{equation} 
for all $w \in  W$, in particular for $w \in \{w_0,w_1 \}$. By
(\ref{eq:hidehoc}) the corresponding instances of (\ref{eq:ciao})
become
\begin{eqnarray}
\label{eq:hullob}
\sum_{i=1}^k \lambda_i f_i 
 \mbox{$\Pr_W$}(w_0)  & = & \mbox{$\Pr_W$}(w_0) \\
\label{eq:hulloa}
 \sum_{i=1}^k \lambda_i f_i  \exp(\beta_{i1})
 \mbox{$\Pr_W$}(w_1)  & = & \mbox{$\Pr_W$}(w_1) 
\end{eqnarray}
where $f_i = Z_i^{-1} \exp(\sum_{\{j: w_0 \in U_j\}} \beta_j)$.
We must also have
\begin{equation}
\label{eq:hulloc}
\sum_{i=1}^k \lambda_i = 1.
\end{equation}
If the $f_i$'s are not all identical, then (\ref{eq:hullob}) and
(\ref{eq:hulloc}) are linearly independent. Then the set of
$(\lambda_1,\ldots, \lambda_k)$ satisfying both these equations must have
dimension $k-2$, and we are done. If the $f_i$ are all identical and
not equal to $1$, there is no $(\lambda_1,\ldots, \lambda_k)$
satisfying both equations and we are done. The only remaining
possibility is that all $f_i$ are equal to $1$. In that case, if
furthermore not all $\beta_{i1}$ are identical,   (\ref{eq:hulloa}) and
(\ref{eq:hullob}) are linearly independent and  we are done. If all
$\beta_{i1}$ are identical and not equal to $0$, there is no  $(\lambda_1,\ldots, \lambda_k)$
satisfying both equations and we are done. If all 
$\beta_{i1}$ are identical and equal to $0$, then all $C_i$ must be
Jeffrey-like by the Claim inside the proof of 
Theorem~\ref{thm:MREnotOK}.
}

\bibliographystyle{chicago}
\bibliography{z,joe,bghk,refs}

\end{document}